\documentclass[10pt,twocolumn,letterpaper]{article}

\usepackage[pagenumbers]{cvpr} %

\usepackage{cuted} %

\usepackage{currfile} %

\usepackage{caption} %

\makeatletter
\renewcommand\paragraph{\@startsection{paragraph}{4}{\z@}%
  {2mm \@plus1ex \@minus.2ex}%
  {-1em}%
  {\normalfont\normalsize\bfseries}}
\makeatother
\newcommand{\finding}[1]{\paragraph{\textcolor{blue!55!black}{#1}}}

\definecolor{cvprblue}{rgb}{0.21,0.49,0.74}
\usepackage[pagebackref,breaklinks,colorlinks,allcolors=cvprblue]{hyperref}

\newcommand{\name}{SceneTeract}
\usepackage{algorithm}
\usepackage{algpseudocodex}
\usepackage{xcolor}
\usepackage{tabularx}
\usepackage{multirow}
\usepackage{makecell}
\usepackage{ragged2e}
\usepackage[table]{xcolor}
\usepackage{circledsteps}
\usepackage{bbm}
\newcommand{\Step}[1]{\Circled{\small$#1$}}

\usepackage{float}
\usepackage{array}
\usepackage{siunitx}
\usepackage{tikz}
\usepackage[breakable]{tcolorbox}

\title{\name{}: Probing and Improving Agent-Aware Activity Reasoning\\in 3D Indoor Scenes}

\author{
Léopold~Maillard\textsuperscript{1,2,}\thanks{Work done during a research visit to Stanford University.}\quad
Francis~Engelmann\textsuperscript{3,4}\quad
Tom~Durand\textsuperscript{2}\quad
Boxiao~Pan\textsuperscript{3}\quad\\
Yang~You\textsuperscript{3}\quad
Leonidas~Guibas\textsuperscript{3}\quad
Maks~Ovsjanikov\textsuperscript{1}
\vspace{1mm}
\\
\textsuperscript{1}École~Polytechnique\quad
\textsuperscript{2}Dassault~Systèmes\quad
\textsuperscript{3}Stanford University\quad
\textsuperscript{4}USI Lugano
\vspace{1mm}
\\
\normalsize\url{https://sceneteract.github.io/}
}

\begin{document}
\addtocontents{toc}{\protect\setcounter{tocdepth}{-1}}
\maketitle
\begin{strip}
\centering
\vspace{-5.1em}
\includegraphics[width=\linewidth]{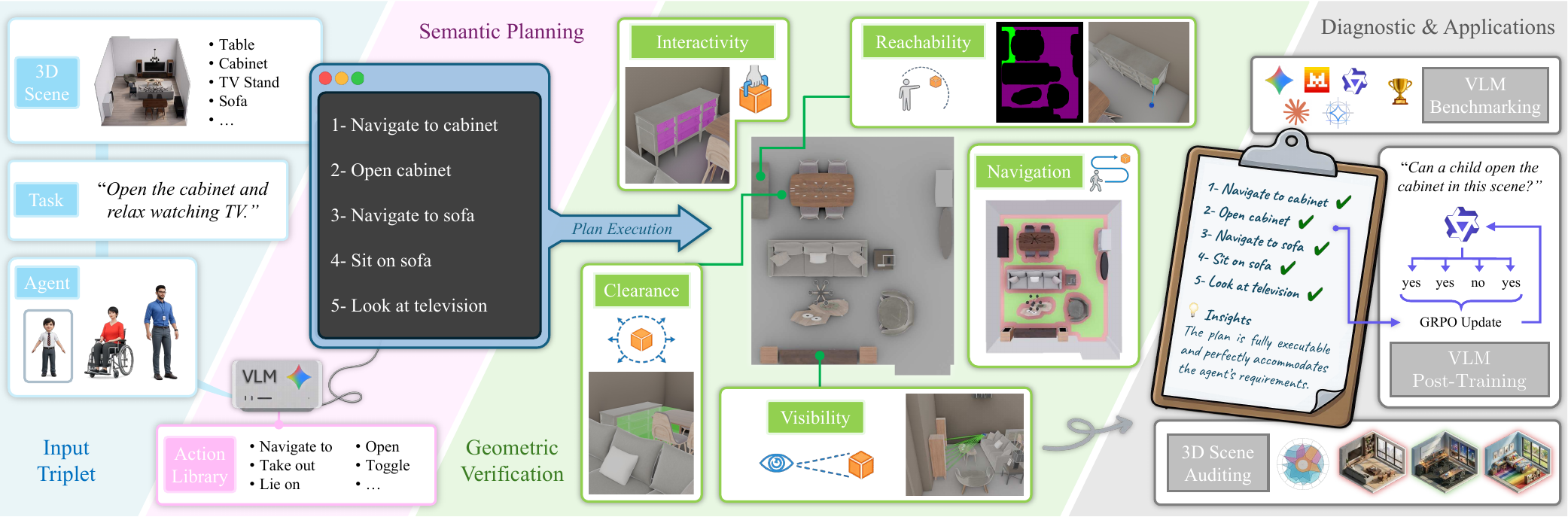}
\captionsetup{hypcap=false}
\captionof{figure}{\name{} verifies whether a 3D scene physically supports an activity for a specific embodied agent. Given a 3D scene, an agent profile, and a target task, a VLM decomposes the activity into atomic actions, and explicit geometric checks validate each step under the agent's properties. Derived from scene geometry, the resulting labels let us (i) uncover accessibility bottlenecks in scenes that \textit{look} realistic, (ii) expose how frontier VLMs hallucinate action feasibility, and (iii) post-train VLMs whose gains transfer to \textit{real} scenes.}
\captionsetup{hypcap=true}
\label{fig:\currfilebase}%
\end{strip}

\begin{abstract}
  Indoor 3D scenes are ultimately meant to be used: an embodied agent should be able to navigate, reach objects, and complete diverse activities. Yet whether a given scene actually supports these activities for a specific agent profile is rarely verified. Existing evaluations of indoor 3D scenes typically focus on visual quality and semantic plausibility.
  In contrast, the feasibility of an activity depends on geometric, agent-specific constraints such as reach, clearance, and navigable space availability. These properties are not captured by visual plausibility metrics, and, as we show, VLMs, which are increasingly used to reason about 3D scenes, often fail to determine action feasibility in a single shot. We present \name{}, a verification interface that separates semantic action understanding from physical feasibility. 
  Given a scene, an activity, and an embodied agent profile, we decompose the activity into atomic actions on scene objects. Explicit \textit{geometric checks} then decide whether each step is executable and return a diagnostic trace explaining failures. In synthetic indoor scenes, \name{} reveals widespread functional and accessibility failures across diverse agent profiles. Moreover, when benchmarked against our verification, we find that existing VLMs systematically over-predict action feasibility, highlighting limited awareness of embodied functional constraints. In response, we post-train a lightweight VLM with verifier feedback, improving its assessment of physical feasibility. Although trained only on renders of synthetic scenes, we demonstrate that scene understanding improvements also generalize to real-world scenes. We will release our verification suite, benchmark labels, and diagnostic trace datasets.
\end{abstract}

\section{Introduction}
\label{sec:intro}

Large-scale 3D datasets~\cite{3dfront,hypersim,hm3d}, generative scene models~\cite{atiss,diffuscene,debara}, and reconstruction pipelines~\cite{mao2025spatiallm,huang2025litereality} have made furnished indoor environments widely available, while Vision-Language Models (VLMs) provide strong priors for object use, task interpretation, and commonsense scene reasoning~\cite{yang2025thinking,yu2025seqafford,layoutgpt,holodeck}. A VLM can often infer that \emph{making tea} involves a mug, that a table is a plausible support surface, and that the activity requires interacting with multiple objects in sequence. Yet this semantic understanding does not answer a basic embodied question: can a specific agent physically execute the activity in the scene?

\begin{figure*}[t]
  \includegraphics[width=\textwidth]{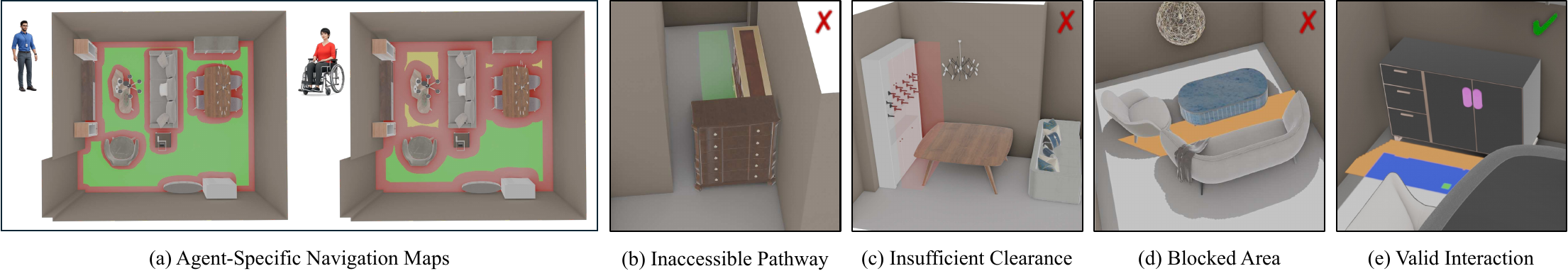}
  \caption{\textbf{Functional patterns and failure modes detected by \name{}}. \textbf{(a)} Navigation maps highlight restricted areas (yellow) disconnected from the main area (green) for agents with larger occupancy radii (impacting the red border). \textbf{(b)} A blocked hallway prevents access to the target cabinet, even though its local interaction clearance (green box) is valid. \textbf{(c)} Insufficient space (red box) to articulate the object. \textbf{(d)} Although the object is partially reachable, the identified interaction area (orange) to \textit{sit} on the couch is obstructed. \textbf{(e)} The identified interactive volume (pink) to \textit{open} the cabinet is within the agent's reach from a valid interaction zone (blue).}
  \label{fig:checks}
  \vspace{-1em}
\end{figure*}

This distinction between semantic plausibility and physical executability is a missing evaluation axis for both 3D scenes and VLMs. A scene may look realistic while containing a blocked hallway, insufficient cabinet clearance, an unreachable shelf, or an interaction zone that is inaccessible to some specific users (Fig.~\ref{fig:checks}). Likewise, a VLM may correctly recognize the relevant objects and still misjudge feasibility because it does not check geometric constraints such as reach, clearance, visibility, or navigable free space. Both gaps have similar roots. Scene evaluation relies on perceptual metrics (\eg, FID~\cite{fid}) and semantic alignment scores~\cite{sceneeval}, both computed based on appearance. A blocked doorway or a few missing centimeters of clearance do not directly translate into significant changes in these metrics~\cite{physcene}, yet either one can make an activity impossible. VLMs, in turn, learn from web images and text in which precise feasibility is rarely given. We refer to the resulting failure mode as \emph{physical hallucination}: confident feasibility judgments that are not backed by geometry.

The problem is inherently agent-conditioned. Affordance is not an intrinsic property of a scene but a relational one, holding between the scene, an activity, and an embodied agent~\cite{gibson2014ecological}. Using the same counter, cabinet, or seat can be effortless for a standing adult, difficult for a child, and impossible for a wheelchair user. Functional evaluation therefore requires explicit agent profiles that specify dimensions, reach, and locomotion mode, together with a concrete procedure for checking whether the relevant physical constraints are satisfied.

How, then, should executability be judged? The most direct option is to ask a VLM whether the activity is possible in the scene. Such judgments inherit the model's perceptual bias and, as we show, often conflate plausibility and feasibility. Yet VLMs are strong at the semantic side of the problem. We therefore introduce \name{}, a verification interface where the VLM produces the semantic proposal and a geometric engine tests this proposal for feasibility (Fig.~\ref{fig:teaser}). Given an open-ended activity (\eg, \emph{prepare a cup of tea and then watch TV}), we decompose it into individual steps, identify the objects each step involves, and localize the functional parts. Each proposed action then triggers a fixed sequence of geometric checks, such as navigability, reachability, and clearance, computed with explicit 3D tools on the scene geometry and conditioned on the agent profile. The output is not a single feasibility label but a diagnostic trace that identifies and quantifies the potential physical limitations. VLM semantics thus remain where they help, while the feasibility decision itself is made by measurement.

Applied at scale, \name{} yields a benchmark of feasibility traces that we deploy in three complementary applications, bridging semantic scene understanding and physical evidence. To summarize, our contributions are:

\begin{enumerate}
  \item \textbf{A missing evaluation axis.} We identify agent-conditioned physical executability as a missing evaluation axis for 3D scenes and VLMs, and demonstrate that plausibility-based judgments fail to capture it.
  \item \textbf{A verification interface.} We present \name{}, a verification interface that maps open-ended activities to verifiable atomic actions and produces executability labels with diagnostic traces that quantify each failure.
  \item \textbf{Benchmark findings.} Using \name{}, we audit synthetic scenes and benchmark frontier and open VLMs, revealing widespread accessibility failures in scenes and systematic physical hallucination.
  \item \textbf{Verifier-feedback improvement.} We show that verifier feedback improves a lightweight VLM's physical feasibility predictions and generalizes from synthetic to real-world scenes, which we evaluate against human labels.
\end{enumerate}

\section{Related Work}
\label{sec:related}

\paragraph{\textbf{Indoor 3D Scenes and Synthesis Methods.}}

Large-scale indoor 3D datasets~\cite{3dfront,hypersim,hm3d} have enabled realistic simulation in virtual environments~\cite{ai2thor,shen2021igibson,savva2019habitat} and rapid progress in data-driven scene synthesis methods~\cite{atiss,diffuscene,debara}. Some approaches improve the practicality of synthesized environments by injecting physical priors (\eg, object spacing, collision, stability, and ergonomic preferences) as additional constraints~\cite{layoutenhancer,kan2018automatic,vitsas2020illumination,xia2026sage,jin2026behavior}. Others generate spatial configurations from specific human motion sequences~\cite{ye2022scene,yi2023mime,hong2024human, li2024physics}, which is rigid and difficult to scale. Concurrent work synthesizes layouts from functional specifications with a check-and-repair loop~\cite{wang2026function2scene}. Instead, we verify whether a scene supports an open-ended activity for a specific embodied agent and use that signal to benchmark and post-train VLMs.

\paragraph{\textbf{Evaluating 3D Environments.}}

Evaluation of 3D scenes is still largely driven by perceptual proxies such as FID~\cite{fid}, KID~\cite{kid}, and Scene Classification Accuracy (SCA)~\cite{ritchie2019fast}, as well as CLIP-based semantic alignment measures~\cite{clipscore}, all computed on 2D renderings that discard much of the spatial and geometric context. Additionally, methods often report indicators of local physical realism, such as rates of overlapping, out-of-bound, or symmetric objects~\cite{legonet, debara, physcene}. In contrast, adjacent vision domains have increasingly adopted holistic benchmarks designed around downstream utility, such as VBench for video generation~\cite{vbench}, MVGBench for multi-view synthesis~\cite{mvgbench}, and WorldScore for world modeling~\cite{worldscore}. For multi-object 3D environments, SceneEval~\cite{sceneeval} is an early step in this direction, combining explicit fidelity metrics with physical plausibility checks, but it does not evaluate whether a specific agent can execute open-ended activities under embodiment constraints. Since supporting activities is the end goal of indoor 3D environments, we argue that verification with respect to the intended embodied user should be prioritized.

\paragraph{\textbf{VLMs in Planning and Physical Reasoning.}}

General-purpose VLMs exhibit strong perception, open-ended scene understanding, and generation capabilities~\cite{gemini,yang2025thinking,yu2025seqafford,holodeck,idesign}, excelling at tasks driven by rich semantic priors, such as high-level action planning~\cite{huang2022language,ahn2022can} and affordance inference~\cite{corsetti2025functionality,yu2025seqafford,qian2024affordancellm}.  In robotics, SayCan~\cite{ahn2022can} and RoboBrain~\cite{ji2025robobrain} combine language decomposition with learned affordance cues to inform robot action execution. In contrast, we decompose a requested activity into atomic actions, verify whether each is physically executable for a specific embodied agent, and use the resulting labels for evaluation and training. At the object level, SceneFun3D~\cite{delitzas2024scenefun3d} studies fine-grained functional-part understanding, a capability \name{} incorporates through training-free part localization~\cite{corsetti2025functionality} as one check within activity-level verification. However, a gap remains in physical and spatial grounding: VLMs can produce confident but physically infeasible judgments in 3D settings~\cite{daxberger2025mm,chen2024spatialvlm}. Recent post-training and reinforcement-learning strategies target spatial and functional reasoning more directly~\cite{pan2026metaspatial,wang2025affordance,liu2026omnieva}. Yet these methods address spatial reasoning for scene coherence, object-level affordances, or robot kinematic feasibility, without verifying open-ended activities across diverse embodied agents. We use instead an agent-aware geometric verifier as a GRPO reward signal~\cite{shao2024deepseekmath}, distilling the relational nature of 3D affordance into VLM reasoning.

\paragraph{\textbf{Tool-Augmented Verification for Spatial Tasks.}}

Recent work increasingly follows a hybrid paradigm for spatial intelligence: VLMs provide broad semantic interpretation, while specialized modules enforce spatial and physical consistency~\cite{mover,fireplace,mobiagent}. Among them, MoVer~\cite{mover} validates and repairs generated graphics animations under spatio-temporal constraints. Closer to our domain, FirePlace~\cite{fireplace} targets 3D object placement in indoor scenes by combining multimodal commonsense priors with explicit geometric refinement and constraint solving, directly coupling semantic plausibility with scene-level physical feasibility. Recently, SpatialCLAW~\cite{cho2026spatialclaw} argued that the interface through which a VLM invokes perception and geometry tools largely determines its spatial reasoning ability. \name{} extends this paradigm to an agent-aware, activity-level verification interface, grounding each step of an open-ended plan against the physical constraints of a target user.

\section{Method}

We detail the \name{} verification interface, illustrated in Fig.~\ref{fig:teaser}. The key idea is to separate semantic activity planning from physical verification: VLMs interpret open-ended activity descriptions and propose action-object steps, while explicit geometric checks decide whether those steps are executable for a specified agent. Given a 3D scene, an agent profile, and a target activity, \name{} produces geometry-derived feasibility labels with diagnostic traces quantifying physical bottlenecks.

\subsection{Task-Agent-Scene Input Triplet}

\begin{table*}[t]
\centering
\caption{\textbf{Atomic actions to boolean properties mapping.} The left panel defines the boolean properties checked by the verifier. The right panel maps them to the action families (Tab.~\ref{tab:action_taxonomy}), where the numbers \Step{1}-\Step{4} denote the verification sequence.}
\label{tab:interface_integrated}
\setlength{\aboverulesep}{0.2ex}
\setlength{\belowrulesep}{0.3ex}
\renewcommand{\arraystretch}{0.9}
\resizebox{\textwidth}{!}{%
\begin{tabular}{
    l                                       %
    @{\hspace{0.05cm}}                      %
    >{\centering\arraybackslash}m{0.9cm}    %
    m{0.75\textwidth}                       %
    *{4}{>{\centering\arraybackslash}m{0.03\textwidth}} %
    }
\toprule
\multicolumn{3}{c}{\textsc{\textbf{Boolean Property Dictionary ($\mathbb{P}$)}}} & \multicolumn{4}{c}{\textsc{\textbf{Actions ($\mathbb{A}$)}}} \\
\cmidrule(lr){1-3} \cmidrule{4-7}
\multicolumn{2}{c}{\textsc{Symbol}} & \textsc{Verification Implementation Overview} &
$\mathbb{A}_{\text{m}}$ &
$\mathbb{A}_{\text{c}}$ &
$\mathbb{A}_{\text{h}}$ &
$\mathbb{A}_{\text{p}}$ \\
\midrule

$\mathcal{P}_{\text{nav}}$ &
\includegraphics[width=0.80cm]{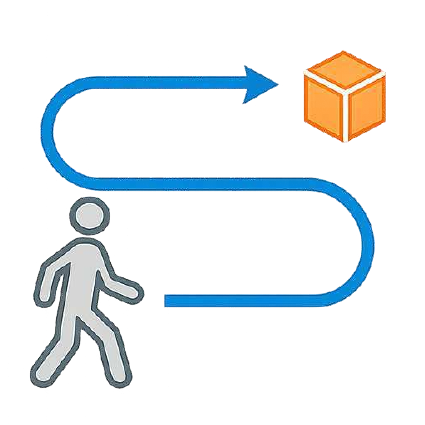} &
\footnotesize \texttt{isNavigableTo}:
Derives an interaction zone around the target object $o$ and verifies path connectivity from the agent to this zone on the 2D navigation map~\cite{physcene} computed using agent width $w_{\text{clear}}$. &
\Step{1} & \Step{1} & \Step{1} \\
\midrule

$\mathcal{P}_{\text{reach}}$ &
\includegraphics[width=0.80cm]{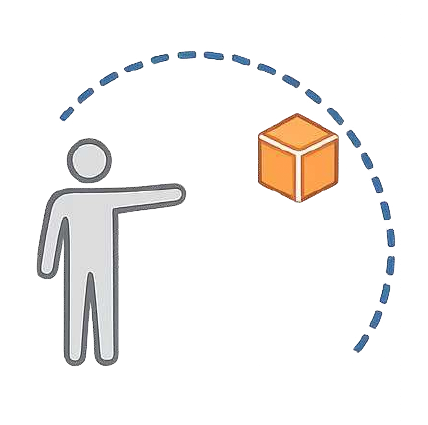} &
\footnotesize \texttt{isReachable}: Computes the minimum Euclidean distance from the agent's connected navigable floor region to the target mesh, vertically translated by $h_{\text{arm}}$, and validated against the maximum reach radius $r_{\text{arm}}$. &
& \Step{2} & \Step{2} & \\
\midrule

$\mathcal{P}_{\text{inter}}$ &
\includegraphics[width=0.80cm]{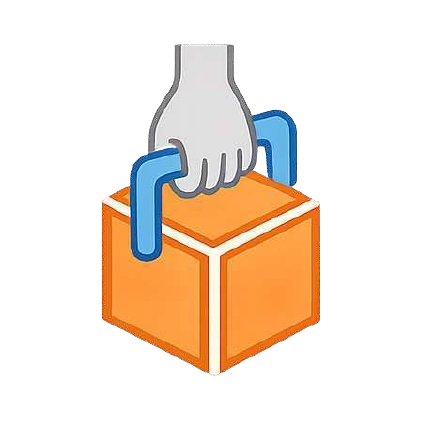} &
\footnotesize \texttt{isInteractable}: Performs a 3D segmentation of target object functional parts (\eg, handles), inspired by~\cite{corsetti2025functionality}. Verifies if the deprojected 3D functional volume is within reach distance $r_{\text{arm}}$. &
& & \Step{3} \\
\midrule

$\mathcal{P}_{\text{clear}}$ &
\includegraphics[width=0.80cm]{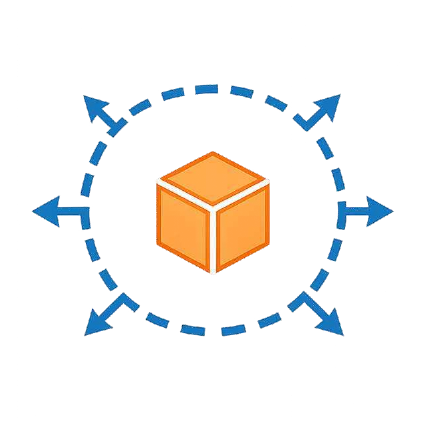} &
\footnotesize \texttt{hasClearance}: Volumetric check in front of target (\eg, doors/drawers) to ensure sufficient kinematic space for articulation. &
& & \Step{4} \\
\midrule

$\mathcal{P}_{\text{vis}}$ &
\includegraphics[width=0.80cm]{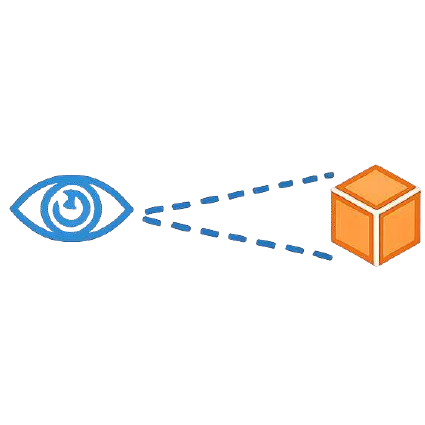} &
\footnotesize \texttt{isVisible}: Line-of-sight raycasting from agent eye height $h_{\text{eye}}$ to target object mesh. Calculates visibility ratio to determine occlusion. &
& & & \Step{1} \\
\bottomrule
\end{tabular}%
}
\vspace{-0.9em}
\end{table*}

We define the verification context as an input triplet \(\langle \mathcal{T}, \mathcal{A}, \mathcal{S} \rangle\) composed of the activity description \(\mathcal{T}\), the agent profile \(\mathcal{A}\) and the scene representation \(\mathcal{S}\). First, \(\mathcal{T}\) consists of a natural language prompt describing a high-level embodied intent (\eg, \textit{retrieve the book from the upper shelf}). Unlike approaches that evaluate scenes without considering their intended users~\cite{sceneeval}, we explicitly model the agent-specific component through an embodied profile, \(\mathcal{A}\). We treat the agent as a parameterized physical entity \(\mathcal{A} = \langle w_{\text{clear}}, h_{\text{eye}}, h_{\text{arm}}, r_{\text{arm}}, m_{\text{loco}} \rangle\), where $w_{\text{clear}}$ denotes the clearance width required for collision-free navigation, $h_{\text{eye}}$ the eye level for visibility checks, $h_{\text{arm}}$ and $r_{\text{arm}}$ are respectively the vertical origin (\ie, shoulder height) and radial bound of the manipulation envelope, and $m_{\text{loco}} \in \{\text{walk}, \text{wheel}\}$ the locomotion mode. This parameterization is deliberately compact: rather than reproducing articulated body kinematics, it captures the few geometric quantities that decide whether common indoor interactions succeed. As we show in Sec.~\ref{experiments}, this simple proxy is sufficient to expose consistent physical bottlenecks and accessibility edge cases across embodiments. Finally, the agent acts within the 3D scene representation, denoted as \(\mathcal{S} = \langle \mathcal{O}, \mathcal{G} \rangle\). Here, \(\mathcal{G}\) represents the scene geometry and navigable surfaces, while \(\mathcal{O} = \{o_1, \dots, o_N\}\) is a set of objects, each defined by its semantic class, 3D bounding box, and mesh geometry.

\subsection{Verification Interface}

Our interface rests on two observations about embodied interaction: (i) complex indoor activities decompose into sequences of simple, \textit{atomic} agent-object interactions, and (ii) the success of each atomic action depends on a finite set of measurable physical constraints. We therefore define a closed vocabulary of atomic actions $\mathbb{A}$, organized into functional families (Tab.~\ref{tab:action_taxonomy}), where each action pairs with a target object to form an interaction tuple $(a,o)$, such as (\textit{open}, \textit{cabinet}). Each family maps to an ordered sequence of boolean properties, all implemented as a geometric check (Tab.~\ref{tab:interface_integrated}). This structure separates \textit{semantic decomposition} from \textit{geometric verification}.

\begin{table}[htbp]
\centering
\caption{\textbf{The \name{} Action Families.} We classify atomic actions into four functional families that dictate the sequence of geometric, boolean property checks required for feasibility verification, as detailed in Tab.~\ref{tab:interface_integrated}.}
\label{tab:action_taxonomy}
{\setlength{\tabcolsep}{7pt}%
\resizebox{1.\columnwidth}{!}{%
\begin{tabular}{l l l}
\toprule
\textsc{\textbf{Symbol}} & \textsc{\textbf{Family}} & \textsc{\textbf{Included Atomic Actions}} \\
\midrule
$\mathbb{A}_{\text{m}}$ & Mobility & \texttt{NavigateTo}, \texttt{SitOn}, \texttt{LieOn} \\
$\mathbb{A}_{\text{c}}$ & Contact & \texttt{Toggle}, \texttt{PickUpFrom}, \texttt{ReleaseOn} \\
$\mathbb{A}_{\text{h}}$ & Handling & \texttt{Open}, \texttt{Close}, \texttt{PutIn}, \texttt{TakeOutOf} \\
$\mathbb{A}_{\text{p}}$ & Perception & \texttt{LookAt} \\
\bottomrule
\end{tabular}%
}}
\vspace{-0.5em}
\end{table}

\paragraph{\textbf{Boolean Property Dictionary.}}

We define a compact dictionary of explicit boolean properties $\mathbb{P}$, where each property evaluates a physically meaningful condition for a candidate action-object pair under the current scene and agent profile. The core properties are $\mathcal{P}_{\text{nav}}$ (navigability), $\mathcal{P}_{\text{reach}}$ (reachability), $\mathcal{P}_{\text{inter}}$ (interactivity), $\mathcal{P}_{\text{clear}}$ (interaction clearance), and $\mathcal{P}_{\text{vis}}$ (visibility). Their geometric implementations and action-family usage are summarized in Tab.~\ref{tab:interface_integrated}.

\paragraph{\textbf{Action-Property Mapping.}}

A fixed action-property interface links $\mathbb{A}$ and $\mathbb{P}$: each action family invokes a small ordered checklist of properties rather than a learned black-box feasibility score. For example, actions in the \textit{Handling} family ($a \in \mathbb{A}_{\text{h}}$) first require navigability and reachability, then functional interactivity and interaction clearance. This ordering reflects the hierarchical dependencies of embodied interaction. The interface is intentionally fixed in this work: \name{} verifies a closed vocabulary of action-object requests using explicit geometric checks. This choice keeps every failure attributable to a specific constraint (navigation, reach, interaction, clearance, or visibility). Still, the interface remains easy to extend, as new actions can often be composed from existing checks. For example, \emph{stand on} can be mapped to navigability, height-parameterized reachability, and clearance. New checks, such as surface stability, can also be added without changing the rest of the pipeline.

\subsection{Execution Pipeline}
\label{sec:method-pipeline}

The \name{} evaluation pipeline is illustrated in Fig.~\ref{fig:teaser}. It proceeds in two sequential phases: \textit{semantic decomposition} and \textit{geometric verification}.

\begin{figure*}[t]
\includegraphics[width=\textwidth]{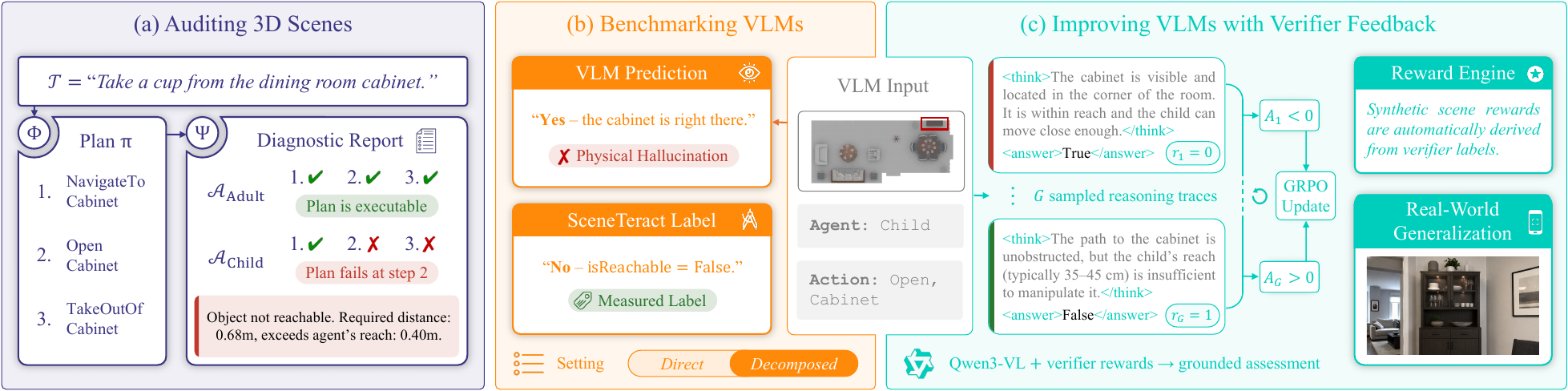}
\caption{Three applications of \name{}, shown for one task, \textit{take a cup from the dining room cabinet}. \textbf{(a) Auditing 3D scenes:} the VLM plan is verified per agent, and each failed step comes with a quantified cause, here the cabinet beyond the child's reach. \textbf{(b) Benchmarking VLMs:} for the same action (\ie, step 2), a VLM's judgment contradicts the measured label, an instance of physical hallucination. \textbf{(c) Improving VLMs:} given the same input, the model samples multiple reasoning traces, each rewarded by agreement with the verifier label. GRPO updates the policy toward grounded feasibility judgments that transfer to in-context images of real scenes.}
\label{fig:applications}
\end{figure*}

\paragraph{\textbf{Semantic Decomposition.}}

We utilize a VLM, denoted as $\Phi$, to perform high-level task planning. The model is conditioned on a system prompt that defines the task constraints and the available atomic action library $\mathbb{A}$. To ensure the generated plan is grounded in the agent's capabilities and the specific scene content, we construct a multimodal input context $\mathcal{C}$. This comprises the activity description $\mathcal{T}$, a textual summary of the agent features and object inventory (extracted from $\mathcal{O}$), and a top-down rendering. Formally, the multimodal input context is constructed as:
\begin{equation}
  \mathcal{C} = \langle \mathcal{T}, \Pi_{\text{txt}}(\mathcal{A}, \mathcal{S}), \Pi_{\text{vis}}(\mathcal{S}) \rangle\,,
\end{equation}
\noindent where $\Pi_{\text{txt}}(\cdot)$ and $\Pi_{\text{vis}}(\cdot)$ are, respectively, textual serializer and visual renderer functions. The VLM decomposes the activity into a semantic plan $\pi$:
\begin{equation}
  \pi = \Phi(\mathcal{C}) = \bigl[\bigl(a^{(1)}, o^{(1)}\bigr), \dots, \bigl(a^{(T)}, o^{(T)}\bigr)\bigr]\,,
\end{equation}
\noindent where $a^{(t)} \in \mathbb{A}$ and $o^{(t)} \in \mathcal{O}$. By combining textual and visual context, the planner prioritizes actions that are plausible before verification begins.

\paragraph{\textbf{Geometric Verification.}}

The verification engine $\Psi$ evaluates each plan step against the scene geometry and the agent's embodiment constraints. For each action-object pair $(a^{(t)}, o^{(t)})$, \name{} retrieves the action family's ordered property checklist and evaluates each check against $(\mathcal{S}, \mathcal{A}, a^{(t)}, o^{(t)})$. Each check returns a boolean outcome and a diagnostic trace, such as an unreachable interaction zone, an out-of-range target surface, or insufficient clearance. The step verdict $v^{(t)} \in \{0,1\}$ is the conjunction of its check outcomes, and the activity label is the conjunction of the step verdicts. Step-level traces aggregate into the final diagnostic report \(\mathcal{R} = \Psi(\pi, \mathcal{S}, \mathcal{A})\).

\subsection{Reporting \& Downstream Applications}
\label{sec:method-application}
The diagnostic report records, for each planned step, whether the action is feasible, which geometric checks failed if any, and the concrete reason for failure (\eg, blocked access, unreachable target geometry, missing functional part, or insufficient clearance). This makes the verifier useful beyond a binary task label: the same report supports scene auditing, failure-mode analysis across agent profiles, and quantitative evaluation of VLM spatial reasoning, as shown in Fig.~\ref{fig:applications} and Sec.~\ref{experiments}. Complete report examples are provided in App.~\ref{supp:qualitative:reports}.

\paragraph{\textbf{Verifier-Feedback Post-Training.}}

Beyond downstream \textit{analysis} tasks, the diagnostic report $\mathcal{R}$ also provides an automated supervision signal for VLM post-training. Rather than relying on Supervised Fine-Tuning (SFT), which is prone to superficial alignment and catastrophic forgetting~\cite{chen2025retaining}, we formulate feasibility post-training as a reinforcement learning problem using Group Relative Policy Optimization (GRPO)~\cite{shao2024deepseekmath}, illustrated in Fig.~\ref{fig:applications}c. For a given atomic action step $t$, we condition a VLM $\Omega$ on a step context $y^{(t)}$ comprising the agent description, the action, and a scene render with the target object highlighted. The model samples a group of $G$ independent Chain-of-Thought (CoT)~\cite{CoT} reasoning paths, each concluding with a feasibility prediction $\hat{v}_i^{(t)} \sim \Omega(y^{(t)})$. Each completion receives a sparse reward $r_i$ driven by its agreement with the verifier label, $\mathbbm{1}[\hat{v}_i^{(t)} = v^{(t)}]$. The policy is then updated by maximizing the group-relative advantage $A_i = (r_i - \mu_r) / (\sigma_r + \epsilon)$, where $\mu_r$ and $\sigma_r$ are the mean and standard deviation of rewards within the group. By contrasting successful reasoning paths against failed ones on the exact same feasibility query, the VLM internalizes agent-conditioned physical constraint.

\section{Experiments}
\label{experiments}

We first validate the verifier's labels against human judgment and parameter perturbations (Sec.~\ref{sec:study}). We then audit how well synthetic scenes support everyday activities across agent profiles (Sec.~\ref{sec:exp-scenes}), benchmark how accurately leading VLMs predict feasibility (Sec.~\ref{sec:exp-benchmark}), and use verification reports as a post-training signal, with gains that transfer to human-labeled ScanNet++~\cite{yeshwanth2023scannet++} scenes (Sec.~\ref{sec:exp-grpo}).

\paragraph{\textbf{Implementation.}}

We deploy our verification pipeline on indoor 3D scenes from 3D-FRONT~\cite{3dfront}, which are populated with object assets from 3D-FUTURE~\cite{3dfuture}. Across all experiments, we use the \textit{living room} and \textit{dining room} subsets, yielding 1,132 unique scenes. We select this dataset for its broad adoption, scale, and heterogeneous residential scenes, providing a rigorous testbed for agent-conditioned functional verification. The semantic planner is Gemini 3 Flash Preview~\cite{gemini}, conditioned on high-quality scene renderings produced by Blender. Gemini also resolves the relevant interaction zones used for agent navigation. Following prior work~\cite{corsetti2025functionality}, functional object parts are first identified with Molmo~\cite{molmo} and then segmented using SAM~\cite{sam}. All geometric verification routines are implemented with libigl~\cite{libigl} and Trimesh~\cite{trimesh}. Comprehensive implementation details for the full \name{} stack are provided in App.~\ref{supp:implementation}. To facilitate follow-up work, we will release both the implementation and the pre-computed verification traces for different activities and agent profiles.

\subsection{Validating the Verification Interface}
\label{sec:study}

We validate the \name{} verifier through a human study over 100 complete execution traces spanning diverse scene-task-agent triplets.

\paragraph{\textbf{Human Agreement.}}
In the planning phase, our VLM-driven planner achieved a 97.0\% logical soundness rate, indicating reliable decomposition of high-level activities into executable sequences. During physical validation, the verification engine showed strong agreement with human judgment, reaching 91.0\% agreement on overall outcomes and 96.3\% agreement across 454 individual verification steps. The utility of the generated diagnostic reports was rated at an average of \(4.30\) out of \(5\), supporting our goal of fine-grained, interpretable failure analysis. Additional details on the study protocol are provided in App.~\ref{supp:dataset:study}.

\paragraph{\textbf{Robustness.}}
Agreement remains strong across the depth of the verification chain: while a low-depth actions such as \textit{SitOn} reaches 94.5\% agreement, a more complex action such as \textit{Open}, which requires navigation, part localization, and clearance checks, maintains 82.4\% agreement. Auditing upstream semantic-grounding failures in the same cascade shows that Gemini interaction-zone proposals account for 1.8\% of audited step errors, Molmo part-localization for 1.3\%, and SAM segmentation for 0\%. Observed label errors are therefore limited and diagnosable rather than opaque. The labels are also stable under parameter perturbations. For instance, across 3,268 reachability checks, varying the reach radius \(r_{\text{arm}}\) over a range of at least 21.9~cm shifts property pass rates by less than 3\%.

\subsection{Auditing Functional Failures in 3D Scenes}
\label{sec:exp-scenes}

\begin{figure}[t]
  \centering
  \begin{minipage}[c]{0.56\columnwidth}
    \centering
    \captionof{table}{Overall task \(\mathcal{T}\) and boolean properties \(\mathbb{P}\) success rates per agent profile on 3D-FRONT~\cite{3dfront}.}
    \label{tab:3df-results}
    \resizebox{\linewidth}{!}{%
      {\renewcommand{\arraystretch}{1.05}%
      \begin{tabular}{l c c c}
        \toprule
        \multirow{2}{*}{\textsc{\textbf{Metric}}} & \multicolumn{3}{c}{\textsc{\textbf{Agent Profile}}} \\
        \cmidrule(lr){2-4}
        & Adult & Child & Wheelchair \\
        \midrule
        Task Success & \cellcolor{red!25}59.0\% & \cellcolor{red!20}66.0\% & \cellcolor{red!38}42.5\% \\
        \texttt{isNavigableTo} & \cellcolor{red!9}84.7\% & \cellcolor{red!7}89.2\% & \cellcolor{red!16}73.0\% \\
        \texttt{isReachable} & \cellcolor{red!5}91.9\% & \cellcolor{red!4}93.7\% & \cellcolor{red!11}80.7\% \\
        \texttt{isInteractable} & \cellcolor{red!14}75.6\% & \cellcolor{red!16}73.4\% & \cellcolor{red!22}64.1\% \\
        \texttt{hasClearance} & \cellcolor{red!19}67.3\% & \cellcolor{red!20}66.3\% & \cellcolor{red!20}66.6\% \\
        \texttt{isVisible} & \cellcolor{red!1}98.7\% & \cellcolor{red!1}97.8\% & \cellcolor{red!1}97.7\% \\
        \bottomrule
      \end{tabular}%
      }
    }
  \end{minipage}\hfill
  \begin{minipage}[c]{0.42\columnwidth}
    \centering
    \includegraphics[width=\linewidth]{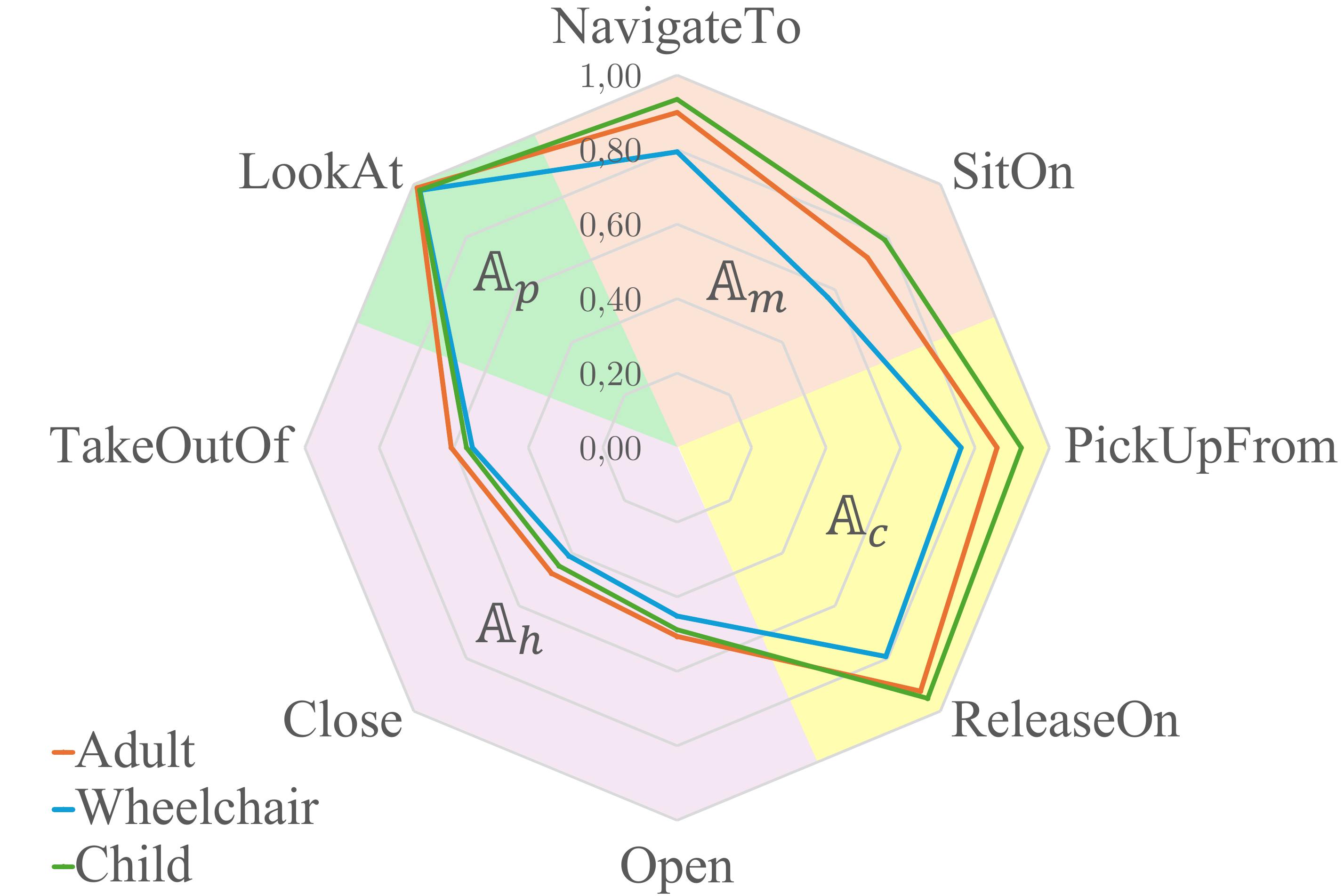}
    \captionof{figure}{Success rates per atomic action \(a\) and agent profile.}
    \label{fig:char_visual}
  \end{minipage}
\end{figure}

3D-FRONT layouts are professionally designed and serve as the standard realism reference for scene generation~\cite{atiss,debara,diffuscene,physcene}, making them visually and semantically plausible by construction. The functional failures reported by \name{} therefore arise despite this plausibility. To benchmark the readiness of modern 3D environments for embodied interaction, we apply \name{} across the 3D-FRONT scenes and evaluate each scene-task configuration under three agent profiles: Adult, Child, and Wheelchair user. This setup tests the agent-dependent nature of affordances (Fig.~\ref{fig:applications}a) and results in 3,396 scene-task-agent verifications. As shown in Tab.~\ref{tab:3df-results}, the dataset exhibits broad feasibility gaps across profiles, indicating that many scene arrangements do not reliably support everyday activities. 

\finding{Finding 1: visually plausible scenes frequently fail basic activities, and the failures are strongly agent-dependent.} Feasibility for wheelchair users is markedly lower than for the other agents, and individual boolean properties show that navigation constraints are a primary bottleneck for this profile, reflecting restrictive spatial layouts. The child profile even scores above the adult: its smaller clearance width eases navigation, the dominant bottleneck, and 3D-FRONT rarely places objects high enough to exceed a child's reach. This pattern is further illustrated in the per-action analysis in Fig.~\ref{fig:char_visual}, where success rates for wheelchair users are consistently bounded by those of the other agents across atomic actions. Conversely, while objects remain almost universally visible across embodiments, the weaker \textit{clearance} results suggest that synthetic scenes are often too cluttered to support robust interaction. Common layout patterns and interaction issues revealed by \name{} are visualized in Fig.~\ref{fig:checks}. Experimental details, including agent property definitions and task acquisition, are provided in App.~\ref{supp:dataset}. These dataset-level diagnoses motivate our next analysis: testing whether VLMs can make reliable feasibility judgments.

\subsection{Benchmarking VLM Feasibility Judgments}
\label{sec:exp-benchmark}

\begin{table*}[!ht]
\centering
\caption{\textbf{VLM Benchmark Results.} Each model is evaluated under two judgment interfaces: \textit{Direct} holistic task prediction and \textit{Decomposed} action-level prediction. MCC values below $0.1$ (\colorbox{red!15}{red}) indicate near-absent discrimination power and make the remaining metrics uninformative. The last rows are a controlled comparison: the same Qwen3-VL-4B backbone~\cite{qwen3}, before and after GRPO post-training with \name{} verifier feedback.}
\label{tab:main_results2}
\resizebox{\textwidth}{!}{%
\begin{tabular}{c ll c ccc ccc}
\toprule
 & \multicolumn{2}{c}{\multirow{2}{*}{\textsc{\textbf{Model Configuration}}}} & \multicolumn{1}{c}{\textsc{\textbf{Action}}} & \multicolumn{3}{c}{\textsc{\textbf{Task}}} & \multicolumn{3}{c}{\textsc{\textbf{Reliability}}} \\
\cmidrule(lr){4-4} \cmidrule(lr){5-7} \cmidrule(lr){8-10}
 & & & Accuracy $\uparrow$ & Accuracy $\uparrow$ & FP $\downarrow$ & MCC $\uparrow$ & InGap $\downarrow$ & HSI$^{*}$ & Cons. $\uparrow$ \\
\midrule
\multirow{6}{*}{\rotatebox{90}{\shortstack{Frontier}}}
 & \multirow{2}{*}{\texttt{Gemini-3-Flash-Preview}~\cite{gemini3flash}}
 & \textit{Direct} & --- & 62.1 & 35.7 & 0.144 & 16.5 & 79.2 & \multirow{2}{*}{58.5}\\
 & & \textit{Decomposed} & \textbf{77.1} & \textbf{69.9} & 12.0 & \textbf{0.390} & 7.5 & 94.7 & \\

 & \multirow{2}{*}{\texttt{Gemini-3.1-Pro-Preview}~\cite{gemini3-1pro}}
 & \textit{Direct} & --- & 61.7 & 22.4 & 0.182 & \textbf{4.5} & 89.3 & \multirow{2}{*}{53.3} \\
 & & \textit{Decomposed} & 72.1 & 61.6 & \textbf{6.0} & 0.321 & 6.9 & 112.7 &\\

 & \multirow{2}{*}{\texttt{Claude-Sonnet-4-6}~\cite{claude-sonnet-4-6}}
 & \textit{Direct} & --- & 62.8 & 21.9 & 0.208 & 12.8 & 97.9 & \multirow{2}{*}{\textbf{81.4}} \\
 & & \textit{Decomposed} & 73.1 & 62.3 & 20.6 & 0.201 & 19.4 & \textbf{100.4} & \\
\midrule
\multirow{4}{*}{\rotatebox{90}{\shortstack{8--12B}}}
 & \multirow{2}{*}{\texttt{Qwen3-VL-8B-Instruct}~\cite{qwen3}}
 & \textit{Direct} & --- & 61.5 & 32.9 & 0.130 & 17.6 & 82.7 & \multirow{2}{*}{69.5}\\
 & & \textit{Decomposed} & 67.5 & \textbf{62.3} & \textbf{20.1} & \textbf{0.204} & \textbf{6.6} & \textbf{99.2} & \\
 & \multirow{2}{*}{\texttt{Gemma3-12B-Instruct}~\cite{kamath2025gemma}}
 & \textit{Direct} & --- & 59.3 & 40.2 & \cellcolor{red!15} -0.055& 22.5 & 69.4 & \multirow{2}{*}{\textbf{93.7}}\\
 & & \textit{Decomposed} & \textbf{75.0} & 61.7 & 36.1 & 0.116 & 17.2 & 78.2 & \\
\midrule
\multirow{8}{*}{\rotatebox{90}{\shortstack{3--4B}}}
 & \multirow{2}{*}{\texttt{Ministral3-3B-Instruct}~\cite{liu2026ministral}}
 & \textit{Direct} & --- & 54.0 & 33.3 & \cellcolor{red!15} -0.049& 15.0 & 80.0 & \multirow{2}{*}{21.7}\\
 & & \textit{Decomposed} & 34.5 & 41.7 & \textbf{0.7} & \cellcolor{red!15} 0.069& 21.6 & 144.7 & \\
 & \multirow{2}{*}{\texttt{Gemma3-4B-Instruct}~\cite{kamath2025gemma}}
 & \textit{Direct} & --- & 59.8 & 40.2 & \cellcolor{red!15} 0.000& 22.5 & 72.3 & \multirow{2}{*}{\textbf{97.9}}\\
 & & \textit{Decomposed} & 74.8 & 60.1 & 38.8 & \cellcolor{red!15} 0.010& 18.9 & 71.9 & \\
\addlinespace
 & \multirow{2}{*}{\texttt{Qwen3-VL-4B-Instruct}~\cite{qwen3}}
 & \textit{Direct} & --- & 60.8 & 32.6 & 0.111 & 20.3 & 87.3 & \multirow{2}{*}{64.6}\\
 & & \textit{Decomposed} & 70.4 & 61.4 & 22.8 & 0.172 & 13.2 & 97.6 & \\
 & \multirow{2}{*}{\texttt{Qwen3-VL-4B} \textit{with} GRPO~\cite{shao2024deepseekmath} (\textbf{ours})}
 & \textit{Direct} & --- & 61.5 & 32.9 & 0.130 & 17.2 & 83.8 & \multirow{2}{*}{62.3} \\
 & & \textit{Decomposed} & \textbf{75.3} & \textbf{69.2} & 14.1 & \textbf{0.364} & \textbf{9.7} & \textbf{98.2} & \\
\bottomrule
\end{tabular}%
}
\captionsetup{justification=raggedright, singlelinecheck=false}
\caption*{\scriptsize $^{*}$For HSI, values closer to 100\% are better.}
\vspace{-1.3em}
\end{table*}

Modern VLMs exhibit strong semantic understanding of 3D scenes, but it remains unclear how reliably they judge whether a scene physically supports a specific action for a specific agent. To quantify this ability, we use \name{}'s labels as the evaluation reference.

\paragraph{\textbf{Setting.}}

We leverage \name{} to generate geometry-verified labels for interactions sampled from 3D-FRONT~\cite{3dfront} scenes and use this corpus to benchmark both frontier (\eg, Gemini~\cite{gemini}, Claude~\cite{claude-sonnet-4-6}) and open VLMs (\eg, Gemma~\cite{kamath2025gemma}, Qwen~\cite{qwen3}) as well as smaller variants. Each model is evaluated at two complementary levels of granularity:

\begin{itemize}
  \item \textbf{Task-Level (\textit{Direct})}: The VLM is provided the global context of the 3D scene alongside an agent's physical archetype and asked to predict the overall feasibility of a complex, multi-step task.
  \item \textbf{Action-Level (\textit{Decomposed})}: Mirroring \name{}'s pipeline, each task is broken down into atomic actions (\eg, \textit{open} the wardrobe, \textit{sit on} the couch). The VLM is provided a render with the target object highlighted and asked to predict the feasibility of that single interaction.
\end{itemize}
For both settings, model predictions are evaluated against \name{}'s grounded labels. Prompts and prediction examples are provided in App.~\ref{supp:benchmark}.

\paragraph{\textbf{Metrics.}} 
We first report Action and Task accuracies, measuring atomic-level spatial perception and full-activity evaluation, respectively. In the \textit{decomposed} setting, a task counts as feasible only if all constituent actions are predicted feasible. The False Positive rate (FP) isolates positive bias, that is, how frequently a model hallucinates impossible tasks as possible. Because our dataset contains natural class imbalances, we also report the Matthews Correlation Coefficient (MCC)~\cite{matthews1975comparison}, a robust unified measure of classification quality~\cite{chicco2020advantages}. We further introduce Consistency (Cons.), the percentage of tasks where a model's \textit{direct} holistic prediction logically matches its \textit{decomposed} conclusion. The Horizon Stability Index (HSI) measures resilience to compounding task complexity as the ratio of task accuracy on long-horizon tasks (3+ action steps) versus short-horizon tasks (1--2 steps), where 100\% indicates invariance to task length. Finally, the Inclusivity Gap (InGap) is the maximum difference in task accuracy across our agent profiles, where a lower gap indicates a model that better internalizes the physical constraints of specific embodiments.

\finding{Finding 2: direct VLM judgments systematically over-predict feasibility.} Tab.~\ref{tab:main_results2} reveals consistent cross-model trends. In the \textit{Direct} setting, all models exhibit substantial physical hallucination, with elevated FP rates and modest MCC despite reasonable task-level accuracy.

\finding{Finding 3: decomposition improves feasibility judgments but does not close the gap.} Decomposition systematically improves functional judgment for most models by reducing false positives and increasing classification quality, with especially large gains for Gemini-3-Flash-Preview. Still, models show distinct reliability profiles: Claude-Sonnet-4-6 is the most self-consistent across direct and decomposed predictions, while Gemini-3.1-Pro-Preview is the most conservative, with the lowest InGap and the lowest FP among models with meaningful discrimination power, but limited task-accuracy gains. Ministral-3's very low decomposed FP rate stems from rejecting almost all actions, which only its near-zero MCC reveals. Finally, performance differences across agent profiles persist even after decomposition, showing that embodiment-aware functional reasoning remains an open challenge.

\subsection{Improving VLM Feasibility Prediction}
\label{sec:exp-grpo}

As formalized in Sec.~\ref{sec:method-application} and illustrated in Fig.~\ref{fig:applications}c, we leverage \name{}'s verification reports not just for static evaluation, but as a training signal to improve the feasibility predictions of VLMs.

\paragraph{\textbf{Setting.}} 
We select the open-weight Qwen3-VL~\cite{qwen3} in its 4B variant as our base model. To build upon its general capabilities, we perform parameter-efficient fine-tuning using Low-Rank Adaptation (LoRA)~\cite{hu2022lora} on attention layers. The model is optimized using GRPO~\cite{shao2024deepseekmath}, driven by the correctness of its final action feasibility prediction against the verifier label. We use a group size \(G=12\) for \(2,800\) steps. Comprehensive details regarding the training hyperparameters, prompts and rewards are provided in App.~\ref{supp:grpo}.

\finding{Finding 4: verifier feedback closes much of the measured gap.} As reported in the last rows of Tab.~\ref{tab:main_results2}, GRPO post-training with \name{} improves agreement with \name{} labels across the main benchmark metrics. Most notably, the model's overall spatial classification quality (MCC) more than doubles in the \textit{decomposed} setting. The gain is primarily driven by reduced false positives: the False Positive (FP) rate drops from $22.8\%$ to $14.1\%$, suggesting better recognition of geometric bottlenecks (\eg, missing clearance or reach). The optimized model also reduces the Inclusivity Gap, indicating improved feasibility prediction across embodied profiles. Overall, with the same backbone, verifier-feedback training yields a lightweight model whose atomic-action and derived task-level accuracy are competitive with proprietary models on this benchmark.

\begin{figure}[htbp]
  \centering
  \includegraphics[width=\columnwidth]{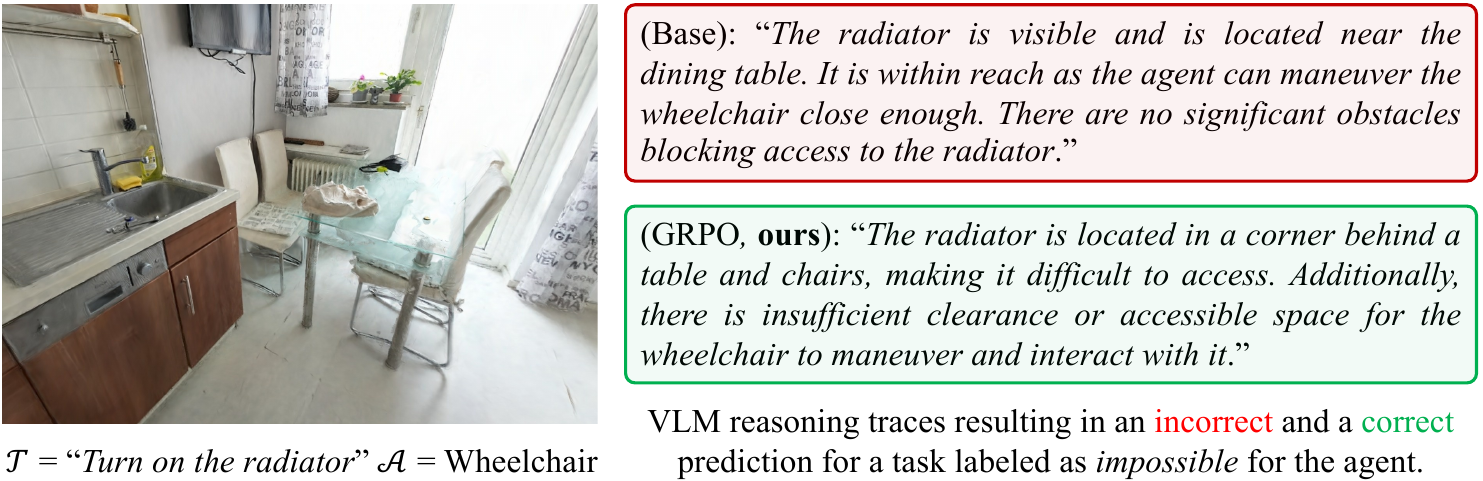}
  \caption{\textbf{Real-scene transfer on ScanNet++~\cite{yeshwanth2023scannet++}.} On a task human-labeled as impossible, the GRPO-optimized variant grounds its trace in the missing clearance and predicts correctly, while the base model overlooks the spatial constraints.}
  \label{fig:scannet-example}
\end{figure}

\begin{table}[htbp]
\centering
\caption{\textbf{Real-scene generalization on ScanNet++~\cite{yeshwanth2023scannet++}.} Qwen3 optimized with \name{} rewards improves over its base backbone across all metrics on human-labeled task feasibility.}
\label{tab:sim2real}
\resizebox{\columnwidth}{!}{%
\begin{tabular}{l c c c c}
\toprule
Model & Acc. $\uparrow$ & FP $\downarrow$ & MCC $\uparrow$ & InGap $\downarrow$ \\
\midrule
Gemini 3 Flash & 80.1\% & 12.8\% & 0.546 & 7.0\% \\
\midrule
Qwen3-VL-4B (Base) & 73.1\% & 23.5\% & 0.361 & 9.9\% \\
Qwen3-VL-4B (GRPO, \textbf{ours}) & \textbf{76.9\%} & \textbf{15.9\%} & \textbf{0.466} & \textbf{8.3\%} \\
\bottomrule
\end{tabular}%
}
\end{table}

\finding{Finding 5: the improvement transfers to human-labeled real scenes.} A natural question is whether these gains reflect genuine physical reasoning beyond the synthetic setting of \name{}-driven post-training. Using in-context images captured from high-quality 3D Gaussian Splatting reconstructions~\cite{li2025scenesplat} of 100 ScanNet++~\cite{yeshwanth2023scannet++} scenes, human annotators defined activities and feasibility labels across the same three agent profiles, yielding 502 real-scene labels. As shown in Tab.~\ref{tab:sim2real}, GRPO improves the base model across all metrics, closing much of the gap to a frontier model. Beyond stronger overall feasibility prediction, the optimized model remarkably corrects 39 physical hallucinations with only 1 regression on impossible tasks ($p < 10^{-10}$, McNemar test). The model is trained only on synthetic renders with \name{} labels, yet improves on images of real scenes with independent human labels, showing generalizable feasibility reasoning rather than overfitting to the verifier. Fig.~\ref{fig:scannet-example} shows a representative real-scene case where the optimized model recognizes a clearance bottleneck that the base model overlooks. These results support the central role of \name{} as a scalable source of grounded supervision rather than only an offline evaluator.

\section{Conclusion}

We introduced \name{}, a verification interface that tests whether an activity is physically executable in a 3D scene for a specific embodied agent. Deployed for scene auditing, VLM benchmarking, and post-training supervision, \name{} showed that (1) visually plausible scenes frequently fail basic activities, (2) modern VLMs systematically over-predict feasibility, and (3) verifier feedback improves a lightweight VLM, with remarkable transfer to human-labeled real scenes. Importantly, \name{} is not only an evaluator but a source of rich feasibility traces that can be used as expressive rewards for understanding targeted physical failures. Together, these results establish agent-conditioned executability as a measurable and improvable axis of 3D scene understanding.

\paragraph{\textbf{Limitations \& Future Work.}}
While easily extensible, \name{}'s interface still relies on a finite set of actions and checks, requiring 3D tool integration and scene preprocessing. The current agent profiles remain simplified proxies and may fall short for interactions that depend on fine-grained posture or two-handed manipulation. We also do not yet model dynamic interactions. Natural next steps include deploying \name{} as a closed-loop critic for scene synthesis, where verifier feedback can guide layouts toward functional validity.

\section*{Acknowledgements}

This work was supported by Dassault Systèmes SE. Léopold Maillard also acknowledges a Google Cloud Credit Grant for Stanford Students Projects. Francis Engelmann acknowledges the support from a SwissAI Grant for Small Projects and an Academic Grant from NVIDIA. Leonidas Guibas and Yang You acknowledge support from an ARL grant W911NF-21-2-0104 and a Vannevar Bush Faculty Fellowship. Yang You is also supported in part by the Outstanding Doctoral Graduates Development Scholarship of Shanghai Jiao Tong University. Maks Ovsjanikov acknowledges support from the ERC Consolidator Grant 101087347 (VEGA), as well as gifts from Ansys Inc. and Adobe Research. The authors also thank Or Litany and Ian Huang for insightful discussions related to the project.

{
    \small
    \bibliographystyle{ieeenat_fullname}
    \bibliography{main}
}

\addtocontents{toc}{\protect\setcounter{tocdepth}{3}}
\maketitlesupplementary

\begin{abstract}
  In this supplementary material, we provide comprehensive implementation details of \name{} (Section~\ref{supp:implementation}); dataset preprocessing, verification trace collection, and the perceptual study (Section~\ref{supp:dataset}); the VLM benchmark setup and prompt templates (Section~\ref{supp:benchmark}); the GRPO post-training formulation, rewards, and hyperparameters (Section~\ref{supp:grpo}); our real-scene evaluation protocol (Section~\ref{supp:scannet}); a discussion of potential societal impact (Section~\ref{supp:societal}); and extended qualitative examples and diagnostic reports (Section~\ref{supp:qualitative}).
\end{abstract}

\setcounter{secnumdepth}{3}
\setcounter{tocdepth}{3} 
\begingroup
    \hypersetup{linkcolor=black}
    \tableofcontents
\endgroup

\appendix

\section{Implementation}
\label{supp:implementation}

This section provides implementation details for \name{}. We first outline the modular software architecture in Section~\ref{supp:implementation:architecture}, then describe the VLM-based action planner and its multimodal inputs in Section~\ref{supp:implementation:planner}, detail the implementation of the geometric feasibility grounding in Section~\ref{supp:implementation:geometry}, and finally report its computational cost in Section~\ref{supp:implementation:cost}.

\subsection{System Overview}
\label{supp:implementation:architecture}

\begin{figure*}[t]
\includegraphics[width=\textwidth]{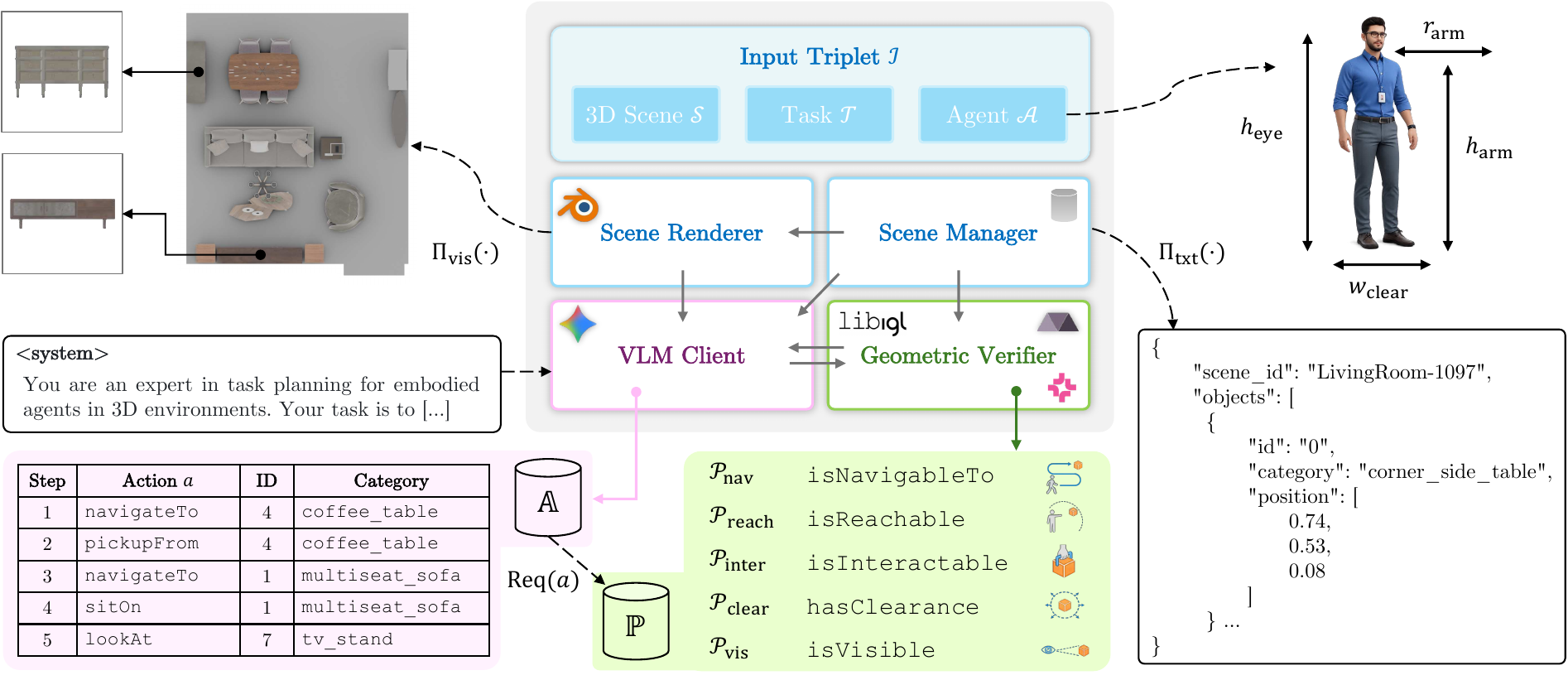}
\caption{Main components of the \name{} implementation. The system is organized around four interacting modules: a Scene Manager (which maintains structured scene state and object properties), a Scene Renderer (which produces visual observations for prompting), a VLM Client (which instantiates the Semantic Planner), and a Geometric Verifier (which executes grounded feasibility checks). The modules communicate via the shared input triplet \(\mathcal{I}=\langle \mathcal{S},\mathcal{T},\mathcal{A}\rangle\) and the action-to-property mapping \(\mathrm{Req}(a)\) that links each atomic action to the relevant geometric checks.}
\label{fig:architecture}
\end{figure*}

The component-level design of \name{} is illustrated in Figure~\ref{fig:architecture}. During the verification pipeline execution, the input triplet is routed across four core modules that we detail below. This separation keeps semantic reasoning, scene I/O, and geometric verification decoupled, making the pipeline easier to maintain and extend.

The \textbf{Scene Manager} maintains the structured scene \(\mathcal{S}\) state, including its geometry and navigable surfaces \(\mathcal{G}\) and object-level features \(\mathcal{O}\) (semantic labels, 3D bounding boxes, and detailed meshes). It exposes this as a human-readable context \(\Pi_{\text{txt}}(\cdot)\) for prompting the VLM planner, as shown in Figure~\ref{fig:architecture} (\textit{bottom-right}). The \textbf{Scene Renderer} creates the top-down renderings \(\Pi_{\text{vis}}(\cdot)\) fed to the planner as well as the views that are used to determine the interactive parts of individual objects, as detailed in Section~\ref{supp:implementation:geometry} and shown in Figure~\ref{fig:architecture} (\textit{top-left}). To implement rendering, we leverage Blender 4.2.12 LTS via its \texttt{bpy} interface. The \textbf{VLM Client} abstracts API calls to local or cloud-based multimodal reasoning models under a unified interface. It is notably used for the planner~\(\Phi\) and the resolution of the target object interaction zones for the agent's navigation, as detailed in Section~\ref{supp:implementation:geometry}. Finally, the \textbf{Geometric Verifier} executes low-level feasibility checks using primarily \texttt{trimesh} and \texttt{libigl} and returns step-wise outcomes under the agent's specific embodiment properties \(\mathcal{A}\).

\subsection{Semantic Planner}
\label{supp:implementation:planner}

As detailed in Sec.~\ref{sec:method-pipeline}, the planning phase relies on a Vision-Language Model to decompose high-level activities \(\mathcal{T}\) into a sequence of atomic actions. In all our experiments, the planner \(\Phi\) is \texttt{gemini-3-flash-preview}, accessed via Google's Vertex AI API. To minimize stochasticity in the planning logic, all VLM generation calls are performed with the temperature set to $0.0$. To condition the planner on both scene layout and embodiment constraints, we prompt the VLM with a multimodal context \(\mathcal{C}\), comprising: (1) the natural-language activity description \(\mathcal{T}\), (2) a structured textual serialization of the scene object inventory (detailing their ID, category and 3D position) together with agent constraints \(\mathcal{A}\), and (3) a top-down rendering generated by the renderer. Guided by the fixed library of available actions \(\mathbb{A}\), the VLM performs one-shot generation of an action sequence \(\pi\), returned as a strictly formatted JSON array of per-step \texttt{(object\_id, semantic\_action)} tuples. We display below the resulting prompt, tailored for the 3D-FRONT~\cite{3dfront} dataset:

\begin{tcolorbox}[breakable, boxrule=0pt, colframe=cvprblue, sharp corners, left=1mm, right=1mm, top=0.2mm, bottom=0.2mm, title=Semantic Planning Prompt]
\footnotesize
{
You are an expert in task planning for embodied agents in 3D environments. Your task is to decompose a high-level action into a sequence of primitive, executable steps, considering the agent's properties and the specific objects available in the scene.
\\\\
\textbf{--- RULES ---}

\textbf{1. Action Vocabulary}: You MUST ONLY use actions from the provided list of allowed actions: [\textcolor{cvprblue}{\{allowed\_actions\}}]

\textbf{2. Grounded Objects}: Each step in your plan MUST refer to a specific \texttt{object\_id} from the provided scene description S.

\textbf{3. Furniture-Centric Actions}:
\begin{itemize}[leftmargin=2em, label=-, nosep]
  \item The scene consists primarily of furniture (cabinets, tables, shelves) and lacks small portable objects.
  \item You MUST target the \textbf{furniture} itself for object interactions.
  \item Surfaces (e.g., tables, shelves, counters, etc.): Use \texttt{pickup\_from} and \texttt{release\_on}.
\end{itemize}

\begin{itemize}[leftmargin=4em, label=-, nosep]
  \item \textit{Example:} ``Get something on the table'' $\rightarrow$ \texttt{[(``navigate\_to'', ``table\_id''), (``pickup\_from'', ``table\_id'')]}
\end{itemize}

\begin{itemize}[leftmargin=2em, label=-, nosep]
  \item Containers (e.g., cabinets, fridges, drawers, etc.): Use \texttt{open}, \texttt{close}, \texttt{put\_in}, \texttt{take\_out\_of}.
\end{itemize}

\begin{itemize}[leftmargin=4em, label=-, nosep]
  \item \textit{Example:} ``Grab something in the cabinet'' $\rightarrow$ \texttt{[(``navigate\_to'', ``cupboard\_id''), (``take\_out\_of'', ``cupboard\_id'')]}
\end{itemize}

\textbf{4. Mapping Implicit Actions}:

\begin{itemize}[leftmargin=2em, label=-, nosep]
  \item Movement: Verbs like `stand up', `turn around', `walk to', `approach' are IMPLICIT. Use \texttt{navigate\_to} when the task only involves positioning the agent.
\end{itemize}

\begin{itemize}[leftmargin=4em, label=-, nosep]
  \item \textit{Example:} ``Go and stand by the window'' $\rightarrow$ \texttt{[(``navigate\_to'', ``window\_id'')]}
\end{itemize}

\textbf{5. Assume Physical Capability}: Do NOT reject a plan because you believe the agent's physical properties (e.g., height, arm length, wheelchair status) make the action difficult or impossible. Assume the agent is capable of the physical movement. Your job is only to determine the \textit{semantic} sequence of interactions. Physical feasibility will be verified by a separate physics engine.

\textbf{6. Output Format}: Your entire output MUST be a single valid JSON object. It must contain a key ``plan'', which is a list of dictionaries. Each dictionary must have two keys: ``object\_id'' and ``semantic\_action''.

\textbf{7. Handle Impossibility}: You should ONLY return a null plan if a task is semantically impossible e.g., because a critical object is missing from the scene (e.g., no `television' for a ``watch TV'' task). In this case, provide a ``reasoning'' string explaining which object is missing. Be creative and ONLY return a null plan as a last resort.
\\\\
\textbf{--- YOUR TASK ---}

INPUT:

Agent A:\\
\textcolor{cvprblue}{\{agent\_json\}}
\\\\
Action D: ``\textcolor{cvprblue}{\{action\_description\}}''
\\\\
Scene S:\\
\textcolor{cvprblue}{\{scene\_json\}}
\\\\
ASSISTANT RESPONSE:
}
\end{tcolorbox}

\subsection{Geometric Verification}
\label{supp:implementation:geometry}

At the core of \name{}, the grounding engine maps each atomic action in the semantic plan to an ordered sequence of boolean physical properties (as defined in Tab.~\ref{tab:interface_integrated}) and assesses them using specialized 3D tools.

\subsubsection{Grounding Procedure ($\Psi$)}
\label{supp:implementation:geometry:procedure}

Algorithm~\ref{alg:grounding} details this orchestration. For each step $(a^{(t)}, o^{(t)})$ in the plan $\pi$, the verifier retrieves the ordered property checklist $\mathrm{Req}(a^{(t)})$ associated with the atomic action and evaluates each property in sequence via $\text{Diagnose}(\mathcal{P}, y^{(t)})$, which returns both a boolean outcome $s$ and a diagnostic trace $d$ explaining the result. The step verdict $v^{(t)}$ is the conjunction of all property outcomes, and the plan-level diagnostic report $\mathcal{R}$ aggregates the per-step verdicts and traces.

\begin{algorithm}[t]
\footnotesize
\caption{Agent-Aware Geometric Grounding ($\Psi$)}
\label{alg:grounding}
\begin{algorithmic}[1]
\State \textbf{Input:} Plan $\pi$, Scene $\mathcal{S}$, Agent $\mathcal{A}$
\State \textbf{Output:} Diagnostic report $\mathcal{R}$
\State $\mathcal{R} \gets [\,]$ \Comment{Initialize verification report}
\For{$t = 1, \dots, |\pi|$} \Comment{Iterate over plan steps}
    \State $(a^{(t)}, o^{(t)}) \gets \pi[t]$ \Comment{Atomic action and target object}
    \State $v^{(t)} \gets 1$, \ $\epsilon^{(t)} \gets [\,]$ \Comment{Step verdict and trace}
    \State $y^{(t)} \gets (\mathcal{S}, \mathcal{A}, a^{(t)}, o^{(t)})$ \Comment{Property-check input}
    \For{each $\mathcal{P} \in \mathrm{Req}(a^{(t)})$} \Comment{Ordered checklist for $a^{(t)}$}
        \State $(s, d) \gets \text{Diagnose}(\mathcal{P}, y^{(t)})$ \Comment{Outcome $s$, diagnostic trace $d$}
        \State $v^{(t)} \gets v^{(t)} \wedge s$
        \State $\epsilon^{(t)}.\text{append}((\mathcal{P}, s, d))$
    \EndFor
    \State $\mathcal{R}.\text{append}((a^{(t)}, o^{(t)}, v^{(t)}, \epsilon^{(t)}))$
\EndFor
\State \textbf{return} $\mathcal{R}$
\end{algorithmic}
\end{algorithm}

We now detail the agent-aware implementation of $\text{Diagnose}(\mathcal{P}, \cdot)$ for each of the five property checks $\mathcal{P}$.

\subsubsection{Navigability (\texttt{is\_Navigable\_To})}

\begin{figure*}[t]
\includegraphics[width=\textwidth]{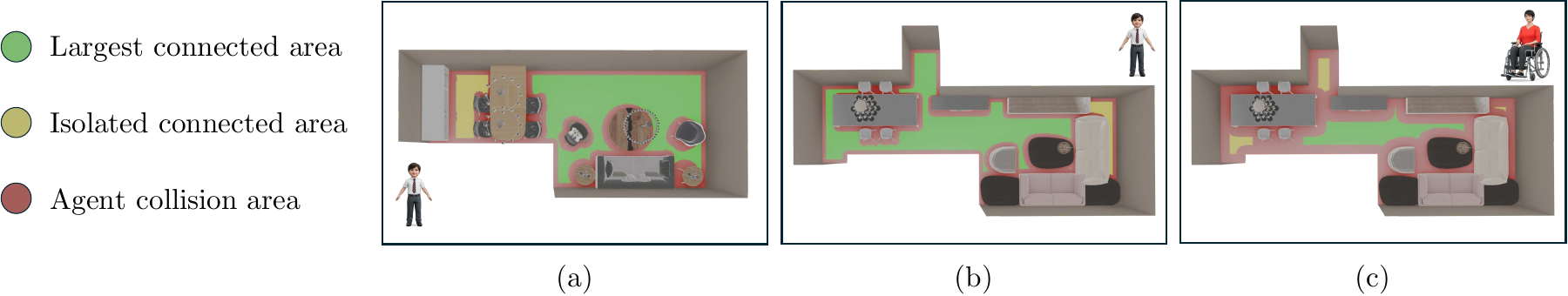}
\caption{Agent-specific navigation map examples. Our method (Algorithm~\ref{alg:nav_map}) identifies areas that are inaccessible to some or all agent profiles. During verification, the agent's position is initialized randomly within the largest connected area (\textit{green}).}
\label{fig:navmap}
\end{figure*}

At the core of the \name{} navigation system lies the agent-aware 2D navigation map $M$. To compute it, we orthographically project the scene's 3D floor geometry and static obstacles onto a 2D plane, producing a discrete occupancy grid. This grid is subsequently eroded by the agent's specific clearance width constraint $w_{\text{clear}}$ using morphological binary erosion. This map directly identifies paths and spatial limits, segmenting the scene into disjoint walkable regions. Prior to execution, the initial agent position is determined by sampling a representative point from the largest connected component of $M$. Algorithm~\ref{alg:nav_map} details this process and results are shown in Figure~\ref{fig:navmap}.

\begin{algorithm}[t]
\footnotesize
\caption{Agent-Aware Navigation Map Generation}
\label{alg:nav_map}
\begin{algorithmic}[1]
\State \textbf{Input:} Scene $\mathcal{S} = \langle \mathcal{O}, \mathcal{G} \rangle$, Agent $\mathcal{A}$, Resolution $R$
\State \textbf{Output:} 2D binary navigation map $M \in \{0,1\}^{R \times R}$
\State $scale, center \gets \text{ComputeBounds}(\mathcal{G}_{\text{floor}})$
\State $M \gets \text{Rasterize}(\mathcal{G}_{\text{floor}}, scale, center)$ \Comment{Initialize binary map with floor geometry}
\For{each object $o \in \mathcal{O}$} \Comment{Project and rasterize obstacles}
    \If{$o.\text{height} > \text{AGENT\_HEIGHT\_THRESHOLD}$}
        \State \textbf{continue} \Comment{Filter objects above the agent's height (\eg, ceiling lamps)}
    \EndIf
    \State $o_{\text{2D}} \gets \text{ProjectMeshToXZ}(o.\text{mesh})$ \Comment{Orthographic projection}
    \State $p_{\text{img}} \gets \text{SceneToImageCoords}(o_{\text{2D}}, scale, center)$
    \State $M[\text{DrawPolygon}(p_{\text{img}})] \gets 0$ \Comment{Mark as occupied}
\EndFor
\State $w_{\text{px}} \gets \text{MeterToPixels}(\mathcal{A}.w_{\text{clear}}, scale)$ \Comment{Determine agent clearance in pixels}
\State $selem \gets \text{Disk}(w_{\text{px}} / 2)$
\State $M \gets \text{BinaryErosion}(M, selem)$ \Comment{Erode to enforce clearance}
\State \textbf{return} $M$
\end{algorithmic}
\end{algorithm}

\paragraph{\textbf{Semantic Interaction Zones.}} 
A major challenge with \textit{naive} navigation is that simply verifying reachability to an object's centroid or bounding box is often insufficient. For instance, the back of a couch might be physically accessible but does not support the action \texttt{SitOn}. To address this, \name{} introduces a semantic resolution step. We first derive four enclosing interaction zones, geometrically modeled as trapezoids extending outward from the faces of the object's oriented bounding box (OBB) onto the floor plane. The renderer then generates an image of the object surrounded by these distinct, colored zones. We then prompt \texttt{gemini-3-flash-preview} with this augmented image to identify the functionally correct zone(s) (\eg, the front of the couch). The prompt template used for this resolution task is provided below:

\begin{tcolorbox}[breakable, boxrule=0pt, colframe=cvprblue, sharp corners, left=1mm, right=1mm, top=0.2mm, bottom=0.2mm, title=Interaction Zone Resolution Prompt]
\footnotesize
{

You are provided with a top-down orthographic view of an indoor scene. The grey area represents the navigable floor. The user wants to perform the action `\textcolor{cvprblue}{\{semantic\_action\}}' on the object `\textcolor{cvprblue}{\{target\_object.category\}}'. Look at the image provided. It shows the object with four colored zones (red, green, blue, yellow) around it. Which colored zone(s) represent the area where an agent must stand to perform this action? Select the zone(s) facing the functional front of the object (e.g., handles/doors) that is/are accessible from the grey floor.
\\\\
Please respond with a JSON object containing a single key 'selected\_zones' which is a list of color names.\\ For example: \{\texttt{'selected\_zones': ['red', 'blue']}\}
}
\end{tcolorbox}

\begin{figure*}[t]
\includegraphics[width=\textwidth]{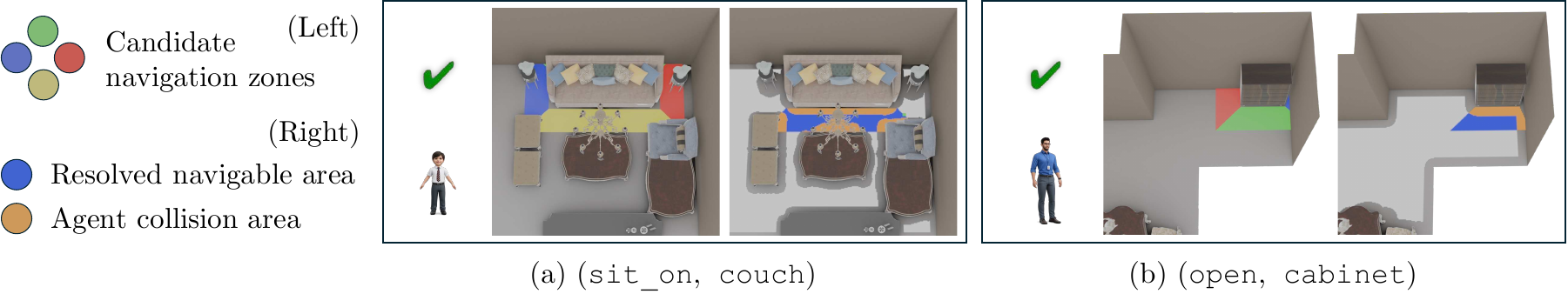}
\caption{Interaction navigation zones (left) and resolution results (right). From a scene rendering where the target object of the current plan step is surrounded by four colored interaction zones, a VLM is prompted to resolve the one(s) in which the agent should stand to perform the action. Connectivity to the resolved area is finally verified using the navigation map (Algorithm~\ref{alg:is_nav_to}).}
\label{fig:navigable}
\end{figure*}

\paragraph{\textbf{Connectivity Check.}} 
Once the correct interaction zone is resolved, the actual \texttt{is\_Navigable\_To} check evaluates if the agent's current position and the resolved zone share the same connected component on the navigation map $M$. If successful, the agent's position is updated to a random valid coordinate within the intersection of the interaction zone and the walkable area, simulating the completion of the navigation step. This routine is detailed in Algorithm~\ref{alg:is_nav_to} and showcased in Figure~\ref{fig:navigable}.

\begin{algorithm}[t]
\footnotesize
\caption{\texttt{is\_Navigable\_To} Verification}
\label{alg:is_nav_to}
\begin{algorithmic}[1]
\State \textbf{Input:} Navigation map $M$, Agent $\mathcal{A}$ with pose $p$, Target Zones $\mathcal{Z}$
\State \textbf{Output:} Boolean success, New Agent Position $p'$
\State $L \gets \text{ConnectedComponents}(M)$ \Comment{Label disjoint walkable regions}
\State $l_{\text{agent}} \gets L[\text{SceneToImageCoords}(p)]$ \Comment{Identify agent's current region}
\For{each zone $Z \in \mathcal{Z}$}
    \State $Z_{\text{mask}} \gets \text{DrawPolygon}(Z)$ \Comment{Rasterize target interaction zone}
    \State $I \gets (Z_{\text{mask}} == 1) \land (L == l_{\text{agent}})$ \Comment{Find intersection with agent's region}
    \If{$\sum I > 0$}
        \State $p_{\text{img\_new}} \gets \text{RandomChoice}(\text{IndicesOf}(I))$ \Comment{Sample new position}
        \State $p' \gets \text{ImageToSceneCoords}(p_{\text{img\_new}})$
        \State \textbf{return} \text{True}, $p'$
    \EndIf
\EndFor
\State \textbf{return} \text{False}, $p$ \Comment{No path to zone; retain original pose}
\end{algorithmic}
\end{algorithm}

\subsubsection{Reachability (\texttt{is\_Reachable})}

The reachability check determines if an agent, characterized by a shoulder height $h_{\text{arm}}$ and a maximum reach radius $r_{\text{arm}}$, can physically extend its arm to touch any part of a target object's mesh $\mathcal{M}_o$. \name{} implements this check by reframing the 3D proximity query as a distance calculation between the floor and a vertically translated (i.e. shifted) version of the target.

\paragraph{\textbf{Shifting Method.}}
Let $P \subset \mathbb{R}^3$ be the set of 3D points representing the agent's connected navigable floor area (derived from $M$ in Sec.~\ref{supp:implementation:geometry}). Mathematically, the task is to verify if there exists a point $p \in P$ such that the distance from the agent's shoulder $(p_x, p_y + h_{\text{arm}}, p_z)$ to the mesh $\mathcal{M}_o$ is within $r_{\text{arm}}$. To solve this efficiently for all $p \in P$ simultaneously, we shift the entire mesh $\mathcal{M}_o$ downwards by $h_{\text{arm}}$, creating a virtual mesh $\mathcal{M}_{\text{shifted}}$. The minimum distance between the static floor points $P$ and $\mathcal{M}_{\text{shifted}}$ then yields the required reach distance.

\begin{figure*}[t]
\includegraphics[width=\textwidth]{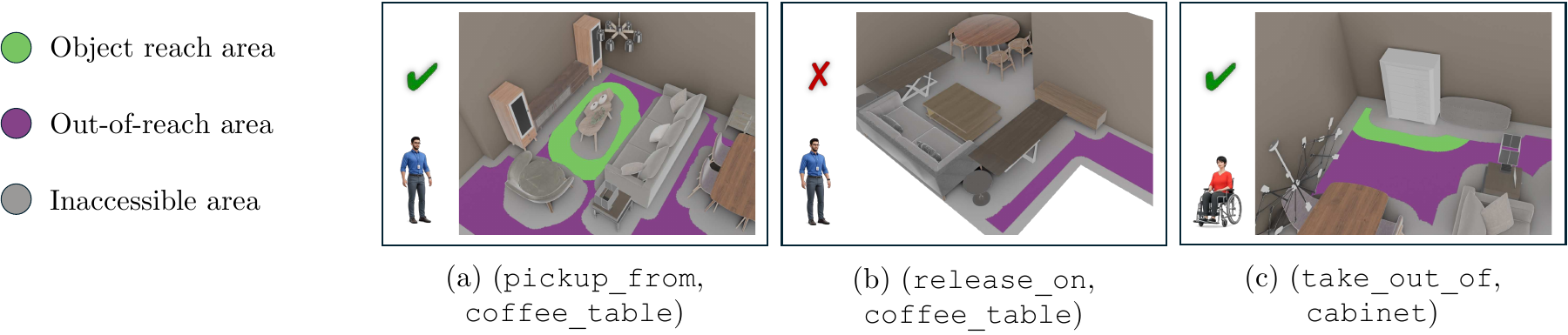}
\caption{Verification results for the \texttt{is\_Reachable} property (Algorithm~\ref{alg:is_reachable}). We can identify the precise floor area from which the agent can reach the target object.}
\label{fig:reach}
\end{figure*}

\paragraph{\textbf{Verification Logic.}}
The system first performs this check at the agent's standard standing height. If unsuccessful, it automatically attempts a secondary check at a minimum crouch height $h_{\text{crouch}}$ (modeled as a fraction of total height), simulating the agent leaning or kneeling to reach lower or further. The specific crouch mobility factors are uniquely defined for each embodied profile (\eg, Adult vs. Wheelchair user), as detailed in Section~\ref{supp:dataset:agents}. The complete procedure is detailed in Algorithm~\ref{alg:is_reachable} and results are shown in Figure~\ref{fig:reach}.

\begin{algorithm}[t]
\footnotesize
\caption{\texttt{is\_Reachable} Verification}
\label{alg:is_reachable}
\begin{algorithmic}[1]
\State \textbf{Input:} Target geometry $G$ (Mesh or Point Cloud), Navigable points $P$, Agent $\mathcal{A}$
\State \textbf{Output:} Boolean success
\State $h_{\text{test}} \in \{ \mathcal{A}.h_{\text{arm}}, \mathcal{A}.h_{\text{crouch}} \}$ \Comment{Heights to test}
\For{each $h \in h_{\text{test}}$}
    \State $G_{\text{shifted}} \gets \{ v - (0, h, 0) \mid v \in G \}$ \Comment{Translate geometry vertically}
    \State $d_{\text{min}} \gets \text{MinDistance}(P, G_{\text{shifted}})$ \Comment{Fast distance query}
    \If{$d_{\text{min}} \le \mathcal{A}.r_{\text{arm}}$}
        \State \textbf{return} \text{True}
    \EndIf
\EndFor
\State \textbf{return} \text{False}
\end{algorithmic}
\end{algorithm}

\subsubsection{Interactability (\texttt{is\_Interactable})}
To verify if an agent can manipulate specific functional parts of an object (\eg, grasping the handle of a wardrobe rather than its main body), we implement a multi-view 3D functional segmentation pipeline. This process ensures that feasibility is grounded in the geometric reality of the interaction point.

\begin{figure*}[t]
\includegraphics[width=\textwidth]{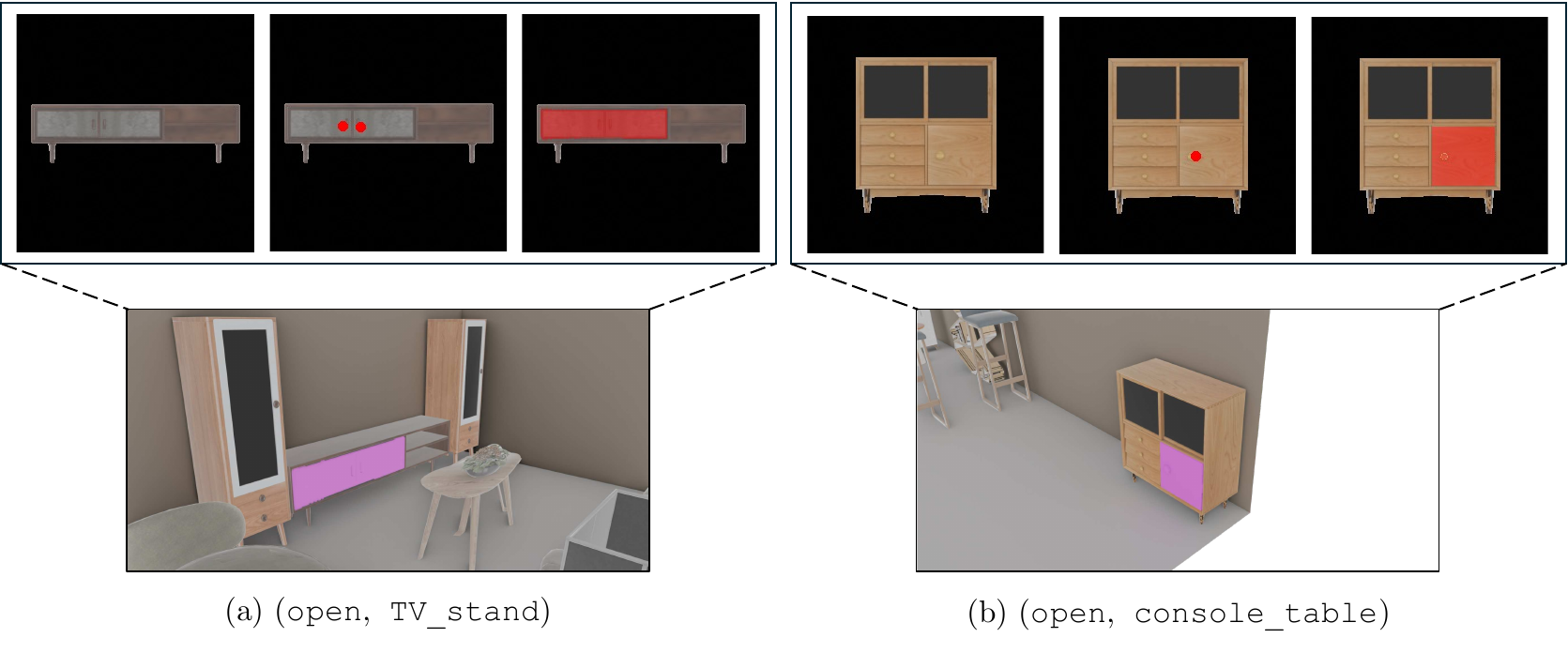}
\caption{Object functional parts identification and 3D deprojection. On the top row, from left to right: object view used to prompt Molmo~\cite{molmo}, predicted interaction point(s) for the intended action, segmentation mask derived by SAM~\cite{sam}. On the bottom row: in-context image of the object featuring the deprojected interactable volume (pink).}
\label{fig:interact}
\end{figure*}

\paragraph{\textbf{Multi-View Segmentation.}}
First, the rendering backend captures a set of orthographic views $\{V_i\}$ (front, back, left, right) of the target object $o^*$. To focus on relevant geometry, we select only the views whose camera vectors most closely align with the resolved semantic interaction zones. For each selected view, we use the Molmo~\cite{molmo} VLM in its 7B version\footnote{\url{https://huggingface.co/allenai/Molmo-7B-D-0924}} to identify the interaction points (if any) on the functional part of the object given the atomic action description $a$. The Molmo prompt template is provided below:

\begin{tcolorbox}[boxrule=0pt, colframe=cvprblue, sharp corners, left=1mm, right=1mm, top=0.2mm, bottom=0.2mm, title=Functional Part Identification Prompt]
\footnotesize
{
if it exists, point to the functional part of the \textcolor{cvprblue}{\{target\_object.category\}} to perform the action ``\textcolor{cvprblue}{\{action\_description\}}''
}
\end{tcolorbox}
\noindent These points serve as sparse prompts for a pretrained\footnote{\url{https://huggingface.co/facebook/sam-vit-huge}} Segment Anything Model~\cite{sam} (SAM), which extracts a precise 2D mask $K_i$. We show examples of identified interactive points and resulting segmentation masks in Figure~\ref{fig:interact}.

\paragraph{\textbf{3D Deprojection.}}
Using the corresponding depth map $D_i$ and the camera's inverse projection matrix $\mathbf{W}_i^{-1}$, we deproject each masked pixel $(u,v) \in K_i$ into world-space coordinates. The union of these points across all processed views forms a unified 3D point cloud representing the interactable volume $V_{\text{inter}}$.

\paragraph{\textbf{Proximity Verification.}}
Finally, we evaluate if $V_{\text{inter}}$ is within reach of the agent. This is achieved by passing the computed volume directly to the reachability verification routine detailed in Algorithm~\ref{alg:is_reachable}. The complete procedure is detailed in Algorithm~\ref{alg:is_interactable}.

\begin{algorithm}[t]
\footnotesize
\caption{\texttt{is\_Interactable} Verification}
\label{alg:is_interactable}
\begin{algorithmic}[1]
\State \textbf{Input:} Target object $o^*$, Navigable points $P$, Atomic action $a$, Agent $\mathcal{A}$
\State \textbf{Output:} Boolean success
\State $\{V_i, D_i, \mathbf{W}_i\} \gets \text{RenderOrthographicViews}(o^*)$
\State $V_{\text{inter}} \gets \emptyset$
\For{each relevant view $i$}
    \State $p_{\text{pts}} \gets \text{MolmoPoint}(V_i, a)$ \Comment{Identify functional part points}
    \State $K_i \gets \text{SAM}(V_i, p_{\text{pts}})$ \Comment{Extract 2D mask}
    \State $V_{\text{inter}} \gets V_{\text{inter}} \cup \text{Deproject}(K_i, D_i, \mathbf{W}_i^{-1})$ \Comment{Accumulate 3D volume}
\EndFor
\State \textbf{return} \text{is\_Reachable}($V_{\text{inter}}, P, \mathcal{A}$)
\end{algorithmic}
\end{algorithm}

\subsubsection{Clearance (\texttt{has\_Clearance})}
Certain actions, particularly those involving articulated containers (\eg, \texttt{Open}, \texttt{PutIn}), require sufficient kinematic space in front of the object to allow for both object articulation (door/drawer swing) and the agent's presence. \name{} verifies this via a volumetric collision check using a \textit{Clearance Box} method.

\paragraph{\textbf{Interaction Volume Generation.}}
For each resolved interaction zone, we define a 3D interaction volume $\mathcal{V}_{\text{clear}}$ modeled as a rectangular box. This box is perfectly aligned with the face of the object's oriented bounding box (OBB) corresponding to the interaction zone and extends outward onto the floor. The dimensions of $\mathcal{V}_{\text{clear}}$ are determined by the object's scale: its width and height match the target object's face, while its depth is set to the minimum of the object's horizontal extents, representing a conservative estimate of the required articulation space.

\begin{figure*}[t]
\includegraphics[width=\textwidth]{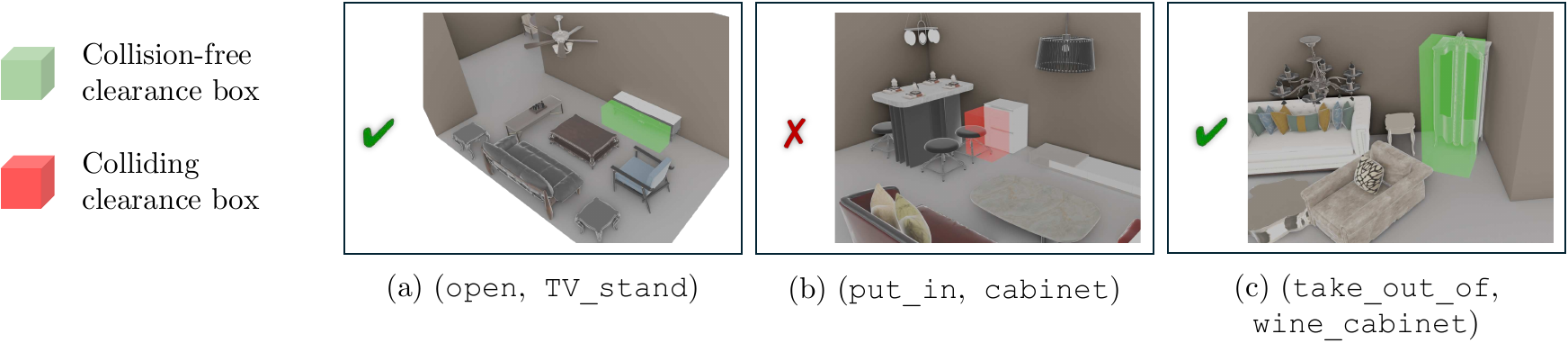}
\caption{Verification results for the \texttt{has\_Clearance} property. To simulate articulation and manipulation, objects coupled with an atomic action from the \textit{Handling} family $\mathbb{A}_\text{h}$ require a collision-free clearance volume facing their interactive side.}
\label{fig:clearance}
\end{figure*}

\paragraph{\textbf{Collision Verification.}}
To determine if the interaction space is obstructed, we utilize the collision manager from \texttt{trimesh} to account for the OBBs of all other static objects in the scene, as well as the 3D meshes of the room architecture (walls and ceiling). We then perform a discrete intersection test between $\mathcal{V}_{\text{clear}}$ and the aggregated scene geometry. The check succeeds if at least one resolved interaction zone provides a collision-free interaction volume that is entirely contained within the room's boundaries. This procedure is summarized in Algorithm~\ref{alg:has_clearance} and property verification results are shown in Figure~\ref{fig:clearance}.

\begin{algorithm}[t]
\footnotesize
\caption{\texttt{has\_Clearance} Verification}
\label{alg:has_clearance}
\begin{algorithmic}[1]
\State \textbf{Input:} Target object $o^*$, Interaction zones $\mathcal{Z}$, Scene obstacles $\mathcal{O} \setminus \{o^*\}$, Scene geometry $\mathcal{G}$
\State \textbf{Output:} Boolean success
\State $\mathcal{C} \gets \text{BuildCollisionEngine}(\mathcal{O} \setminus \{o^*\}, \mathcal{G}_{\text{walls}})$ \Comment{Aggregate scene geometry}
\For{each zone $Z \in \mathcal{Z}$}
    \State $\mathcal{V}_{\text{clear}} \gets \text{ComputeClearanceBox}(o^*, Z)$ \Comment{Align box with OBB face}
    \State $\text{is\_collision} \gets \text{CheckCollision}(\mathcal{C}, \mathcal{V}_{\text{clear}})$ \Comment{Check overlap with obstacles}
    \State $\text{is\_inside} \gets \text{ContainsPoint}(\mathcal{G}_{\text{floor}}, \text{Center}(\mathcal{V}_{\text{clear}}))$ \Comment{Box is within the room}
    \If{$\neg \text{is\_collision} \land \text{is\_inside}$}
        \State \textbf{return} \text{True}
    \EndIf
\EndFor
\State \textbf{return} \text{False}
\end{algorithmic}
\end{algorithm}

\subsubsection{Visibility (\texttt{is\_Visible})}
The visibility check determines if a target object is in the agent's line of sight, accounting for potential occlusions by other furniture or the room's architectural structure. \name{} employs a multi-ray casting approach to ensure robustness against partial occlusions.

\paragraph{\textbf{Posture-Aware Eye Placement.}}
Visibility is computed from the agent's eye position $e$, which is modeled using: $e_y = h_{\text{shoulder}} + \delta_{\text{eye}}$, where $\delta_{\text{eye}}$ is a fixed offset from the shoulder to the eyes. To account for the agent's dynamic state, $h_{\text{shoulder}}$ is adjusted based on the current posture: for standing agents, it matches the base shoulder height, while for sitting agents (excluding those in wheelchairs), it is reduced to 70\% of the standing height. For agents in wheelchairs, the shoulder height remains constant at a pre-defined sitting level. To simulate natural head movement and avoid brittle failures from single-pixel obstructions, we sample a set of $N_{\text{eye}}$ perturbed eye positions $\{e_j\}$ within a small radius around the theoretical center.

\paragraph{\textbf{Multi-Ray Sampling.}}
Rather than testing a single point, we cast rays from each eye position $e_j$ to a set of $N_{\text{target}}$ keypoints on the target object $o^*$. These keypoints include the object's geometric centroid and the eight vertices of its oriented bounding box (OBB). A ray is considered clear if it does not intersect any mesh in the occlusion set $\mathcal{M}_{\text{occ}} = \{ \mathcal{M}_{o} \mid o \in \mathcal{O} \setminus \{o^*\} \} \cup \mathcal{G}$ before reaching the target point.

\begin{figure*}[t]
\includegraphics[width=\textwidth]{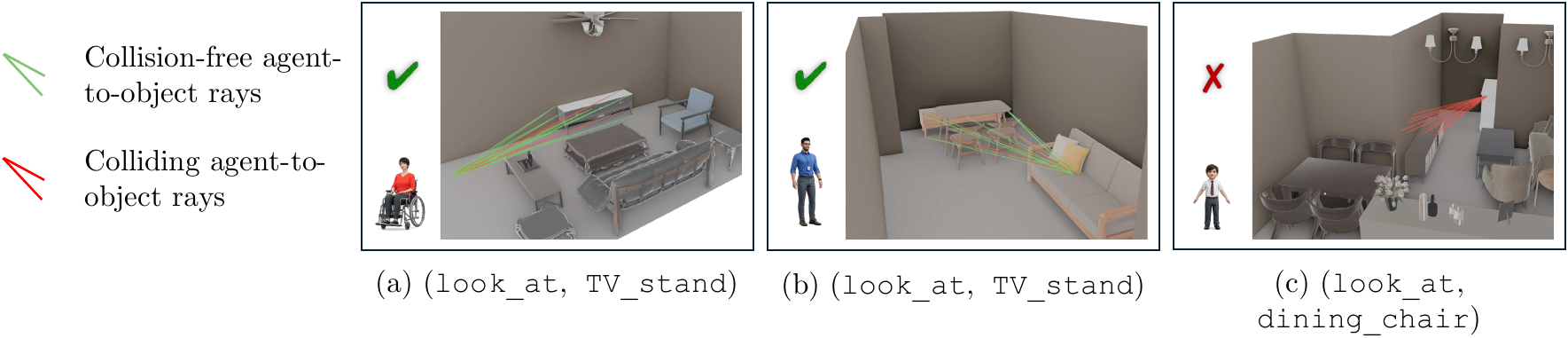}
\caption{Verification results for the \texttt{is\_Visible} property. While full object visibility is often ensured (a), the verification remains permissive and also accepts partial visibility (b). In rarer cases, visibility may still fail, typically due to architectural occlusions such as walls (c).}
\label{fig:visible}
\end{figure*}

\paragraph{\textbf{Visibility Criteria.}}
We calculate the visibility ratio $v = N_{\text{clear}} / (N_{\text{eye}} \cdot N_{\text{target}})$, where $N_{\text{clear}}$ is the number of unobstructed rays. The object is deemed visible if $v$ exceeds a threshold $\tau = 0.15$, ensuring that a significant portion of the object is perceivable. This process is detailed in Algorithm~\ref{alg:is_visible} and verification results are presented in Figure~\ref{fig:visible}.

\begin{algorithm}[t]
\footnotesize
\caption{\texttt{is\_Visible} Verification}
\label{alg:is_visible}
\begin{algorithmic}[1]
\State \textbf{Input:} Target object $o^*$, Agent $\mathcal{A}$ with pose $p$, Occlusion meshes $\mathcal{M}_{\text{occ}}$
\State \textbf{Output:} Boolean success
\State $\{e_j\} \gets \text{GetEyePositions}(\mathcal{A})$ \Comment{Sample perturbed eye points}
\State $\{t_k\} \gets \text{GetTargetPoints}(o^*)$ \Comment{Centroid and OBB corners}
\State $N_{\text{clear}} \gets 0$
\For{each eye position $e_j$}
    \For{each target point $t_k$}
        \State $\mathbf{r} \gets t_k - e_j$ \Comment{Ray vector}
        \If{$\neg \text{Intersects}(\mathcal{M}_{\text{occ}}, e_j, \mathbf{r}, \text{length}=\|\mathbf{r}\|)$}
            \State $N_{\text{clear}} \gets N_{\text{clear}} + 1$
        \EndIf
    \EndFor
\EndFor
\State $v \gets N_{\text{clear}} / (|\{e_j\}| \cdot |\{t_k\}|)$
\State \textbf{return} $v > \tau$
\end{algorithmic}
\end{algorithm}

\subsection{Computational Cost}
\label{supp:implementation:cost}

Averaged over the $3{,}396$ verification traces used in Sec.~\ref{sec:exp-scenes} and generated as described in the following Section~\ref{supp:dataset}, verifying a single scene-agent-task triplet takes a median of 28.4s (mean 36.5s), including scene rendering, geometric checks, and all VLM calls, for a plan averaging 4.6 steps, with 88.5\% of traces completing in under one minute. Runtime is consistent across the three agent profiles (within 1.2s of each other). Semantic planning and interaction-zone resolution jointly consume a median of 5{,}709 input and 237 output tokens per trace via \texttt{gemini-3-flash-preview}. At current pricing for this model (\$0.50 / \$3.00 per 1M input/output tokens), this corresponds to approximately \$0.36 per 100 verification traces (mean: \$0.40). This cost is naturally dependent on the provider and model, and querying a hosted API is not a strict requirement, as \name{} also supports locally-hosted models.

\section{Dataset Acquisition}
\label{supp:dataset}

This section details the empirical setup used to evaluate the 3D-FRONT~\cite{3dfront} scenes and the obtained diagnostic reports used subsequently to benchmark and improve VLMs. It covers the processing of the scene data, the parameterization of the embodied agents, the generation of complex tasks, and the protocol for the human perceptual study.

\subsection{3D-FRONT Dataset Processing}
\label{supp:dataset:3dfront}
We utilize the 3D-FRONT~\cite{3dfront} dataset, a large-scale repository of synthetic indoor environments largely adopted by the scene synthesis community. For our evaluation, we adopt the preprocessed\footnote{\url{https://github.com/PhyScene/PhyScene}} version of the dataset provided by PhyScene~\cite{physcene}, which ensures consistent scene organization and valid initial layouts. We specifically focus on the \textit{living room} and \textit{dining room} subsets, yielding a total of 1,132 unique residential scenes. Each environment is populated with high-quality object assets from 3D-FUTURE~\cite{3dfuture}. To ensure geometric fidelity during verification, raw meshes are programmatically loaded, centered, and rescaled to match the exact physical bounding box dimensions provided by the scene layout annotations. The object meshes are then transformed into the global world coordinates using the provided spatial poses. Similarly, the room's floor plan is constructed dynamically from layout vertices and faces to provide a closed, walkable ground plane. By leveraging these detailed, spatially aligned 3D structures, \name{} is able to perform the precise geometric and physical verification checks detailed in Section~\ref{supp:implementation:geometry}.

\subsection{Agent Profiles Definition}
\label{supp:dataset:agents}

A core principle of \name{} is that functional affordance is relational: an environment that is accessible to one user may not be to another. We encode these embodiment constraints using a simplified user model parameterized by continuous variables. For our 3D-FRONT evaluation and VLM experiments, we instantiate three distinct agent profiles: an \textbf{Adult}, a \textbf{Child}, and a \textbf{Wheelchair User}. 

The \textbf{Adult} profile represents a bipedal human user parameterized with average adult reach and mobility dimensions. The \textbf{Child} profile is configured to approximate the physical dimensions of a ten-year-old, featuring a reduced standing height, lower visual vantage point, and shorter arm reach. The \textbf{Wheelchair User} profile represents an adult operating a wheelchair: this configuration requires an increased navigation clearance (\(w_{\text{clear}}\)) of \qty{0.65}{\meter}, utilizes a static shoulder height reduction (factor of 0.70) to reflect a seated posture, and accounts for limited trunk mobility through a reduced crouch factor. We emphasize that these profiles are intended as representative instances of diverse embodiments used to evaluate the sensitivity of the verification procedure; they are not designed as exhaustive archetypes or mean representations of any specific group. The exact physical parameters used in our verification engine are summarized in Table~\ref{tab:agent_profiles}.

\begin{table*}[t]
\centering
\caption{Explicit physical parameters configuring the three embodied agent profiles used for geometric verification in \name{}. All distance measurements are in meters.}
\label{tab:agent_profiles}
\resizebox{\textwidth}{!}{%
\begin{tabular}{l c c c c c c c c c c}
\toprule
\multicolumn{1}{c}{\multirow{2}{*}{\textsc{\textbf{Profile}}}} & & \multicolumn{2}{c}{\textbf{Locomotion}} & & \multicolumn{3}{c}{\textbf{Anatomical Stack}} & & \textbf{Mobility} & \textbf{Manipulation} \\
\cmidrule{3-4} \cmidrule{6-8} \cmidrule{10-11}
& & \thead{Type\\($m_{\text{loco}}$)} & \thead{Clearance Width\\($w_{\text{clear}}$)} & & \thead{Standing Shoulder\\Height} & \thead{Shoulder-to-Eye\\Offset ($\delta_{\text{eye}}$)} & \thead{Eye-to-Top\\Offset} & & \thead{Crouch/Lean\\Factor} & \thead{Reach Radius\\($r_{\text{arm}}$)} \\
\midrule
\textbf{Adult}           & & Bipedal & 0.40 & & 1.45 & +0.20 & +0.10 & & 0.40 & 0.70 \\
\textbf{Child}           & & Bipedal & 0.30 & & 0.85 & +0.15 & +0.10 & & 0.50 & 0.40 \\
\textbf{Wheelchair User} & & Wheeled & 0.65 & & 1.45$^\dagger$ & +0.20 & +0.10 & & 0.10 & 0.70 \\
\bottomrule
\addlinespace[3pt]
\multicolumn{11}{l}{\footnotesize $^\dagger$For the Wheelchair User, the effective shoulder height is statically reduced by a posture scale of 0.70 to simulate a seated position.}
\end{tabular}%
}
\end{table*}

\subsection{Task Generation \& Acquisition}
\label{supp:dataset:tasks}

The 3D-FRONT dataset provides rich spatial and semantic layouts, but it does not natively contain activity or task annotations (\(\mathcal{T}\)). To evaluate the functionality of these scenes, we must synthesize realistic, multi-step activities in natural language. To do so we leverage a LLM (\texttt{gemini-2.5-flash}) to automatically generate context-aware tasks.

For each scene, the LLM is provided with a serialized JSON list of available objects (categories and IDs) and instructed to imagine a logical sequence of actions an embodied user might perform in that specific room. The prompt strictly enforces that the LLM must only refer to objects actually present in the provided list, preventing it from hallucinating non-existent items (\eg, asking the agent to ``read a book'' when no book object exists). The exact prompt template used for this generation step is provided below:

\begin{tcolorbox}[breakable, boxrule=0pt, colframe=cvprblue, sharp corners, left=1mm, right=1mm, top=0.2mm, bottom=0.2mm, title=Task Generation System Prompt]
\footnotesize
{
You are an expert at generating realistic, human-centric activities for 3D environments. Your goal is to create a single, natural language task description that an embodied agent (like a human or robot) could perform in the given scene.
\\\\
\textbf{Instructions}
\begin{enumerate}[leftmargin=*, topsep=2pt, partopsep=2pt]
  \item \textbf{Analyze the Scene:} You will be provided with a JSON list of objects present in the scene. Pay attention to their categories (e.g., ``chair'', ``table'', ``cabinet'') and IDs.
  \item \textbf{Create a Narrative:} Imagine a logical sequence of actions someone would do in this room.
  \item \textbf{Formulate the Task:} Write a concise, single-sentence instruction describing this activity.
  \item \textbf{Constraints:}
\end{enumerate}
\begin{itemize}[leftmargin=2em, label=*, nosep]
  \item \textbf{Strict Object Adherence:} You MUST ONLY refer to objects that are explicitly listed in the JSON. Do not invent objects (e.g., if there is no ``book'', do not say ``read a book'').
  \item \textbf{Referencing Objects:} When referring to an object, you do NOT need to use its exact ID (like \texttt{chair\_123}). Instead, use natural language (e.g., ``the chair'', ``the living room table''). If there are multiple similar objects, use spatial descriptors if possible (e.g., ``the chair near the table''), but generic references are acceptable.
  \item \textbf{Multi-Step:} The task should ideally involve 3-4 distinct actions (e.g., ``Navigate to X, then interact with Y'').
  \item \textbf{Agent Agnostic:} Do not assume specific agent capabilities (like ``flying'') unless it's a standard human action (walking, sitting, reaching).
\end{itemize}
\textbf{Input Format}\\
The input will be a JSON string representing the scene objects.

\textbf{Output Format}\\
Return \textbf{only} the task description string. Do not include any introductory text or markdown formatting.
}
\end{tcolorbox}

As stated in Sec.~\ref{sec:exp-scenes}, applying \name{} to the resulting scene-task samples for each of the three agent profiles produces the detailed verification traces and diagnostic reports used for our subsequent VLM auditing and post-training experiments. Examples of these reports are provided in Section~\ref{supp:qualitative:reports}. We will make the resulting dataset available to the research community to support further work in agent-aware 3D scene understanding.

\subsection{Perceptual Study Protocol}
\label{supp:dataset:study}
While \name{} verifications are ultimately grounded in strict geometric assessments, \name{} relies on a hybrid pipeline where high-level plans and semantic interaction zones are resolved by pre-trained models. Therefore, we conducted a manual perceptual study to ensure that this integration of semantic reasoning and rigid physical checks yields outcomes that appropriately align with human common sense and practical intuition. We sampled 100 complete execution traces across diverse scene-agent-task triplets from the evaluation dataset. 

\begin{figure*}[t]
    \centering
    
    \begin{subfigure}{\linewidth}
        \centering
        \includegraphics[width=0.81\linewidth]{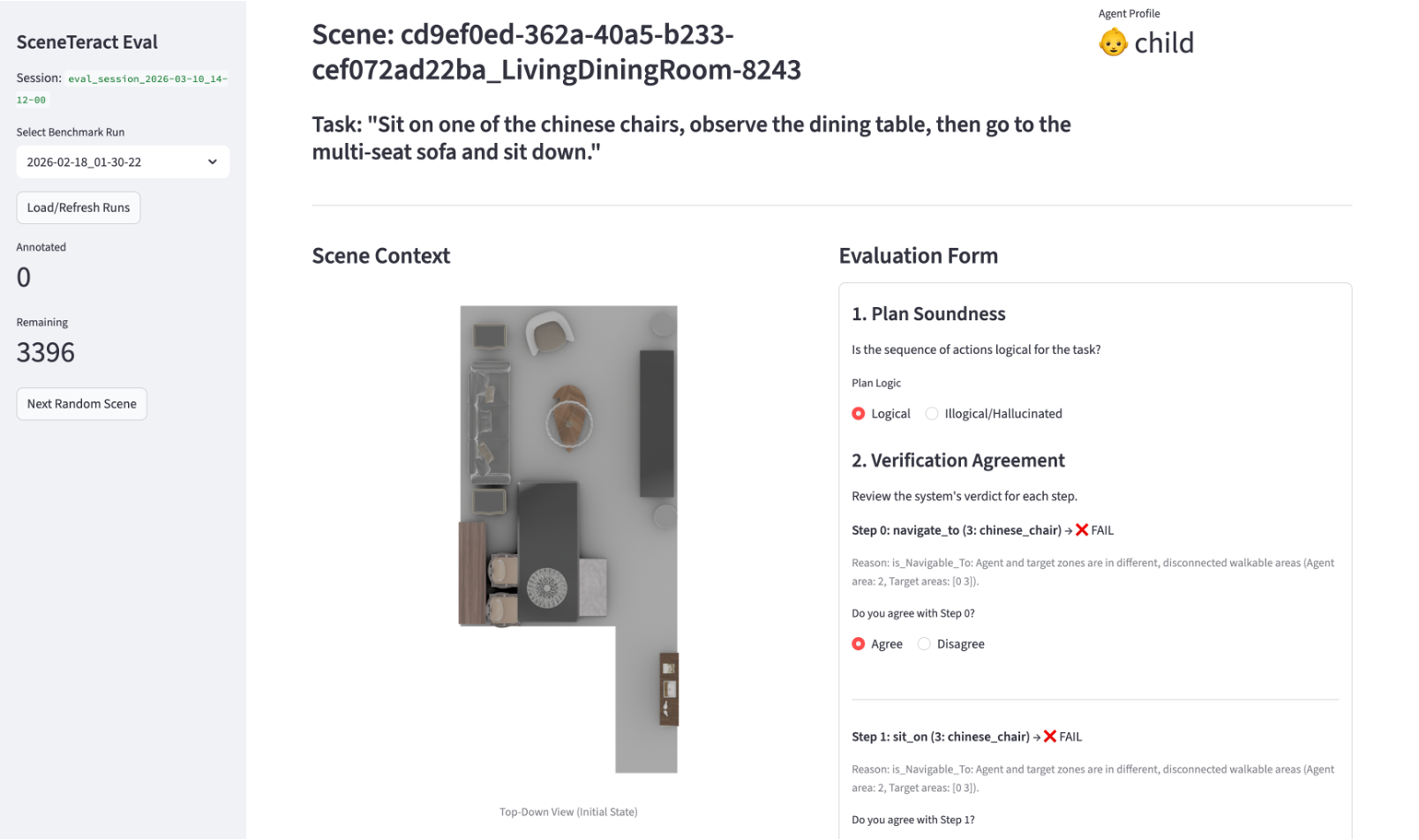} %
        \label{fig:study_top}
    \end{subfigure}
        
    \begin{subfigure}{\linewidth}
        \centering
        \includegraphics[width=0.81\linewidth]{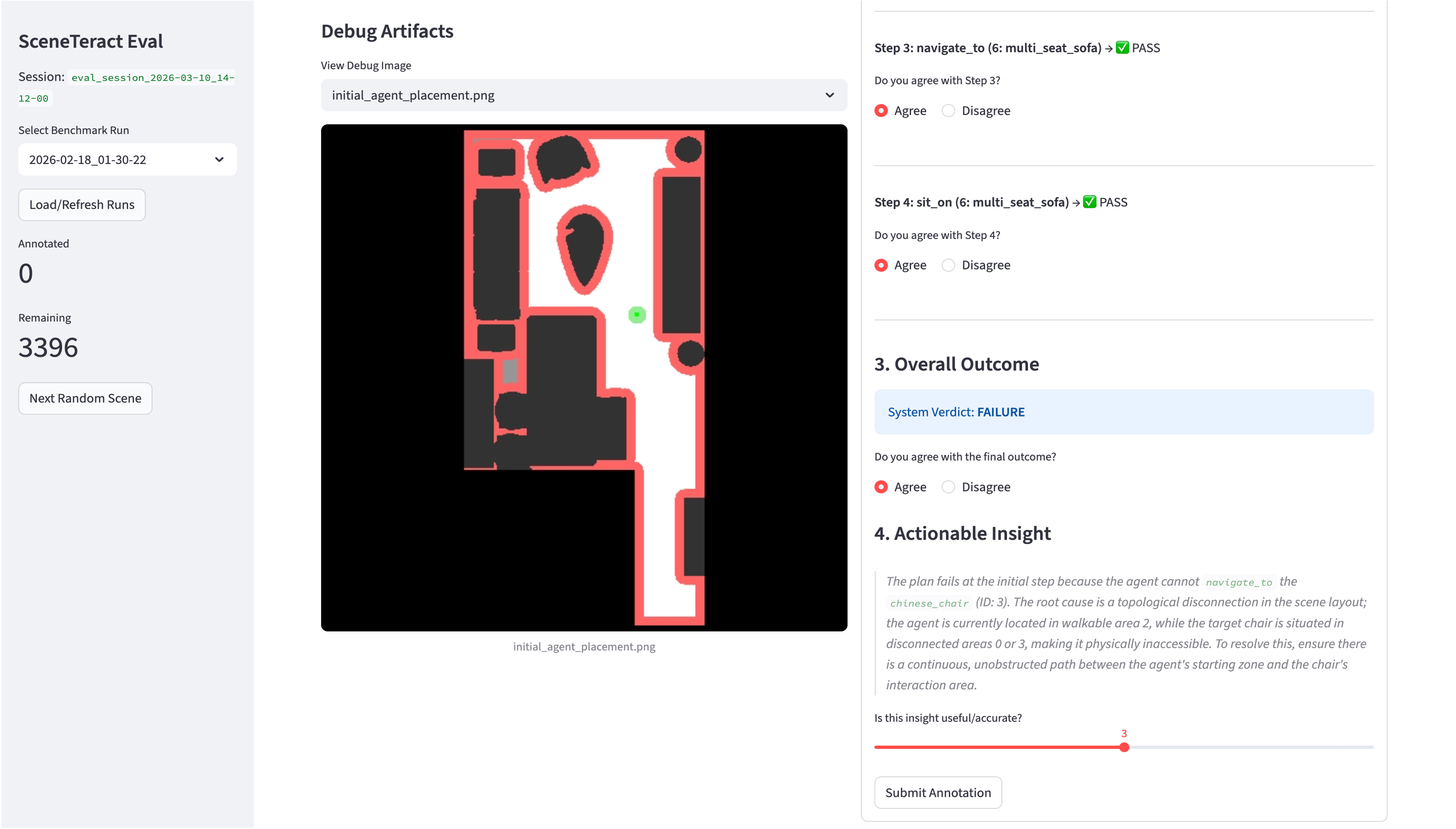} %
        \label{fig:study_bottom}
    \end{subfigure}
    
    \caption{Perceptual study interface. The interface provides comprehensive context for each execution trace, including the target activity, the agent profile, a top-down visual rendering of the scene, and various debug images generated by the grounding engine. Evaluation is conducted by reviewing the logical soundness of the VLM-generated plan, verifying agreement with the system's geometric pass/fail verdicts at both the atomic action step level and the overall task level, and finally rating the utility of the generated actionable insight.}
    \label{fig:streamlit_interface}
\end{figure*}

\paragraph{\textbf{Evaluation Interface.}} 
A multimodal annotation interface was used to expose the internal reasoning of \name{}, as illustrated in Figure~\ref{fig:streamlit_interface}. For each trace, the evaluator was presented with: (1) a top-down rendering of the initial scene state, (2) the target natural-language task and the specific agent profile (Adult, Child, or Wheelchair User), (3) the generated multi-step action plan, and (4) the step-by-step geometric verification results. To ensure sufficient context for physical judgment, the interface also displayed debug artifacts generated by the grounding engine, such as 2D navigation occupancy maps and 3D reachability/visibility debug renders.

\paragraph{\textbf{Grading Criteria.}} 
Each execution trace was evaluated across four primary dimensions:
\begin{enumerate}
    \item \textbf{Plan Soundness:} Whether the high-level plan decomposed by the VLM was logically consistent and appropriate for the given task and scene inventory.
    \item \textbf{Action-Level Verification Agreement:} Agreement with the binary Pass/Fail verdict returned by the grounding engine for each individual atomic action in the plan.
    \item \textbf{Overall Outcome Agreement:} Whether the final success or failure status of the complete activity matched physical judgment.
    \item \textbf{Insight Utility:} A rating on a scale of 1 to 5 stars evaluating the accuracy and diagnostic value of the final "Actionable Insight" summary generated by the system.
\end{enumerate}

The high agreement rates reported in Sec.~\ref{sec:study} (91.0\% on overall outcomes and 96.3\% on individual steps) demonstrate that the \name{} grounded verification engine provides a reliable, human-aligned proxy for physical feasibility in complex 3D environments.

\section{VLM Benchmark}
\label{supp:benchmark}

This section expands on the VLM benchmarking methodology introduced in Sec.~\ref{sec:exp-benchmark}, providing formal definitions of the evaluation tracks and the prompt templates used to query the models. We evaluate the models at two levels of granularity: a holistic task-level assessment (\textit{Direct}) and a step-by-step atomic assessment (\textit{Decomposed}). Qualitative predictions for both settings are shown in Section~\ref{sec:additional:vlm}.

\subsection{Direct Task-Level Evaluation}
\label{supp:benchmark:direct}

In the \textit{Direct} setting, the VLM is tasked with evaluating the overall physical feasibility of a complex, multi-step activity (\(\mathcal{T}\)) in a zero-shot manner. This tests the model's ability to holistically reason about spatial navigation, object constraints, and agent embodiment over long horizons without explicit step-by-step guidance. Similar to the context used for the semantic planner described in Section~\ref{supp:implementation:planner}, the model is provided with a multimodal prompt comprising:
\begin{enumerate}
    \item A top-down visual rendering of the room, as shown in Section~\ref{sec:additional:vlm}.
    \item A textual serialization of the agent's explicit physical capabilities.
    \item A textual JSON serialization of the scene's object inventory and coordinates.
    \item The high-level natural language task description.
\end{enumerate}

The exact prompt template used for this evaluation is provided below:

\begin{tcolorbox}[breakable, boxrule=0pt, colframe=cvprblue, sharp corners, left=1mm, right=1mm, top=0.2mm, bottom=0.2mm, title=Task-Level Evaluation Prompt]
\footnotesize
{
Consider an embodied agent in a 3D room. 

\textbf{AGENT SPECS:}
\textcolor{cvprblue}{\{agent\_json\}}

\textbf{WORLD CONTEXT:}
The following JSON describes all objects in the room and their coordinates:
\textcolor{cvprblue}{\{scene\_json\}}

\textbf{GOAL:}
The agent's task is to: ``\textcolor{cvprblue}{\{task\_description\}}''

\textbf{NOTE ON LOCOMOTION:}
The task description may contain specific movement verbs (e.g., ``walk'', ``step'', ``stand up''). Please interpret all such verbs as a general intent to ``navigate to'' or ``move''. Do not mark a task as impossible simply because a wheelchair user cannot literally ``walk'' or ``stand''; focus exclusively on whether the agent can spatially navigate to and physically reach the targets.

\textbf{IMAGES PROVIDED:}
1. Top-Down View: A clean overhead view of the room.

\textbf{QUESTION:}
Based on the provided images and the agent's physical capabilities, is it possible for the agent to complete this task? 
Consider, notably, if paths are wide enough for the agent to navigate, if objects are within reach, and if there is enough clearance to interact with items.

Respond strictly in JSON format: 
\texttt{\{
  ``prediction'': bool, 
  ``reasoning'': ``A concise explanation of why the task is possible or impossible, mentioning specific spatial constraints.''
\}}
}
\end{tcolorbox}

\subsection{Decomposed Action-Level Evaluation}
\label{supp:benchmark:decomposed}

In the \textit{Decomposed} setting, we evaluate the model's spatial perception and reasoning at the atomic level. Mirroring the internal logic of \name{}, complex tasks are broken down into single interaction units (\eg, \texttt{open} the \texttt{wardrobe}). For this evaluation, the VLM is provided with a targeted, object-centric prompt:
\begin{enumerate}
    \item A top-down visual rendering of the room, with the specific target object highlighted by a red bounding box, as shown in Section~\ref{sec:additional:vlm}.
    \item A textual serialization of the agent's explicit physical capabilities.
    \item A textual JSON serialization of the scene's object inventory.
    \item The specific atomic action intent (\eg, \texttt{open}) and the target object's category (\eg, \texttt{cabinet}).
\end{enumerate}

The model is evaluated strictly on whether its binary prediction for this single step matches the geometrically verified label generated by \name{}. The prompt template used for the atomic action evaluation is provided below:

\begin{tcolorbox}[breakable, boxrule=0pt, colframe=cvprblue, sharp corners, left=1mm, right=1mm, top=0.2mm, bottom=0.2mm, title=Action-Level Evaluation Prompt]
\footnotesize
{
Consider an embodied agent in a 3D room. 

\textbf{AGENT SPECS:}
\textcolor{cvprblue}{\{agent\_json\}}

\textbf{WORLD CONTEXT:}
The following JSON describes all objects in the room and their coordinates:
\textcolor{cvprblue}{\{scene\_json\}}

\textbf{GOAL:}
The agent needs to perform the following action: ``\textcolor{cvprblue}{\{semantic\_action\}}'' on the ``\textcolor{cvprblue}{\{target\_object\}}''.

\textbf{IMAGES PROVIDED:}
The provided image is a top-down overhead view of the room with the target object highlighted by a red bounding box.

\textbf{QUESTION:}
Based on the provided image and the agent's physical capabilities, can the agent successfully perform the action ``\textcolor{cvprblue}{\{semantic\_action\}}'' on the ``\textcolor{cvprblue}{\{target\_object\}}''? 
Consider if the agent can navigate to the object, if the object is within reach, and if there is enough clearance.

Respond strictly in JSON format: 
\texttt{\{
  ``prediction'': bool, 
  ``reasoning'': ``A concise explanation of why the action is possible or impossible, mentioning specific spatial constraints.''
\}}
}
\end{tcolorbox}

Note that the task-level evaluation can still be done in this setup with the following consideration : a task is considered feasible only if the model predicts that \textit{all} constituent atomic actions are feasible. 

\subsection{Implementation Details}
\label{supp:benchmark:implementation}

\paragraph{\textbf{Models and Inference.}}
To ensure reproducibility, all models were evaluated with a sampling temperature set to $0.0$. For proprietary frontier models, we utilized the Google Vertex AI API with all other generation parameters set to their defaults. Open-weight models were loaded in \texttt{bfloat16} precision and served locally on a single NVIDIA RTX 5000 Ada 32GB GPU using the vLLM~\cite{vllm} engine to optimize multimodal inference throughput. We evaluated the following official Hugging Face implementations:

\begin{itemize}
    \item \href{https://huggingface.co/Qwen/Qwen3-VL-8B-Instruct}{\texttt{Qwen/Qwen3-VL-8B-Instruct}}
    \item \href{https://huggingface.co/Qwen/Qwen3-VL-4B-Instruct}{\texttt{Qwen/Qwen3-VL-4B-Instruct}}
    \item \href{https://huggingface.co/google/gemma-3-12b-it}{\texttt{google/gemma-3-12b-it}}
    \item \href{https://huggingface.co/google/gemma-3-4b-it}{\texttt{google/gemma-3-4b-it}}
    \item \href{https://huggingface.co/mistralai/Ministral-3b-instruct}{\texttt{mistralai/Ministral-3b-instruct}}
\end{itemize}

\paragraph{\textbf{Evaluation Split.}}
To ensure strict isolation between evaluation and any subsequent post-training (detailed in Section~\ref{supp:grpo}), we performed an 80/20 scene-based split on the generated dataset. Crucially, this split was performed at the \textit{scene} level rather than the task level, guaranteeing that no physical room layouts seen during the training phase leaked into the evaluation phase. All VLM benchmarking results reported in Sec.~\ref{sec:exp-benchmark} and in this appendix are computed exclusively on this hold-out set, which comprises 227 unique scenes, 681 distinct task-level verification traces, and 1,600 individual atomic actions.

\section{GRPO Post-Training}
\label{supp:grpo}

This section details the reinforcement learning phase, where we leverage \name{}'s grounded verification engine as a reward signal to align a Vision-Language Model's spatial and functional reasoning with physical reality.

\subsection{Training Architecture \& Hyperparameters}
\label{supp:grpo:hyperparams}

We utilize the open-weight \texttt{Qwen3-VL-4B-Instruct} as our base reasoning model. To maintain the model's general multimodal capabilities while injecting functional awareness, we perform a parameter-efficient fine-tuning using Low-Rank Adaptation (LoRA)~\cite{hu2022lora} on the query and value projection layers of the attention blocks. The training was conducted on a single NVIDIA RTX 5000 Ada 32GB GPU using the \texttt{trl}~\cite{vonwerra2020trl} library. The exact hyperparameters used for the GRPO alignment are summarized in Table~\ref{tab:grpo_hyperparams}.

\begin{table}[htbp]
\centering
\caption{Hyperparameters used for GRPO post-training.}
\label{tab:grpo_hyperparams}
\resizebox{\columnwidth}{!}{%
\begin{tabular}{l l}
\toprule
\textbf{Hyperparameter} & \textbf{Value} \\
\midrule
Base Model & \texttt{Qwen3-VL-4B-Instruct} \\
Fine-tuning Method & LoRA ($r=8, \alpha=32, \text{dropout}=0.1$) \\
Target Modules & \texttt{q\_proj, v\_proj} \\
Batch Size & 12 \\
Training Steps & 2,800 \\
Learning Rate & $2 \times 10^{-5}$ \\
Group Size ($G$) & 12 \\
Max Completion Length & 512 tokens \\
Max Prompt Length & 2048 tokens \\
Precision & \texttt{bfloat16} \\
Attention Implementation & Flash Attention 2 \\
\bottomrule
\end{tabular}%
}
\end{table}

\subsection{Formulation \& Rewards}
\label{supp:grpo:formulation}

\paragraph{\textbf{Prompts.}}

The training data consists of the atomic action evaluation samples described in Section~\ref{supp:benchmark:decomposed}. Each sample includes a top-down rendering with the target object highlighted. The model is guided by a system prompt that enforces a Chain-of-Thought (CoT)~\cite{CoT} reasoning format and a user prompt that provides the agent description and action intent.

\begin{tcolorbox}[boxrule=0pt, colframe=cvprblue, sharp corners, left=1mm, right=1mm, top=0.2mm, bottom=0.2mm, title=GRPO System Prompt]
\footnotesize
{
You are an embodied AI spatial critic. The user asks a question about spatial feasibility, and you must solve it. You must first think about the reasoning process in your mind, considering spatial constraints like navigation, reach, and clearance. Then, provide the final answer. Your reasoning process and your final boolean answer MUST be enclosed within \texttt{<think> </think>} and \texttt{<answer> </answer>} tags, respectively. For example:
\\\\
\texttt{<think> reasoning process here </think>}\\
\texttt{<answer>True</answer>}
}
\end{tcolorbox}

\begin{tcolorbox}[boxrule=0pt, colframe=cvprblue, sharp corners, left=1mm, right=1mm, top=0.2mm, bottom=0.2mm, title=GRPO User Prompt Template]
\footnotesize
{
Consider an embodied agent in a 3D room. 

\textbf{AGENT DESCRIPTION:} The agent is \textcolor{cvprblue}{\{agent\_desc\}}.

\textbf{GOAL:}
The agent needs to perform the following action: ``\textcolor{cvprblue}{\{semantic\_action\}}'' on the ``\textcolor{cvprblue}{\{target\_object\}}''.

\textbf{IMAGES PROVIDED:}
The provided image is a top-down overhead view of the room with the target object highlighted by a red bounding box.

\textbf{QUESTION:}
Based on the provided image and the agent's physical capabilities, can the agent successfully perform the action ``\textcolor{cvprblue}{\{semantic\_action\}}'' on the ``\textcolor{cvprblue}{\{target\_object\}}''? 
Consider if the agent can navigate to the object, if the object is within reach, and if there is enough clearance.
}
\end{tcolorbox}

\paragraph{\textbf{Reward Design.}}

We optimize the model policy using Group Relative Policy Optimization (GRPO)~\cite{shao2024deepseekmath}. For each training sample, we generate a group of $G=12$ independent generations. To address the natural class imbalance in the generated dataset (where only $\sim 25\%$ of interactions result in physical failure), we apply an upsampling to the failed cases during dataset preparation, raising the proportion of \texttt{False} samples to $\sim 40\%$. The model receives a cumulative reward defined as follows:
\begin{equation}
  R = r_{\text{format}} + r_{\text{correct}} + r_{\text{spatial}}\,, \quad \text{where}
\end{equation}
\begin{itemize}
    \item \textbf{Format Reward ($r_{\text{format}}$):} A binary reward ($+0.5$) is assigned if the completion strictly adheres to the mandated \texttt{<think>...</think>} XML-style template. We apply a severe penalty ($-1.0$) if the model simply reproduces the placeholder text from the system prompt.
    \item \textbf{Asymmetric Correctness Reward ($r_{\text{correct}}$):} To further counter class imbalance, a correct \texttt{False} prediction yields a higher reward ($+3.0$) than a correct \texttt{True} prediction ($+2.0$), encouraging the model to prioritize the identification of physical failure modes.
    \item \textbf{Spatial Reasoning Reward ($r_{\text{spatial}}$):} For true negative samples, we provide a targeted reward ($+1.0$) if the model's reasoning trace explicitly mentions keywords related to the geometric property that caused the failure in the \name{} engine. The mapping between geometric properties and their triggering keywords is detailed in Table~\ref{tab:grpo_keywords}.
\end{itemize}

\begin{table}[htbp]
\centering
\caption{Keyword mapping for the Spatial Reasoning Reward ($r_{\text{spatial}}$). A reward is triggered if the model's generated \texttt{<think>} trace contains any of the target keywords corresponding to the specific physical failure identified by the verification engine.}
\label{tab:grpo_keywords}
\resizebox{\columnwidth}{!}{%
\begin{tabular}{l l}
\toprule
\textbf{Failed Geometric Property} & \textbf{Target Keywords} \\
\midrule
\texttt{is\_Navigable\_To} & \textit{path, walk, navigate, access, blocked, way, stuck} \\
\texttt{is\_Reachable} & \textit{reach, distance, too far, height, arm, long, short, touch} \\
\texttt{is\_Interactable} & \textit{handle, grip, part, grasp, rim, side} \\
\texttt{has\_Clearance} & \textit{clearance, blocked, fit, narrow, collision, hit, space, swing} \\
\bottomrule
\end{tabular}%
}
\end{table}

\subsection{Training Dynamics}

\begin{figure*}[t]
\includegraphics[width=\textwidth]{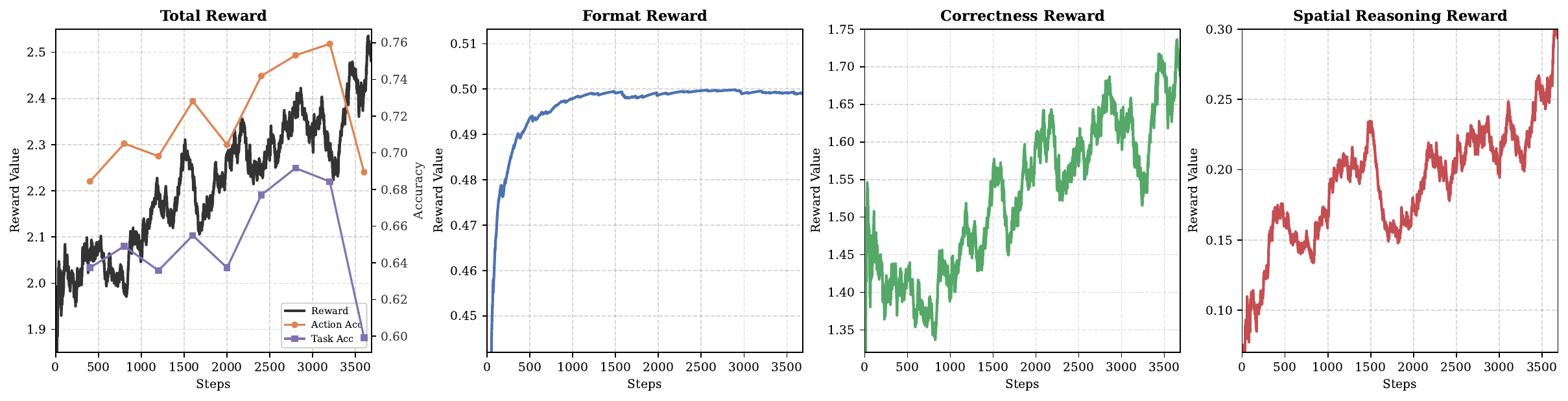}
\caption{Training dynamics during the GRPO alignment phase. The plots display the progression of the smoothed rewards over optimization steps. The \textit{Total Reward} panel overlays the test-set accuracies (evaluated every 400 steps), showing that reward optimization effectively improves functional reasoning performance, before observing a typical case of reward hacking.}
\label{fig:grpo_rewards}
\end{figure*}

Figure~\ref{fig:grpo_rewards} illustrates the evolution of rewards and evaluation accuracy over the GRPO post-training process. 

\paragraph{\textbf{Reward Convergence.}}
The \textit{Format Reward} converges rapidly, indicating the model quickly learns the XML-style reasoning structure. Both the \textit{Asymmetric Correctness Reward} and the \textit{Spatial Reasoning Reward} show steady improvement, confirming that the model is learning to identify physical bottlenecks and provide the correct semantic justifications for its feasibility judgments.

\paragraph{\textbf{Accuracy Alignment.}}
The \textit{Total Reward} plot (top-left) includes an overlay of the model's accuracy on the 20\% hold-out test set, evaluated every 400 steps. We observe a strong positive correlation between reward maximization and downstream performance, with both atomic action and derived task-level accuracies reaching their peak as the spatial reward stabilizes around step \(2,800\). 

\paragraph{\textbf{Over-Optimization.}}
It is worth noting that if training is extended significantly, we observe a classic case of over-optimization (\eg, reward hacking). While the training rewards continue to climb, the hold-out test accuracy begins to drop sharply past approximately step 3,000. This divergence indicates that the model begins to overfit to the training set and exploit the reward structure by generating increasingly erratic reasoning traces that prioritize keyword matching over physical logic. Consequently, we utilize the peak-performing checkpoint at step 2,800 for all final evaluations to preserve generalized spatial reasoning capabilities. The robust performance gains achieved with these baseline configurations suggest that \name{}-guided RL alignment is a highly promising direction for future research, with significant potential for further optimization.

\subsection{Example of Reasoning Trace}

To illustrate the effect our reward formulation, we provide an example of a generated reasoning trace during optimization. In this scenario, the \textbf{Wheelchair User} agent is tasked with the action \texttt{sit\_on} for a target \texttt{dining\_chair}. The model generated the following Chain-of-Thought completion:

\begin{tcolorbox}[breakable, boxrule=0pt, colframe=white, sharp corners, left=1mm, right=1mm, top=0.2mm, bottom=0.2mm]
\footnotesize
{
\texttt{<think>}\\
The agent is seated in a manual wheelchair. To sit on the dining chair, the agent must first navigate to it, which requires a clear path and sufficient space for the wheelchair to approach and stop. The dining chair is positioned at the head of the dining table on the right side of the room.
There are no objects obstructing the direct path to it. However, the chair is located in a corner or tight space next to the table, and there is a dining table with multiple chairs surrounding it, which may reduce clearance. The wheelchair's width (typically around 60-70 cm) must fit into the space around the chair, which appears to be tight with no apparent clearance or buffer space.
Additionally, the red bounding box indicates the chair is selected, but visually, the space around it seems too narrow to allow the wheelchair to maneuver in and sit without hitting adjacent furniture. Hence, the agent cannot successfully perform the action ``sit\_on'' on the ``dining\_chair'' due to insufficient clearance and navigational constraints.\\
\texttt{<answer>False</answer>}
}
\end{tcolorbox}
Based on this generation and the \name{} grounded label, the reward engine assigned the following scores:
\begin{itemize}
    \item \textbf{Format ($+0.5$):} Correct use of \texttt{<think>} and \texttt{<answer>} tags.
    \item \textbf{Correctness ($+3.0$):} Correctly predicted \texttt{False}, capturing the higher asymmetric reward for identifying a failure.
    \item \textbf{Spatial Reasoning ($+1.0$):} The \name{} engine failed this action due to an \texttt{is\_Navigable\_To} violation. The model successfully captured the dense reward by explicitly mentioning relevant physical keywords (e.g., \textit{``navigate''}, \textit{``path''}, \textit{``fit''}) in its reasoning trace.
\end{itemize}

Interestingly, while specific dimensions are not provided in the prompt, we observe that the model independently derives a sensible clearance width for the wheelchair (\ie, \textit{``typically around 60-70 cm''}), which closely aligns with the \qty{0.65}{\meter} parameter used by our verification engine. This provides an indication that the model grounds its assessment process in its internal common-sense physical knowledge.

\section{Real-Scene Evaluation Protocol}
\label{supp:scannet}

This section details the acquisition, annotation, and evaluation protocol for the human-labeled real-scene set used in our sim-to-real generalization results (Sec.~\ref{sec:exp-grpo}).

\subsection{3D Gaussian Splatting Acquisition}
\label{supp:scannet:3dgs}

\begin{figure*}[t]
\centering
\includegraphics[width=0.8\textwidth]{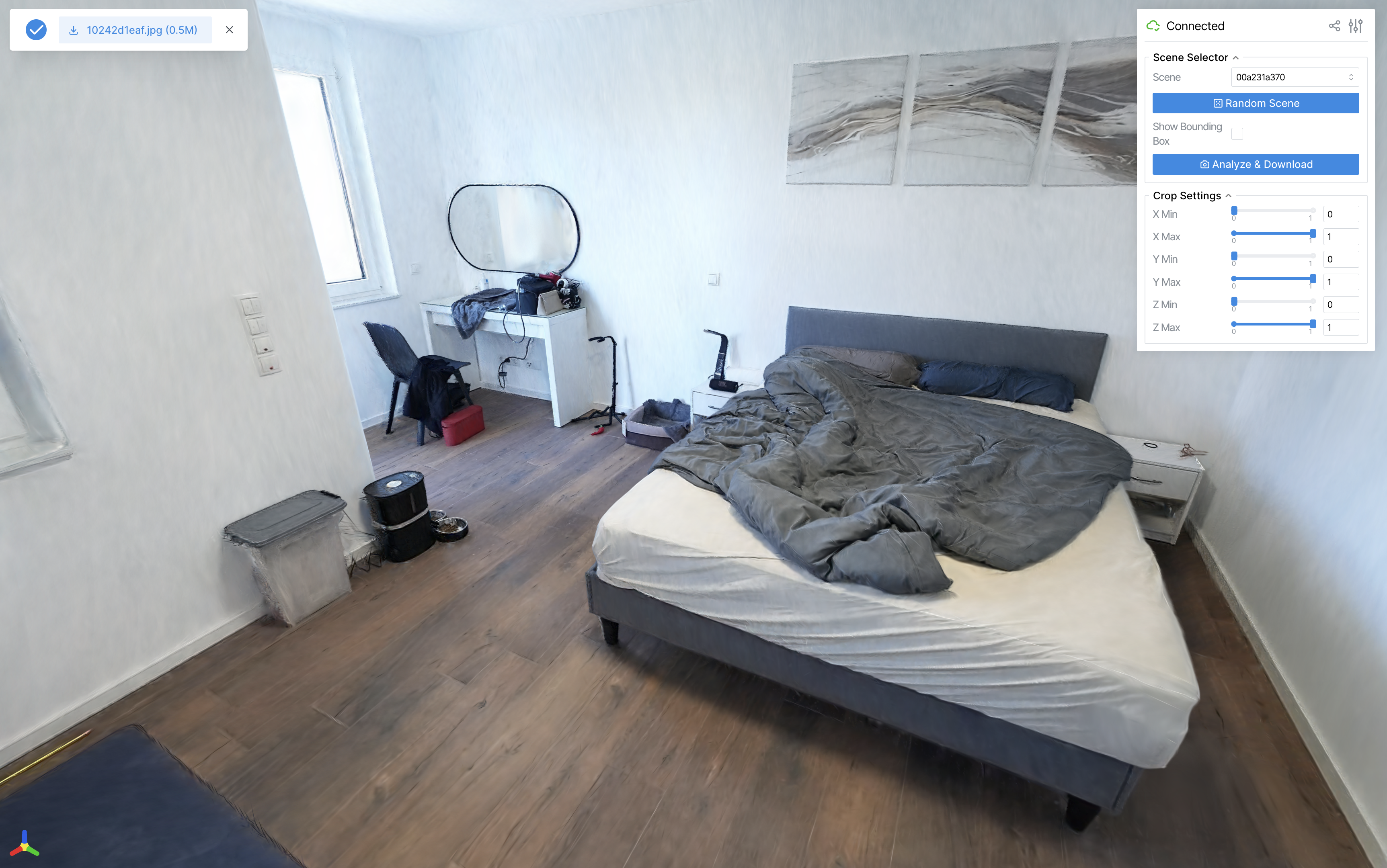}
\caption{3D Gaussian Splatting navigation interface used to capture in-context images from ScanNet++~\cite{yeshwanth2023scannet++} reconstructions. We navigate the reconstructed splat volume and capture a snapshot from the current viewpoint.}
\label{fig:scannet_viser}
\end{figure*}

We source 100 scenes from ScanNet++~\cite{yeshwanth2023scannet++}, using pre-computed 3D Gaussian Splatting reconstructions from the SceneSplat~\cite{li2025scenesplat} release\footnote{\url{https://huggingface.co/datasets/GaussianWorld/scene_splat_7k}}. To capture in-context images from these reconstructions, we built a lightweight browser-based viewer on top of Viser~\cite{yi2025viser}, illustrated in Figure~\ref{fig:scannet_viser}. For each scene, we freely navigate the reconstructed splat volume and capture a snapshot from the current camera viewpoint at \(1920{\times}1080\) resolution.

To accelerate the subsequent annotation stage, each captured snapshot is passed to a VLM prompted to suggest one plausible and one implausible candidate action for the visible scene content. These suggestions are stored as image metadata and serve purely as an optional example or drafting aid that annotators may rework or discard; the feasibility labels themselves are set exclusively by human annotators (Section~\ref{supp:scannet:annotation}).

\subsection{Human Annotation Protocol}
\label{supp:scannet:annotation}

\begin{figure*}[t]
\centering
\includegraphics[width=0.8\textwidth]{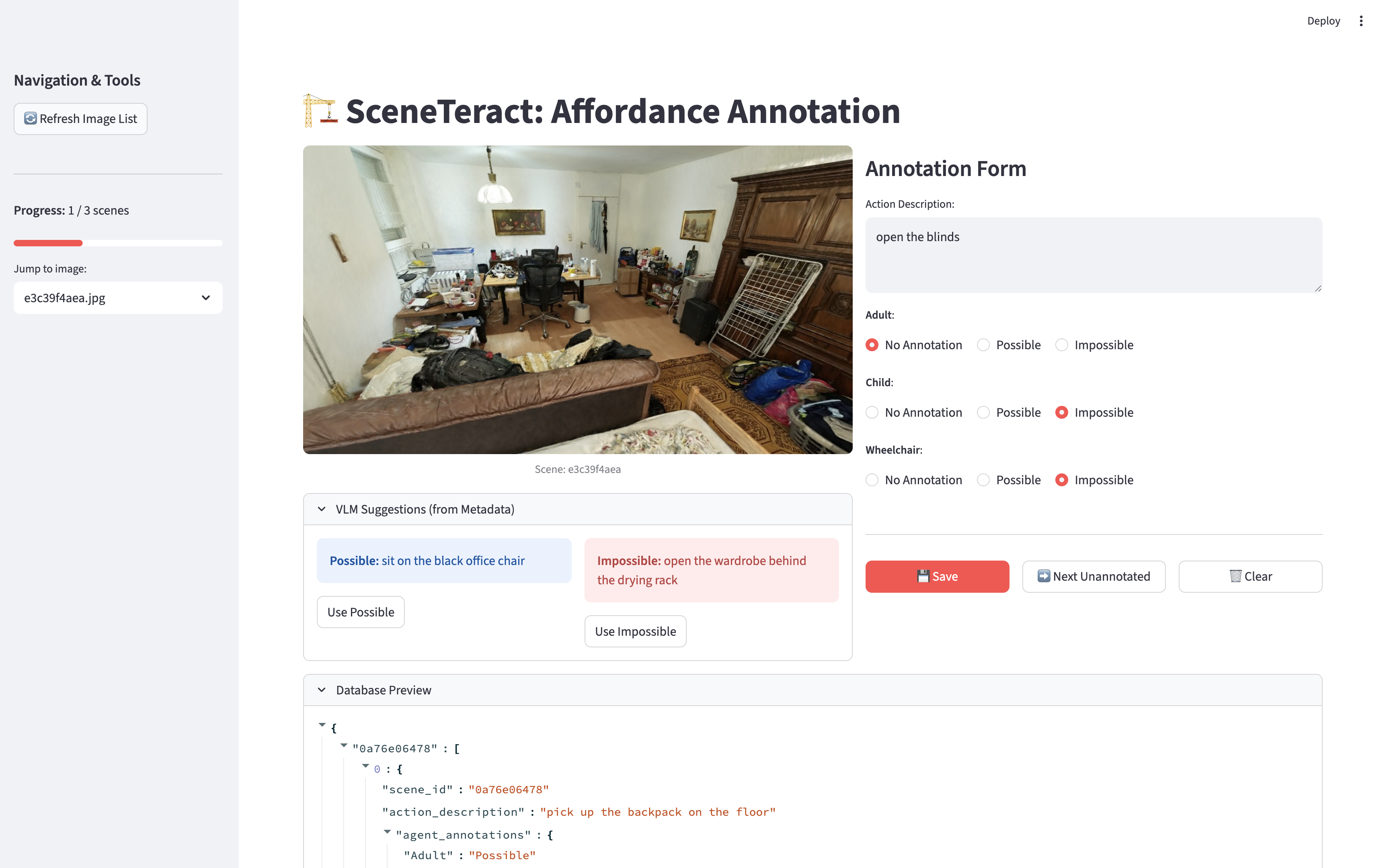}
\caption{Human annotation interface for the real-scene evaluation set. For each captured scene, the annotator provides an action description and independently labels its feasibility for the Adult, Child, and Wheelchair agent profiles.}
\label{fig:scannet_annotator}
\end{figure*}

Captured snapshots are annotated using a dedicated Streamlit interface, shown in Figure~\ref{fig:scannet_annotator}. For each scene, the annotator writes a natural-language action description, optionally starting from the VLM-suggested candidate (Section~\ref{supp:scannet:3dgs}) and freely editing it, and independently assigns a feasibility label, \texttt{Possible} or \texttt{Impossible}, for each of the three agent profiles (Adult, Child, Wheelchair). A scene may accumulate multiple annotated actions, and the interface tracks per-scene annotation progress. Three annotators shared the annotation workload across the 100 scenes. Aggregating labeled (scene, action, agent) triplets, and excluding entries left unannotated for a given profile, yields the 502 real-scene labels reported in Sec.~\ref{sec:exp-grpo}.

\subsection{VLM Feasibility Prediction Prompt}
\label{supp:scannet:eval}

To evaluate a VLM against the human-labeled real-scene set, we query it with a prompt structurally matching the GRPO user prompt template (Section~\ref{supp:grpo:formulation}), adapted for real-world images and phrased agent descriptions. The agent profile is expressed in natural language as ``an average-height adult standing'' (Adult), ``a standing 10-year-old child'' (Child), or ``an adult seated in a manual wheelchair'' (Wheelchair). The full prompt template is shown below:

\begin{tcolorbox}[breakable, boxrule=0pt, colframe=cvprblue, sharp corners, left=1mm, right=1mm, top=0.2mm, bottom=0.2mm, title=VLM Feasibility Prediction Prompt]
\footnotesize
{
Consider an embodied agent in a 3D room.
\\\\
\textbf{AGENT DESCRIPTION:} The agent is \textcolor{cvprblue}{\{agent\_description\}}.

\textbf{GOAL:}
The agent needs to perform the following action: ``\textcolor{cvprblue}{\{action\}}'' on the ``visible target''.

\textbf{IMAGES PROVIDED:}
The provided image is a real-world view of the room containing the target of the action.

\textbf{QUESTION:}
Based on the provided image and the agent's physical capabilities, can the agent successfully perform the action ``\textcolor{cvprblue}{\{action\}}''?
Consider if the agent can navigate to the object, if the object is within reach, and if there is enough clearance.

Respond strictly in JSON format: \{\{``prediction'': bool, ``reasoning'': string\}\}
}
\end{tcolorbox}

\section{Societal Impact}
\label{supp:societal}

We believe our method will mainly yield positive societal impact by promoting accessibility in 3D scene understanding and embodied AI. Motivating this work, the rapid advancement of 3D scene synthesis holds significant potential across industries, from virtual reality and architectural design to domestic robotics. While other vision fields, like video generation~\cite{vbench} and multi-view synthesis~\cite{mvgbench}, have recently adopted comprehensive benchmarks tailored to their respective domains, evaluation paradigms in scene synthesis predominantly focus on visual realism, as discussed in Sec.~\ref{sec:related}. This overlooks the practical and real-life implications of a 3D environment and can lead to the generation of spaces that exclude diverse populations.

\name{} addresses this by highlighting the concept of relational affordances. By providing a scalable procedure to verify functionality against explicit, diverse embodiment constraints, our work encourages the development of universally accessible virtual environments. Furthermore, by exposing the gap in frontier VLMs and offering a first pathway to correct it, \name{} contributes to the safety and reliability of embodied AI systems, ensuring they respect physical limitations before executing actions in the real world.

\paragraph{\textbf{Limitations and Risks.}} While \name{} promotes inclusivity, the parameterization of human bodies into geometric constraints (\eg, bounding boxes and reach radii) is inherently reductive. This simplification should be especially taken into consideration when it comes to applying these assessments to the physical world, where complex ergonomic, cognitive, and dynamic factors play a critical role. There is a risk that developers may interpret a successful verification trace as an absolute guarantee of real-world accessibility. We emphasize that our agent profiles are representative instances designed to probe algorithmic spatial reasoning, and should not replace comprehensive ergonomic or clinical accessibility standards.

\section{Additional Results}
\label{supp:qualitative}

This section presents additional qualitative results for \name{}. We first include complete diagnostic reports \(\mathcal{R}\) from our verification engine in Section~\ref{supp:qualitative:reports}, then provide extended VLM predictions at both task (\ie, \textit{Direct}) and atomic-action (\ie, \textit{Decomposed}) levels in Section~\ref{sec:additional:vlm}.

\subsection{Plans \& Diagnostic Reports}
\label{supp:qualitative:reports}

At the end of a verification run, \name{} automatically generates a human-readable \textit{Actionable Insight} summarizing the outcome. This is achieved by prompting \texttt{gemini-3-flash-preview} with the raw JSON verification report, which contains the step-by-step pass/fail states and quantitative geometric error messages (\eg, ``Required distance: 1.15m, exceeds Agent's reach: 0.70m''), and instructing it to synthesize a concise diagnostic paragraph explaining the root physical cause of failure or confirming complete execution success. The system prompt used for this generation is provided below.

\begin{tcolorbox}[breakable, boxrule=0pt, colframe=cvprblue, sharp corners, left=1mm, right=1mm, top=0.2mm, bottom=0.2mm, title=Insight Generation System Prompt]
\footnotesize
{
\textbf{\# Role}
You are the \textbf{Diagnostic Expert} for SceneTeract, a human-centric 3D evaluation framework. Your goal is to translate a structured technical verification report into a clear, concise, and insightful summary for a human user.

\textbf{\# Input}
You will receive a \textbf{Verification Report} in JSON format. This report details a sequence of actions an agent attempted to perform in a 3D scene, and for each action, a series of geometric property checks (\eg, \texttt{is\_Navigable\_To}, \texttt{is\_Reachable}) with their pass/fail status and specific error messages.

\textbf{\# Task}
Analyze the report and generate a \textbf{single paragraph} (approx. 2-4 sentences) of "Actionable Insight".
\begin{itemize}[leftmargin=2em, label=-, nosep]
    \item \textbf{If the plan FAILED:} Identify the \textit{first} point of failure (the "root cause"). Explain \textit{why} it fails in plain English, incorporating the specific quantitative data from the error message. Be specific (\eg, "The agent cannot reach the mug because it is 1.2m away, exceeding the agent's 0.8m reach limit.").
    \item \textbf{If the plan PASSED:} Confirm that the entire activity is functionally valid for this specific agent. Highlight a key success factor if relevant. Keep it brief and positive. Use \textbf{present tense}.
\end{itemize}

\textbf{\# Style Guidelines}
\begin{itemize}[leftmargin=2em, label=-, nosep]
    \item \textbf{Tone:} Professional, objective, and helpful.
    \item \textbf{Format:} A single paragraph. No bullet points.
    \item \textbf{Content:} Focus on the \textit{physical} and \textit{agent-specific} reasons.
\end{itemize}
}
\end{tcolorbox}

Below, we provide diverse execution traces, including both successful and failed scenarios across different agent profiles. We added colored dots to the relevant objects to help the reader follow the plans.

\newcommand{\inlinestamp}[3]{%
    \tikz[baseline=(node.base)] \node (node) [
        draw=#1,                
        text=#1,                
        line width=1.2pt,       
        double=white,           
        double distance=1pt,    
        rounded corners=3pt,    
        inner sep=3pt,          
        rotate=#3,              
        font=\rmfamily\bfseries %
    ] {#2};%
}

\newcommand{\pass}{\inlinestamp{green!60!black}{PASS}{12}}
\newcommand{\fail}{\inlinestamp{red!70!black}{FAIL}{-10}}

\onecolumn
\subsubsection*{Report \#1: Child in LivingDiningRoom-34204}

\begin{table}[H]
\centering
\begin{minipage}[c]{0.45\textwidth}
    \raggedright
    \textbf{Scene:} \texttt{LivingDiningRoom-34204} \\ [0.6em]
    \includegraphics[width=\textwidth]{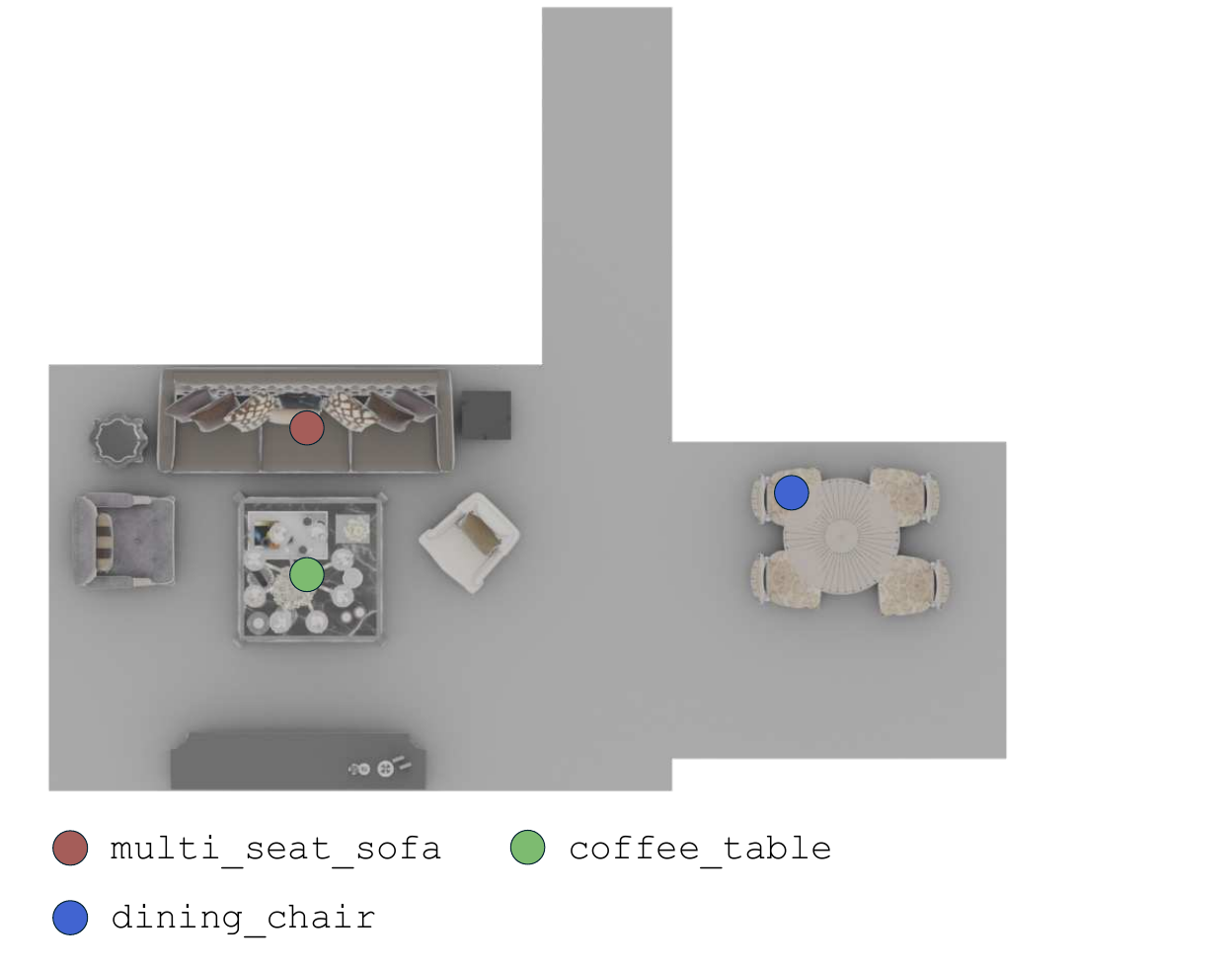}
\end{minipage}
\hfill
\begin{minipage}[c]{0.5\textwidth}
    \raggedright
    \textbf{Agent:} \textsc{child} \\[1.2em]
    \textbf{Task:} ``Sit on the multi-seat sofa, then look at the coffee table, and then move to sit on a dining chair at the dining table.''
\end{minipage}
\end{table}

\paragraph{Overall Success:} \pass

\paragraph{Execution Trace:}
\begin{enumerate}
\item \textbf{Step 1:} (\texttt{navigate\_to}, \texttt{multi\_seat\_sofa}) [ID: 11] --- \textcolor{green!70!black}{\textbf{PASS}}
\begin{itemize}
    \item \texttt{is\_Navigable\_To}: \textcolor{green!70!black}{Pass} - \textit{A collision-free path was found to an interaction zone.}
\end{itemize}
\item \textbf{Step 2:} (\texttt{sit\_on}, \texttt{multi\_seat\_sofa}) [ID: 11] --- \textcolor{green!70!black}{\textbf{PASS}}
\begin{itemize}
    \item \texttt{is\_Navigable\_To}: \textcolor{green!70!black}{Pass} - \textit{A collision-free path was found to an interaction zone.}
\end{itemize}
\item \textbf{Step 3:} (\texttt{look\_at}, \texttt{coffee\_table}) [ID: 8] --- \textcolor{green!70!black}{\textbf{PASS}}
\begin{itemize}
    \item \texttt{is\_Visible}: \textcolor{green!70!black}{Pass} - \textit{Object is robustly visible (77.8\% clear, Centroid: Visible).}
\end{itemize}
\item \textbf{Step 4:} (\texttt{navigate\_to}, \texttt{dining\_chair}) [ID: 2] --- \textcolor{green!70!black}{\textbf{PASS}}
\begin{itemize}
    \item \texttt{is\_Navigable\_To}: \textcolor{green!70!black}{Pass} - \textit{A collision-free path was found to an interaction zone.}
\end{itemize}
\item \textbf{Step 5:} (\texttt{sit\_on}, \texttt{dining\_chair}) [ID: 2] --- \textcolor{green!70!black}{\textbf{PASS}}
\begin{itemize}
    \item \texttt{is\_Navigable\_To}: \textcolor{green!70!black}{Pass} - \textit{A collision-free path was found to an interaction zone.}
\end{itemize}
\end{enumerate}

\begin{tcolorbox}[boxrule=0.5pt, colframe=cvprblue, sharp corners, left=2mm, right=2mm, top=2mm, bottom=2mm, title=\textbf{Actionable Insight}]
The plan is fully executable, as the agent can successfully navigate to and interact with all designated furniture items. The scene layout allows for collision-free access to both the sofa and dining chair, and the agent maintains a clear line of sight to the coffee table with 77.8\% visibility. This confirms that the spatial arrangement and object placements are functionally valid for the agent's physical constraints and movement capabilities.
\end{tcolorbox}
\vspace{2em}

\clearpage
\subsubsection*{Report \#2: Adult Wheelchair in LivingDiningRoom-34204}

\begin{table}[htbp]
\centering
\begin{minipage}[c]{0.45\textwidth}
    \raggedright
    \textbf{Scene:} \texttt{LivingDiningRoom-34204} \\ [0.6em]
    \includegraphics[width=\textwidth]{supplementary/figures/qualitative/report_1.pdf}
\end{minipage}
\hfill
\begin{minipage}[c]{0.5\textwidth}
    \raggedright
    \textbf{Agent:} \textsc{adult wheelchair} \\[1.2em]
    \textbf{Task:} ``Sit on the multi-seat sofa, then look at the coffee table, and then move to sit on a dining chair at the dining table.''
\end{minipage}
\end{table}

\paragraph{Overall Success:} \fail %

\paragraph{Execution Trace:}
\begin{enumerate}
\item \textbf{Step 1:} (\texttt{navigate\_to}, \texttt{multi\_seat\_sofa}) [ID: 11] --- \textcolor{red}{\textbf{FAIL}}
\begin{itemize}
    \item \texttt{is\_Navigable\_To}: \textcolor{red}{Fail} - \textit{Target zones are entirely in non-walkable areas.}
\end{itemize}
\item \textbf{Step 2:} (\texttt{sit\_on}, \texttt{multi\_seat\_sofa}) [ID: 11] --- \textcolor{red}{\textbf{FAIL}}
\begin{itemize}
    \item \texttt{is\_Navigable\_To}: \textcolor{red}{Fail} - \textit{Target zones are entirely in non-walkable areas.}
\end{itemize}
\item \textbf{Step 3:} (\texttt{look\_at}, \texttt{coffee\_table}) [ID: 8] --- \textcolor{green!70!black}{\textbf{PASS}}
\begin{itemize}
    \item \texttt{is\_Visible}: \textcolor{green!70!black}{Pass} - \textit{Object is robustly visible (77.8\% clear, Centroid: Visible).}
\end{itemize}
\item \textbf{Step 4:} (\texttt{navigate\_to}, \texttt{dining\_chair}) [ID: 2] --- \textcolor{green!70!black}{\textbf{PASS}}
\begin{itemize}
    \item \texttt{is\_Navigable\_To}: \textcolor{green!70!black}{Pass} - \textit{A collision-free path was found to an interaction zone.}
\end{itemize}
\item \textbf{Step 5:} (\texttt{sit\_on}, \texttt{dining\_chair}) [ID: 2] --- \textcolor{green!70!black}{\textbf{PASS}}
\begin{itemize}
    \item \texttt{is\_Navigable\_To}: \textcolor{green!70!black}{Pass} - \textit{A collision-free path was found to an interaction zone.}
\end{itemize}
\end{enumerate}

\begin{tcolorbox}[boxrule=0.5pt, colframe=cvprblue, sharp corners, left=2mm, right=2mm, top=2mm, bottom=2mm, title=\textbf{Actionable Insight}]
The plan fails at the initial step because the agent cannot \texttt{navigate\_to} the \texttt{multi\_seat\_sofa} (ID 11). The root cause is that all designated interaction zones for the sofa are located entirely within non-walkable areas, preventing the agent from establishing a valid path to the object. To resolve this, the sofa should be repositioned or the surrounding floor geometry adjusted to ensure the agent's navigation mesh overlaps with the sofa's access points.
\end{tcolorbox}

\clearpage
\subsubsection*{Report \#3: Adult in LivingDiningRoom-8799}

\begin{table}[htbp]
\centering
\begin{minipage}[c]{0.45\textwidth}
    \raggedright
    \textbf{Scene:} \texttt{LivingDiningRoom-8799} \\ [0.6em]
    \includegraphics[width=\textwidth]{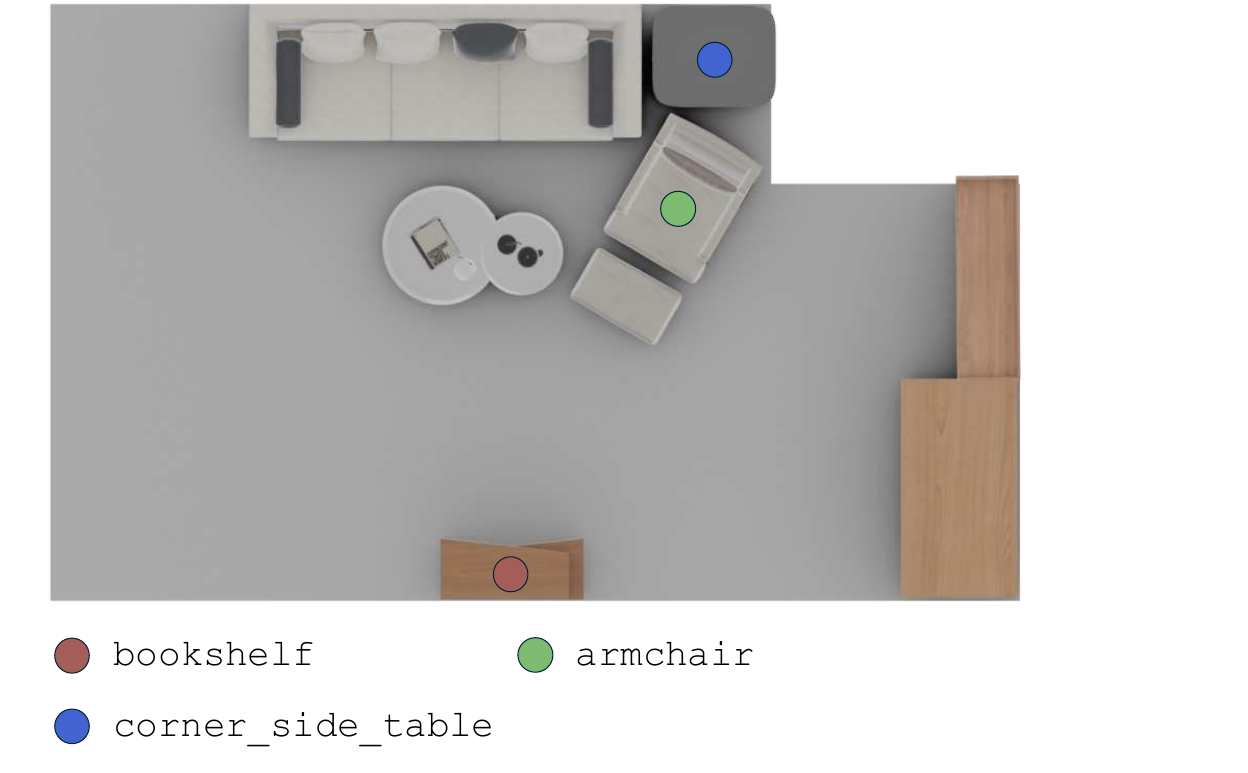}
\end{minipage}
\hfill
\begin{minipage}[c]{0.5\textwidth}
    \raggedright
    \textbf{Agent:} \textsc{adult} \\[1.2em]
    \textbf{Task:} ``Retrieve an item from the bookshelf, walk to the armchair and sit down, then place the item on the corner side table.''
\end{minipage}
\end{table}

\paragraph{Overall Success:} \fail

\paragraph{Execution Trace:}
\begin{enumerate}
\item \textbf{Step 1:} (\texttt{navigate\_to}, \texttt{bookshelf}) [ID: 0] --- \textcolor{green!70!black}{\textbf{PASS}}
\begin{itemize}
    \item \texttt{is\_Navigable\_To}: \textcolor{green!70!black}{Pass} - \textit{A collision-free path was found to an interaction zone.}
\end{itemize}
\item \textbf{Step 2:} (\texttt{pickup\_from}, \texttt{bookshelf}) [ID: 0] --- \textcolor{green!70!black}{\textbf{PASS}}
\begin{itemize}
    \item \texttt{is\_Navigable\_To}: \textcolor{green!70!black}{Pass} - \textit{A collision-free path was found to an interaction zone.}
    \item \texttt{is\_Reachable}: \textcolor{green!70!black}{Pass} - \textit{Object is reachable. Required distance: 0.21m, Agent's reach: 0.70m.}
\end{itemize}
\item \textbf{Step 3:} (\texttt{navigate\_to}, \texttt{armchair}) [ID: 5] --- \textcolor{green!70!black}{\textbf{PASS}}
\begin{itemize}
    \item \texttt{is\_Navigable\_To}: \textcolor{green!70!black}{Pass} - \textit{A collision-free path was found to an interaction zone.}
\end{itemize}
\item \textbf{Step 4:} (\texttt{sit\_on}, \texttt{armchair}) [ID: 5] --- \textcolor{green!70!black}{\textbf{PASS}}
\begin{itemize}
    \item \texttt{is\_Navigable\_To}: \textcolor{green!70!black}{Pass} - \textit{A collision-free path was found to an interaction zone.}
\end{itemize}
\item \textbf{Step 5:} (\texttt{release\_on}, \texttt{corner\_side\_table}) [ID: 6] --- \textcolor{red}{\textbf{FAIL}}
\begin{itemize}
    \item \texttt{is\_Navigable\_To}: \textcolor{red}{Fail} - \textit{Agent and target zones are in different, disconnected walkable areas (Agent area: 1, Target areas: [0 2]).}
    \item \texttt{is\_Reachable}: \textcolor{red}{Fail} - \textit{Object not reachable. Required distance: 0.78m, exceeds Agent's reach: 0.70m.}
\end{itemize}
\end{enumerate}

\begin{tcolorbox}[boxrule=0.5pt, colframe=cvprblue, sharp corners, left=2mm, right=2mm, top=2mm, bottom=2mm, title=\textbf{Actionable Insight}]
The plan fails because the agent cannot \texttt{release\_on} the \texttt{corner\_side\_table} after sitting on the armchair. The primary issue is a lack of connectivity; the agent is located in a disconnected walkable area from the table's interaction zones. Furthermore, even if the agent could approach, the table remains physically unreachable at a distance of 0.78m, which exceeds the agent's 0.70m reach limit.
\end{tcolorbox}
\vspace{2em}

\clearpage
\subsubsection*{Report \#4: Adult Wheelchair in LivingDiningRoom-24410}

\begin{table}[htbp]
\centering
\begin{minipage}[c]{0.45\textwidth}
    \raggedright
    \textbf{Scene:} \texttt{LivingDiningRoom-24410} \\ [0.6em]
    \includegraphics[width=\textwidth]{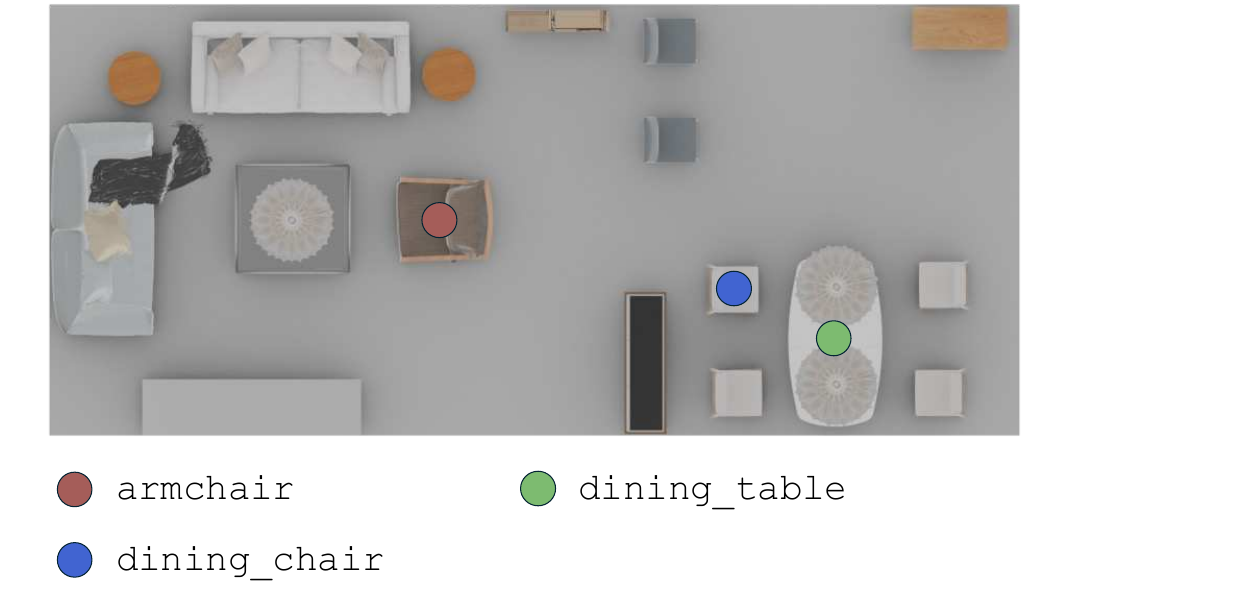}
\end{minipage}
\hfill
\begin{minipage}[c]{0.5\textwidth}
    \raggedright
    \textbf{Agent:} \textsc{adult wheelchair} \\[1.2em]
    \textbf{Task:} ``Walk to the armchair, sit down, then move to the dining table and sit on a dining chair.''
\end{minipage}
\end{table}

\paragraph{Overall Success:} \pass

\paragraph{Execution Trace:}
\begin{enumerate}
\item \textbf{Step 1:} (\texttt{navigate\_to}, \texttt{armchair}) [ID: 0] --- \textcolor{green!70!black}{\textbf{PASS}}
\begin{itemize}
    \item \texttt{is\_Navigable\_To}: \textcolor{green!70!black}{Pass} - \textit{A collision-free path was found to an interaction zone.}
\end{itemize}
\item \textbf{Step 2:} (\texttt{sit\_on}, \texttt{armchair}) [ID: 0] --- \textcolor{green!70!black}{\textbf{PASS}}
\begin{itemize}
    \item \texttt{is\_Navigable\_To}: \textcolor{green!70!black}{Pass} - \textit{A collision-free path was found to an interaction zone.}
\end{itemize}
\item \textbf{Step 3:} (\texttt{navigate\_to}, \texttt{dining\_table}) [ID: 8] --- \textcolor{green!70!black}{\textbf{PASS}}
\begin{itemize}
    \item \texttt{is\_Navigable\_To}: \textcolor{green!70!black}{Pass} - \textit{A collision-free path was found to an interaction zone.}
\end{itemize}
\item \textbf{Step 4:} (\texttt{sit\_on}, \texttt{dining\_chair}) [ID: 9] --- \textcolor{green!70!black}{\textbf{PASS}}
\begin{itemize}
    \item \texttt{is\_Navigable\_To}: \textcolor{green!70!black}{Pass} - \textit{A collision-free path was found to an interaction zone.}
\end{itemize}
\end{enumerate}

\begin{tcolorbox}[boxrule=0.5pt, colframe=cvprblue, sharp corners, left=2mm, right=2mm, top=2mm, bottom=2mm, title=\textbf{Actionable Insight}]
The plan is fully executable, as the agent can successfully navigate to and interact with all specified furniture items. The scene layout provides collision-free paths to the armchair, dining table, and dining chair, confirming that the spatial arrangement and interaction zones are well-suited for this agent's movement and seating requirements.
\end{tcolorbox}
\vspace{2em}

\clearpage
\subsubsection*{Report \#5: Child in LivingDiningRoom-13989}

\begin{table}[htbp]
\centering
\begin{minipage}[c]{0.45\textwidth}
    \raggedright
    \textbf{Scene:} \texttt{LivingDiningRoom-13989} \\ [0.6em]
    \centering
    \includegraphics[width=0.68\textwidth]{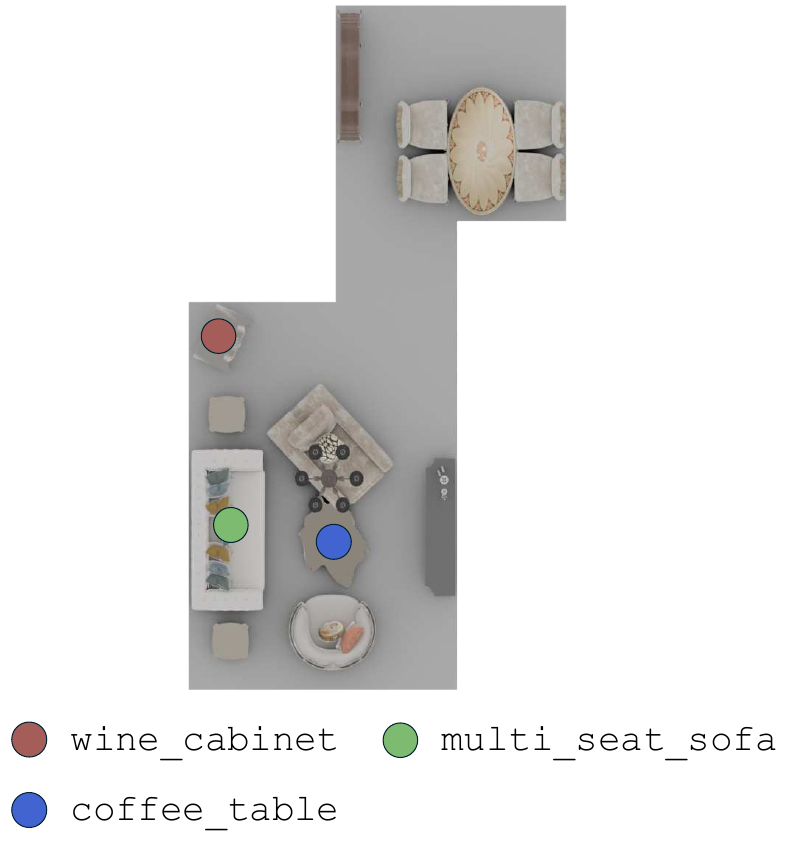}
\end{minipage}
\hfill
\begin{minipage}[c]{0.5\textwidth}
    \raggedright
    \textbf{Agent:} \textsc{child} \\[1.2em]
    \textbf{Task:} ``Go to the wine cabinet, take an item, then sit on the multi-seat sofa and place the item on the coffee table.''
\end{minipage}
\end{table}
\vspace{-1.8em}
\paragraph{Overall Success:} \fail

\paragraph{Execution Trace:}
\begin{enumerate}
\item \textbf{Step 1:} (\texttt{navigate\_to}, \texttt{wine\_cabinet}) [ID: 7] --- \textcolor{green!70!black}{\textbf{PASS}}
\begin{itemize}
    \item \texttt{is\_Navigable\_To}: \textcolor{green!70!black}{Pass} - \textit{A collision-free path was found to an interaction zone.}
\end{itemize}
\item \textbf{Step 2:} (\texttt{open}, \texttt{wine\_cabinet}) [ID: 7] --- \textcolor{green!70!black}{\textbf{PASS}}
\begin{itemize}
    \item \texttt{is\_Navigable\_To}: \textcolor{green!70!black}{Pass} - \textit{A collision-free path was found to an interaction zone.}
    \item \texttt{is\_Reachable}: \textcolor{green!70!black}{Pass} - \textit{Object is reachable. Required distance: 0.15m, Agent's reach: 0.40m.}
    \item \texttt{is\_Interactable}: \textcolor{green!70!black}{Pass} - \textit{Interactable volume is reachable. Required distance: 0.15m, Agent's reach: 0.40m.}
    \item \texttt{has\_Clearance}: \textcolor{green!70!black}{Pass} - \textit{Found 1 collision-free interaction zones.}
\end{itemize}
\item \textbf{Step 3:} (\texttt{take\_out\_of}, \texttt{wine\_cabinet}) [ID: 7] --- \textcolor{green!70!black}{\textbf{PASS}}
\begin{itemize}
    \item \texttt{is\_Navigable\_To}: \textcolor{green!70!black}{Pass} - \textit{A collision-free path was found to an interaction zone.}
    \item \texttt{is\_Reachable}: \textcolor{green!70!black}{Pass} - \textit{Object is reachable. Required distance: 0.15m, Agent's reach: 0.40m.}
    \item \texttt{is\_Interactable}: \textcolor{green!70!black}{Pass} - \textit{Interactable volume is reachable. Required distance: 0.15m, Agent's reach: 0.40m.}
    \item \texttt{has\_Clearance}: \textcolor{green!70!black}{Pass} - \textit{Found 1 collision-free interaction zones.}
\end{itemize}
\item \textbf{Step 4:} (\texttt{close}, \texttt{wine\_cabinet}) [ID: 7] --- \textcolor{green!70!black}{\textbf{PASS}}
\begin{itemize}
    \item \texttt{is\_Navigable\_To}: \textcolor{green!70!black}{Pass} - \textit{A collision-free path was found to an interaction zone.}
    \item \texttt{is\_Reachable}: \textcolor{green!70!black}{Pass} - \textit{Object is reachable. Required distance: 0.15m, Agent's reach: 0.40m.}
    \item \texttt{is\_Interactable}: \textcolor{green!70!black}{Pass} - \textit{Interactable volume is reachable. Required distance: 0.15m, Agent's reach: 0.40m.}
    \item \texttt{has\_Clearance}: \textcolor{green!70!black}{Pass} - \textit{Found 1 collision-free interaction zones.}
\end{itemize}
\item \textbf{Step 5:} (\texttt{navigate\_to}, \texttt{multi\_seat\_sofa}) [ID: 1] --- \textcolor{green!70!black}{\textbf{PASS}}
\begin{itemize}
    \item \texttt{is\_Navigable\_To}: \textcolor{green!70!black}{Pass} - \textit{A collision-free path was found to an interaction zone.}
\end{itemize}
\item \textbf{Step 6:} (\texttt{sit\_on}, \texttt{multi\_seat\_sofa}) [ID: 1] --- \textcolor{red}{\textbf{FAIL}}
\begin{itemize}
    \item \texttt{is\_Navigable\_To}: \textcolor{red}{Fail} - \textit{Agent and target zones are in different, disconnected walkable areas (Agent area: 1, Target areas: [0 6]).}
\end{itemize}
\item \textbf{Step 7:} (\texttt{release\_on}, \texttt{coffee\_table}) [ID: 3] --- \textcolor{green!70!black}{\textbf{PASS}}
\begin{itemize}
    \item \texttt{is\_Navigable\_To}: \textcolor{green!70!black}{Pass} - \textit{A collision-free path was found to an interaction zone.}
    \item \texttt{is\_Reachable}: \textcolor{green!70!black}{Pass} - \textit{Object is reachable (via crouching). Required distance: 0.15m, Agent's reach: 0.40m.}
\end{itemize}
\end{enumerate}

\begin{tcolorbox}[boxrule=0.5pt, colframe=cvprblue, sharp corners, left=2mm, right=2mm, top=2mm, bottom=2mm, title=\textbf{Actionable Insight}]
The plan fails because the agent cannot \texttt{sit\_on} the \texttt{multi\_seat\_sofa}. While the agent can navigate to the general vicinity of the sofa, the specific interaction zones required for sitting are located in disconnected walkable areas (Areas 0 and 6) that are inaccessible from the agent's current position in Area 1. This suggests a connectivity issue in the scene's navigation mesh or an obstruction preventing the agent from reaching the functional side of the sofa.
\end{tcolorbox}
\vspace{2em}

\clearpage
\subsubsection*{Report \#6: Adult in LivingDiningRoom-25741}

\begin{table}[htbp]
\centering
\begin{minipage}[c]{0.45\textwidth}
    \raggedright
    \textbf{Scene:} \texttt{LivingDiningRoom-25741} \\ [0.6em]
    \centering
    \includegraphics[width=0.8\textwidth]{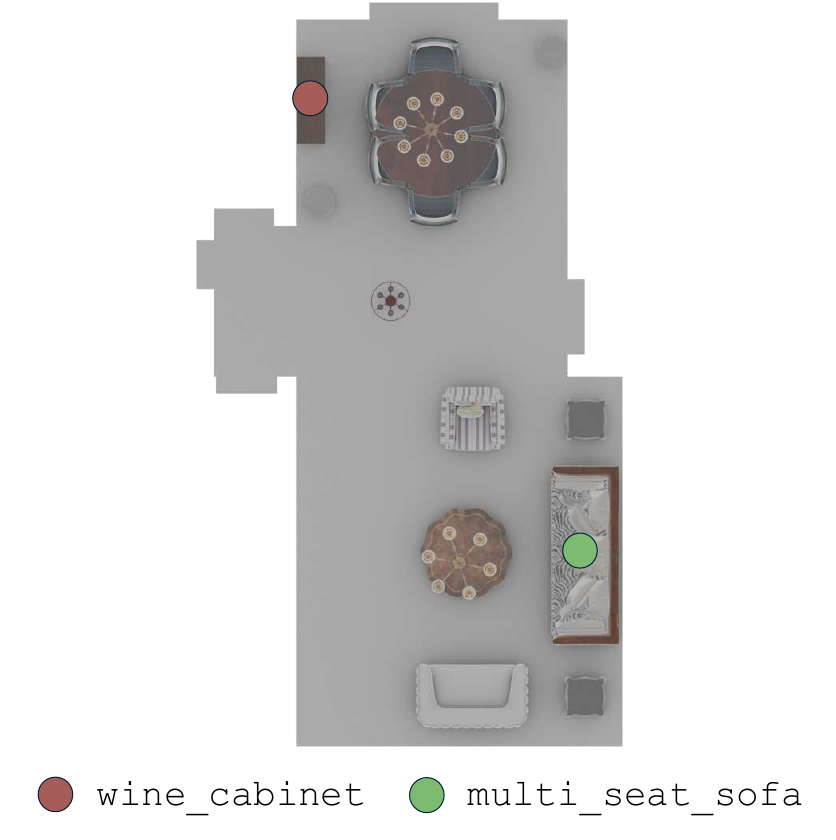}
\end{minipage}
\hfill
\begin{minipage}[c]{0.5\textwidth}
    \raggedright
    \textbf{Agent:} \textsc{adult} \\[1.2em]
    \textbf{Task:} ``Go to the wine cabinet, take an item, then walk to the multi-seat sofa and sit down.''
\end{minipage}
\end{table}

\paragraph{Overall Success:} \pass

\paragraph{Execution Trace:}
\begin{enumerate}
\item \textbf{Step 1:} (\texttt{navigate\_to}, \texttt{wine\_cabinet}) [ID: 11] --- \textcolor{green!70!black}{\textbf{PASS}}
\begin{itemize}
    \item \texttt{is\_Navigable\_To}: \textcolor{green!70!black}{Pass} - \textit{A collision-free path was found to an interaction zone.}
\end{itemize}
\item \textbf{Step 2:} (\texttt{open}, \texttt{wine\_cabinet}) [ID: 11] --- \textcolor{green!70!black}{\textbf{PASS}}
\begin{itemize}
    \item \texttt{is\_Navigable\_To}: \textcolor{green!70!black}{Pass} - \textit{A collision-free path was found to an interaction zone.}
    \item \texttt{is\_Reachable}: \textcolor{green!70!black}{Pass} - \textit{Object is reachable. Required distance: 0.18m, Agent's reach: 0.70m.}
    \item \texttt{is\_Interactable}: \textcolor{green!70!black}{Pass} - \textit{Interactable volume is reachable. Required distance: 0.18m, Agent's reach: 0.70m.}
    \item \texttt{has\_Clearance}: \textcolor{green!70!black}{Pass} - \textit{Found 1 collision-free interaction zones.}
\end{itemize}
\item \textbf{Step 3:} (\texttt{take\_out\_of}, \texttt{wine\_cabinet}) [ID: 11] --- \textcolor{green!70!black}{\textbf{PASS}}
\begin{itemize}
    \item \texttt{is\_Navigable\_To}: \textcolor{green!70!black}{Pass} - \textit{A collision-free path was found to an interaction zone.}
    \item \texttt{is\_Reachable}: \textcolor{green!70!black}{Pass} - \textit{Object is reachable. Required distance: 0.18m, Agent's reach: 0.70m.}
    \item \texttt{is\_Interactable}: \textcolor{green!70!black}{Pass} - \textit{Interactable volume is reachable. Required distance: 0.18m, Agent's reach: 0.70m.}
    \item \texttt{has\_Clearance}: \textcolor{green!70!black}{Pass} - \textit{Found 1 collision-free interaction zones.}
\end{itemize}
\item \textbf{Step 4:} (\texttt{close}, \texttt{wine\_cabinet}) [ID: 11] --- \textcolor{green!70!black}{\textbf{PASS}}
\begin{itemize}
    \item \texttt{is\_Navigable\_To}: \textcolor{green!70!black}{Pass} - \textit{A collision-free path was found to an interaction zone.}
    \item \texttt{is\_Reachable}: \textcolor{green!70!black}{Pass} - \textit{Object is reachable. Required distance: 0.18m, Agent's reach: 0.70m.}
    \item \texttt{is\_Interactable}: \textcolor{green!70!black}{Pass} - \textit{Interactable volume is reachable. Required distance: 0.18m, Agent's reach: 0.70m.}
    \item \texttt{has\_Clearance}: \textcolor{green!70!black}{Pass} - \textit{Found 1 collision-free interaction zones.}
\end{itemize}
\item \textbf{Step 5:} (\texttt{navigate\_to}, \texttt{multi\_seat\_sofa}) [ID: 4] --- \textcolor{green!70!black}{\textbf{PASS}}
\begin{itemize}
    \item \texttt{is\_Navigable\_To}: \textcolor{green!70!black}{Pass} - \textit{A collision-free path was found to an interaction zone.}
\end{itemize}
\item \textbf{Step 6:} (\texttt{sit\_on}, \texttt{multi\_seat\_sofa}) [ID: 4] --- \textcolor{green!70!black}{\textbf{PASS}}
\begin{itemize}
    \item \texttt{is\_Navigable\_To}: \textcolor{green!70!black}{Pass} - \textit{A collision-free path was found to an interaction zone.}
\end{itemize}
\end{enumerate}

\begin{tcolorbox}[boxrule=0.5pt, colframe=cvprblue, sharp corners, left=2mm, right=2mm, top=2mm, bottom=2mm, title=\textbf{Actionable Insight}]
The plan is fully executable, as the agent can successfully navigate the environment and interact with all target objects. The agent maintains ample physical clearance for the wine cabinet, with a required reach of only 0.18m against a 0.70m capacity, and can transition smoothly from the cabinet to sitting on the sofa. This confirms that the scene layout and object spacing are well-optimized for this agent's specific reach and mobility constraints.
\end{tcolorbox}
\vspace{2em}

\subsection{VLM Predictions}
\label{sec:additional:vlm}

Below, we showcase VLM prediction results under the benchmark settings described in Section~\ref{supp:benchmark}. We present outputs for both task-level and atomic-action-level evaluations. Notably, in the \textit{Decomposed} action-level setting, the prompt rendering highlights the target object with a red bounding box.

\subsubsection*{VLM Task Evaluation: Adult Wheelchair in LivingRoom-69328}

\begin{table}[htbp]
\centering
\begin{minipage}[c]{0.45\textwidth}
    \raggedright
    \textbf{Scene:} \texttt{LivingRoom-69328} \\ [0.6em]
    \includegraphics[width=\textwidth]{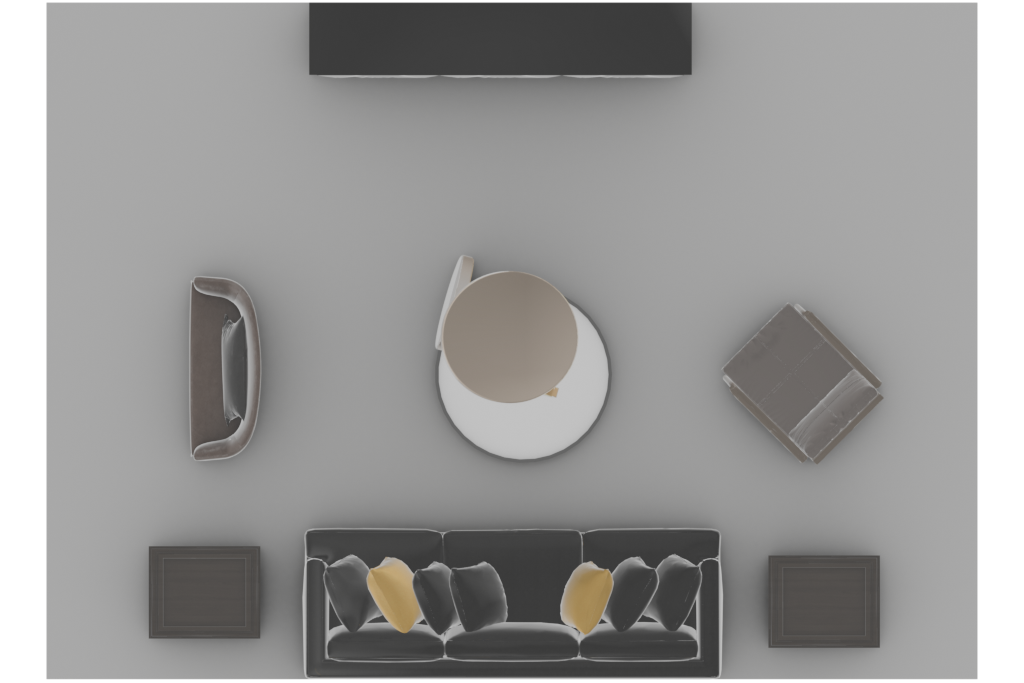}
\end{minipage}
\hfill
\begin{minipage}[c]{0.5\textwidth}
    \raggedright
    \textbf{Agent:} \textsc{adult wheelchair} \\[1.2em]
    \textbf{Task:} ``Navigate to the lounge chair, sit down, and then look towards the coffee table.\\[1.2em]
    \textbf{\name{} Verification:} \textcolor{green!70!black}{\textbf{PASS}}
\end{minipage}
\end{table}

\paragraph{Model Predictions:}\mbox{}
\vspace{0.5em}

\noindent\textbf{\texttt{Gemini-3-Flash-Preview}} (\textcolor{green!70!black}{$\checkmark$ Correct}) \\
\textbf{Prediction:} \textcolor{green!70!black}{\textbf{Possible (PASS)}} \\
\textbf{Reasoning:} \textit{The room features an open floor plan with wide paths and no obstructions. There is ample space for a manual wheelchair to navigate to the lounge chair and sufficient clearance for the agent to perform a transfer. Once seated, the coffee table is directly within the agent's line of sight.}
\vspace{0.8em}

\noindent\textbf{\texttt{Gemini-3.1-Pro-Preview}} (\textcolor{green!70!black}{$\checkmark$ Correct}) \\
\textbf{Prediction:} \textcolor{green!70!black}{\textbf{Possible (PASS)}} \\
\textbf{Reasoning:} \textit{The room has an open layout with ample clearance between all pieces of furniture. There is plenty of space for the wheelchair user to navigate to the lounge chair, position the wheelchair beside it to transfer, and once seated, there are no obstructions blocking the line of sight to the coffee table in the center of the room.}
\vspace{0.8em}

\noindent\textbf{\texttt{Claude-Sonnet-4-6}} (\textcolor{green!70!black}{$\checkmark$ Correct}) \\
\textbf{Prediction:} \textcolor{green!70!black}{\textbf{Possible (PASS)}} \\
\textbf{Reasoning:} \textit{The room has sufficient open floor space for wheelchair navigation. The lounge chair (large beige rounded chair in the center) is accessible from multiple directions with adequate clearance between furniture pieces. The coffee table (white circular object adjacent to the lounge chair) is in close proximity. Interpreting 'sit down' as navigating to and positioning at the lounge chair, the agent can navigate there and orient toward the coffee table without significant spatial obstruction.}
\vspace{0.8em}

\noindent\textbf{\texttt{Gemma3-12B-Instruct}} (\textcolor{green!70!black}{$\checkmark$ Correct}) \\
\textbf{Prediction:} \textcolor{green!70!black}{\textbf{Possible (PASS)}} \\
\textbf{Reasoning:} \textit{The room appears to have sufficient space for a wheelchair to navigate. The lounge chair is accessible, and the coffee table is positioned in a way that the agent, seated in the lounge chair, could reasonably look towards it. There are no obvious obstructions preventing the agent from reaching the chair and then orienting themselves towards the table.}
\vspace{0.8em}

\noindent\textbf{\texttt{Gemma3-4B-Instruct}} (\textcolor{green!70!black}{$\checkmark$ Correct}) \\
\textbf{Prediction:} \textcolor{green!70!black}{\textbf{Possible (PASS)}} \\
\textbf{Reasoning:} \textit{The agent, in their wheelchair, can navigate to the lounge chair. There is sufficient space around the lounge chair to maneuver the wheelchair and enter it. Once seated, the coffee table is within reach for looking, and there is enough clearance between the chair and the table. The paths appear wide enough for the wheelchair to traverse.}
\vspace{0.8em}

\noindent\textbf{\texttt{Ministral3-3B-Instruct}} (\textcolor{green!70!black}{$\checkmark$ Correct}) \\
\textbf{Prediction:} \textcolor{green!70!black}{\textbf{Possible (PASS)}} \\
\textbf{Reasoning:} \textit{The agent can navigate to the lounge chair and the coffee table based on the spatial layout. The room appears to have sufficient open space for a wheelchair to maneuver. The lounge chair is positioned within reachable distance from the seating area of the sofa, and the coffee table is centrally located, allowing the agent to look towards it once seated in the lounge chair without obstruction.}
\vspace{0.8em}

\noindent\textbf{\texttt{Qwen3-VL-4B-Instruct}} (\textcolor{red}{$\times$ Incorrect}) \\
\textbf{Prediction:} \textcolor{red}{\textbf{Impossible (FAIL)}} \\
\textbf{Reasoning:} \textit{The lounge chair is positioned too far from the coffee table and lacks clear, unobstructed pathways for the wheelchair to navigate. The agent cannot reach the lounge chair without significant spatial constraints, and even if it could, the coffee table is not directly accessible from the chair's position without obstruction or requiring a turn that may not be feasible for a wheelchair.}
\vspace{0.8em}

\noindent\textbf{\texttt{Qwen3-VL-4B-Instruct}} \textit{with} GRPO (\textbf{ours}) (\textcolor{green!70!black}{$\checkmark$ Correct}) \\
\textbf{Prediction:} \textcolor{green!70!black}{\textbf{Possible (PASS)}} \\
\textbf{Reasoning:} \textit{The top-down view shows clear spatial pathways for the wheelchair to navigate, with no obstructive furniture or narrow passages. The lounge chair is accessible and positioned to allow the agent to sit down, and the coffee table is within reach and visible in the layout, enabling the agent to look towards it after sitting.}
\vspace{0.8em}

\vspace{1em}
\vspace{2em}

\clearpage
\subsubsection*{VLM Task Evaluation: Adult Wheelchair in LivingDiningRoom-108722}

\begin{table}[htbp]
\centering
\begin{minipage}[c]{0.45\textwidth}
    \raggedright
    \textbf{Scene:} \texttt{LivingDiningRoom-108722} \\ [0.6em]
    \includegraphics[width=\textwidth]{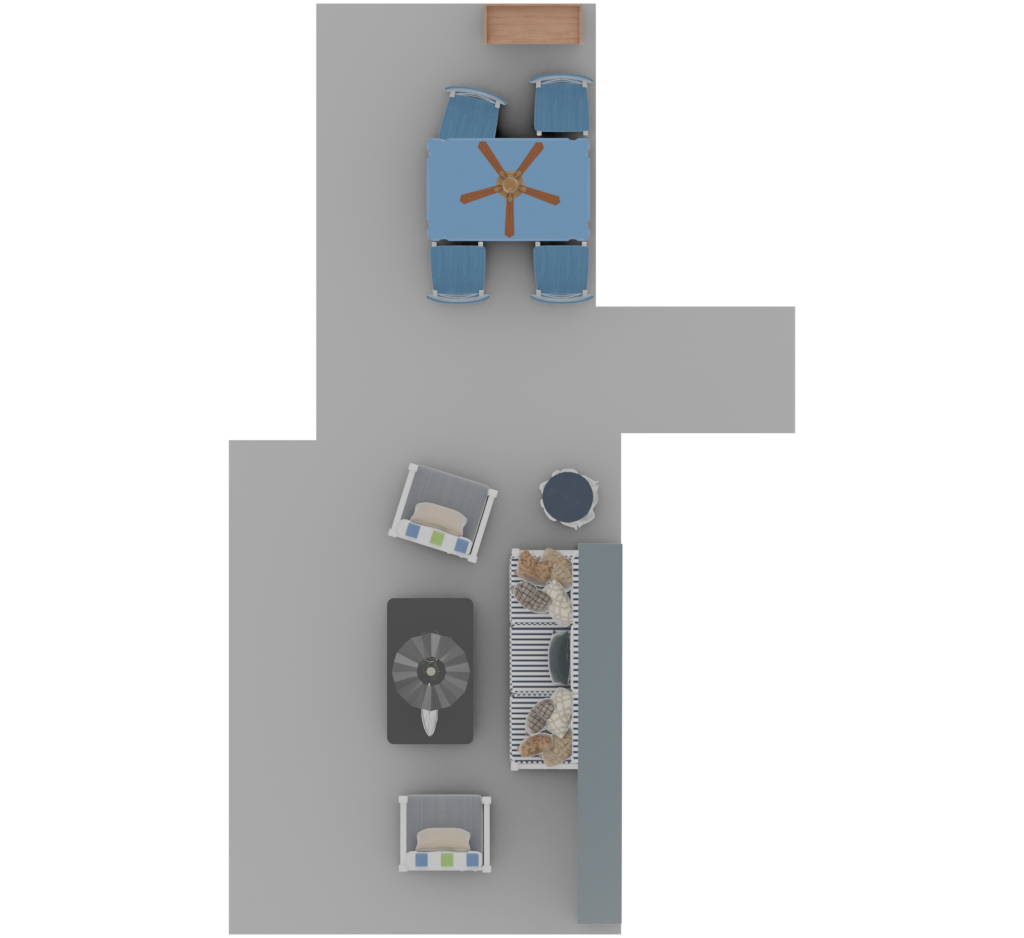}
\end{minipage}
\hfill
\begin{minipage}[c]{0.5\textwidth}
    \raggedright
    \textbf{Agent:} \textsc{adult wheelchair} \\[1.2em]
    \textbf{Task:} ``Walk to the bookshelf, then sit on the multi-seat sofa and look at the coffee table.\\[1.2em]
    \textbf{\name{} Verification:} \textcolor{red}{\textbf{FAIL}}
\end{minipage}
\end{table}

\paragraph{Model Predictions:}\mbox{}
\vspace{0.5em}

\noindent\textbf{\texttt{Gemini-3-Flash-Preview}} (\textcolor{green!70!black}{$\checkmark$ Correct}) \\
\textbf{Prediction:} \textcolor{red}{\textbf{Impossible (FAIL)}} \\
\textbf{Reasoning:} \textit{The bookshelf is located at the far end of the room, directly behind the dining table and chairs. The spatial clearance between the dining set and the wall where the bookshelf is situated is too narrow for a manual wheelchair to navigate through, making the bookshelf physically inaccessible to the agent.}
\vspace{0.8em}

\noindent\textbf{\texttt{Gemini-3.1-Pro-Preview}} (\textcolor{green!70!black}{$\checkmark$ Correct}) \\
\textbf{Prediction:} \textcolor{red}{\textbf{Impossible (FAIL)}} \\
\textbf{Reasoning:} \textit{The task is impossible because the pathways to the multi-seat sofa are blocked. The clearance between the coffee table and the sofa, as well as the gaps between the coffee table and the adjacent armchairs, are far too narrow to accommodate the width of a wheelchair, preventing the agent from reaching the sofa to sit on it.}
\vspace{0.8em}

\noindent\textbf{\texttt{Claude-Sonnet-4-6}} (\textcolor{red}{$\times$ Incorrect}) \\
\textbf{Prediction:} \textcolor{green!70!black}{\textbf{Possible (PASS)}} \\
\textbf{Reasoning:} \textit{The room has sufficient open floor space for wheelchair navigation throughout. The bookshelf at the top of the room is accessible via the open corridor past the dining area. The multi-seat sofa in the lower living area has open space on at least one side for wheelchair approach. The coffee table is positioned directly in front of the sofa area, making it visible from that location. No narrow bottlenecks or blocking obstacles prevent the wheelchair from reaching these targets.}
\vspace{0.8em}

\noindent\textbf{\texttt{Gemma3-12B-Instruct}} (\textcolor{red}{$\times$ Incorrect}) \\
\textbf{Prediction:} \textcolor{green!70!black}{\textbf{Possible (PASS)}} \\
\textbf{Reasoning:} \textit{The room appears to have sufficient space for a wheelchair to navigate. There are clear paths between the starting location, the bookshelf, the sofa, and the coffee table. The image suggests adequate clearance around the furniture for the agent to maneuver and reach the coffee table while seated on the sofa.}
\vspace{0.8em}

\noindent\textbf{\texttt{Gemma3-4B-Instruct}} (\textcolor{red}{$\times$ Incorrect}) \\
\textbf{Prediction:} \textcolor{green!70!black}{\textbf{Possible (PASS)}} \\
\textbf{Reasoning:} \textit{The agent can navigate from their starting position to the bookshelf and then to the sofa. There are no narrow passages or obstructions preventing wheelchair access. The sofa and coffee table are within reach of the agent. The path appears clear and wide enough for the wheelchair to traverse.}
\vspace{0.8em}

\noindent\textbf{\texttt{Ministral3-3B-Instruct}} (\textcolor{red}{$\times$ Incorrect}) \\
\textbf{Prediction:} \textcolor{green!70!black}{\textbf{Possible (PASS)}} \\
\textbf{Reasoning:} \textit{The agent can navigate to the bookshelf and the multi-seat sofa due to sufficient clear pathways in the room. The top-down view shows ample space for a manual wheelchair to maneuver. The bookshelf is positioned within reachable distance from the seating area, and the multi-seat sofa is accessible for the agent to sit on. The coffee table is also within the reachable area on the sofa, allowing the agent to look at it.}
\vspace{0.8em}

\noindent\textbf{\texttt{Qwen3-VL-4B-Instruct}} (\textcolor{green!70!black}{$\checkmark$ Correct}) \\
\textbf{Prediction:} \textcolor{red}{\textbf{Impossible (FAIL)}} \\
\textbf{Reasoning:} \textit{The agent's path from the dining area to the bookshelf is blocked by the dining table and chairs, and the bookshelf is not visible in the provided image. Additionally, the multi-seat sofa is located adjacent to the bookshelf, but the agent's wheelchair would require clear space to maneuver around the table and reach the sofa, which is not spatially feasible given the layout.}
\vspace{0.8em}

\noindent\textbf{\texttt{Qwen3-VL-4B-Instruct}} \textit{with} GRPO (\textbf{ours}) (\textcolor{green!70!black}{$\checkmark$ Correct}) \\
\textbf{Prediction:} \textcolor{red}{\textbf{Impossible (FAIL)}} \\
\textbf{Reasoning:} \textit{The agent cannot navigate to the bookshelf or multi-seat sofa due to spatial constraints; the bookshelf is not clearly accessible, and the sofa is obstructed or not reachable from the agent's position, preventing completion of the task.}
\vspace{0.8em}

\vspace{1em}
\vspace{2em}

\clearpage
\subsubsection*{Action-Level Evaluation: Sit On in LivingDiningRoom-15286}

\begin{table}[htbp]
\centering
\begin{minipage}[c]{0.45\textwidth}
    \raggedright
    \textbf{Scene:} \texttt{LivingDiningRoom-15286} \\ [0.6em]
    \includegraphics[width=\textwidth]{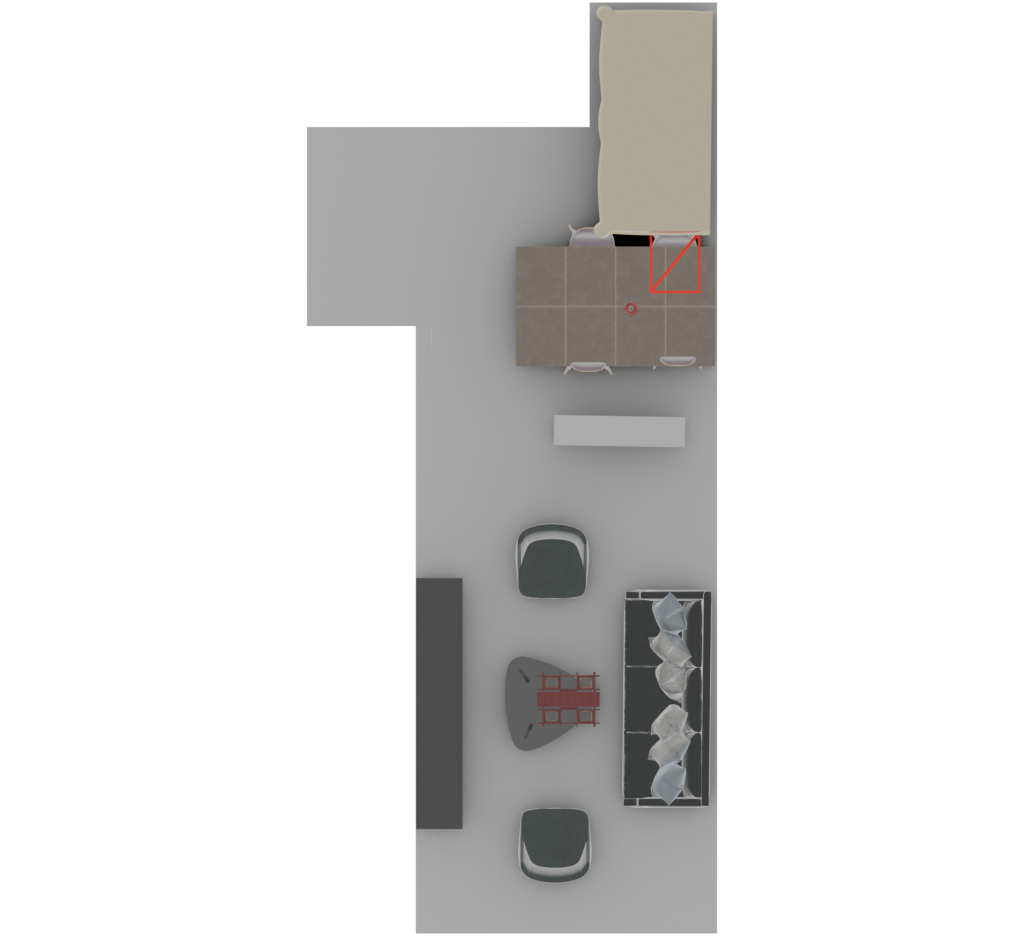}
\end{minipage}
\hfill
\begin{minipage}[c]{0.5\textwidth}
    \raggedright
    \textbf{Agent:} \textsc{adult wheelchair} \\
    \textbf{Action Detail:} (\texttt{sit\_on}, \texttt{dining\_chair}) \\[1.2em]
    \textbf{\name{} Verification:} \textcolor{red}{\textbf{FAIL (Is Navigable To)}}
\end{minipage}
\end{table}

\paragraph{Model Predictions:}\mbox{}
\vspace{0.5em}

\noindent\textbf{\texttt{Gemini-3-Flash-Preview}} (\textcolor{green!70!black}{$\checkmark$ Correct}) \\
\textbf{Prediction:} \textcolor{red}{\textbf{Impossible (FAIL)}}
\vspace{0.8em}

\noindent\textbf{\texttt{Gemini-3.1-Pro-Preview}} (\textcolor{green!70!black}{$\checkmark$ Correct}) \\
\textbf{Prediction:} \textcolor{red}{\textbf{Impossible (FAIL)}}
\vspace{0.8em}

\noindent\textbf{\texttt{Claude-Sonnet-4-6}} (\textcolor{green!70!black}{$\checkmark$ Correct}) \\
\textbf{Prediction:} \textcolor{red}{\textbf{Impossible (FAIL)}}
\vspace{0.8em}

\noindent\textbf{\texttt{Qwen3-VL-8B-Instruct}} (\textcolor{green!70!black}{$\checkmark$ Correct}) \\
\textbf{Prediction:} \textcolor{red}{\textbf{Impossible (FAIL)}}
\vspace{0.8em}

\noindent\textbf{\texttt{Gemma3-12B-Instruct}} (\textcolor{red}{$\times$ Incorrect}) \\
\textbf{Prediction:} \textcolor{green!70!black}{\textbf{Possible (PASS)}}
\vspace{0.8em}

\noindent\textbf{\texttt{Gemma3-4B-Instruct}} (\textcolor{red}{$\times$ Incorrect}) \\
\textbf{Prediction:} \textcolor{green!70!black}{\textbf{Possible (PASS)}}
\vspace{0.8em}

\noindent\textbf{\texttt{Ministral3-3B-Instruct}} (\textcolor{red}{$\times$ Incorrect}) \\
\textbf{Prediction:} \textcolor{green!70!black}{\textbf{Possible (PASS)}}
\vspace{0.8em}

\noindent\textbf{\texttt{Qwen3-VL-4B-Instruct}} (\textcolor{red}{$\times$ Incorrect}) \\
\textbf{Prediction:} \textcolor{green!70!black}{\textbf{Possible (PASS)}}
\vspace{0.8em}

\noindent\textbf{\texttt{Qwen3-VL-4B-Instruct}} \textit{with} GRPO (\textbf{ours}) (\textcolor{green!70!black}{$\checkmark$ Correct}) \\
\textbf{Prediction:} \textcolor{red}{\textbf{Impossible (FAIL)}}
\vspace{0.8em}

\vspace{2em}

\clearpage
\subsubsection*{Action-Level Evaluation: Release On in LivingRoom-6833}

\begin{table}[htbp]
\centering
\begin{minipage}[c]{0.45\textwidth}
    \raggedright
    \textbf{Scene:} \texttt{LivingRoom-6833} \\ [0.6em]
    \includegraphics[width=\textwidth]{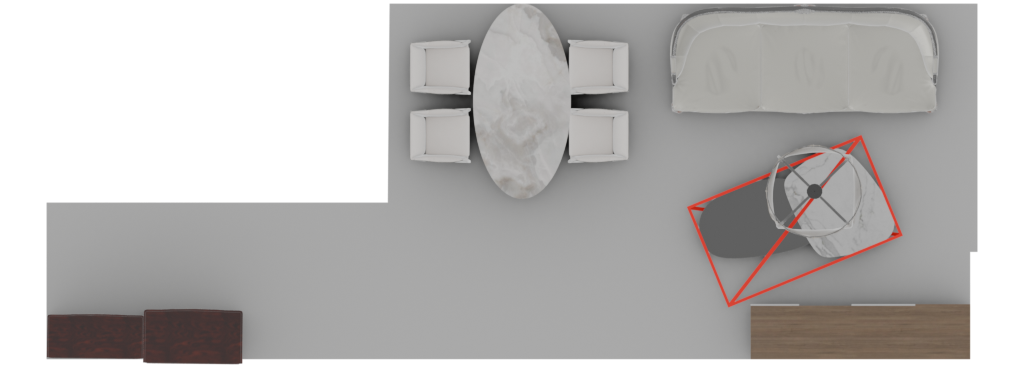}
\end{minipage}
\hfill
\begin{minipage}[c]{0.5\textwidth}
    \raggedright
    \textbf{Agent:} \textsc{child} \\
    \textbf{Action Detail:} (\texttt{release\_on}, \texttt{coffee\_table}) \\ [1.2em]
    \textbf{\name{} Verification:} \textcolor{green!70!black}{\textbf{PASS}}
\end{minipage}
\end{table}

\paragraph{Model Predictions:}\mbox{}
\vspace{0.5em}

\noindent\textbf{\texttt{Gemini-3-Flash-Preview}} (\textcolor{green!70!black}{$\checkmark$ Correct}) \\
\textbf{Prediction:} \textcolor{green!70!black}{\textbf{Possible (PASS)}}
\vspace{0.8em}

\noindent\textbf{\texttt{Gemini-3.1-Pro-Preview}} (\textcolor{green!70!black}{$\checkmark$ Correct}) \\
\textbf{Prediction:} \textcolor{green!70!black}{\textbf{Possible (PASS)}}
\vspace{0.8em}

\noindent\textbf{\texttt{Claude-Sonnet-4-6}} (\textcolor{green!70!black}{$\checkmark$ Correct}) \\
\textbf{Prediction:} \textcolor{green!70!black}{\textbf{Possible (PASS)}}
\vspace{0.8em}

\noindent\textbf{\texttt{Qwen3-VL-8B-Instruct}} (\textcolor{red}{$\times$ Incorrect}) \\
\textbf{Prediction:} \textcolor{red}{\textbf{Impossible (FAIL)}}
\vspace{0.8em}

\noindent\textbf{\texttt{Gemma3-12B-Instruct}} (\textcolor{green!70!black}{$\checkmark$ Correct}) \\
\textbf{Prediction:} \textcolor{green!70!black}{\textbf{Possible (PASS)}}
\vspace{0.8em}

\noindent\textbf{\texttt{Gemma3-4B-Instruct}} (\textcolor{green!70!black}{$\checkmark$ Correct}) \\
\textbf{Prediction:} \textcolor{green!70!black}{\textbf{Possible (PASS)}}
\vspace{0.8em}

\noindent\textbf{\texttt{Ministral3-3B-Instruct}} (\textcolor{red}{$\times$ Incorrect}) \\
\textbf{Prediction:} \textcolor{red}{\textbf{Impossible (FAIL)}}
\vspace{0.8em}

\noindent\textbf{\texttt{Qwen3-VL-4B-Instruct}} (\textcolor{red}{$\times$ Incorrect}) \\
\textbf{Prediction:} \textcolor{red}{\textbf{Impossible (FAIL)}}
\vspace{0.8em}

\noindent\textbf{\texttt{Qwen3-VL-4B-Instruct}} \textit{with} GRPO (\textbf{ours}) (\textcolor{green!70!black}{$\checkmark$ Correct}) \\
\textbf{Prediction:} \textcolor{green!70!black}{\textbf{Possible (PASS)}}
\vspace{0.8em}

\vspace{2em}

\twocolumn

\end{document}